\pdfoutput=1

\documentclass[11pt,a4paper]{article} 
\usepackage{setspace}
\usepackage[hidelinks]{hyperref}

\usepackage{amsmath,amsfonts,amsthm,amssymb,bm}

\def\eqref#1{equation~\ref{#1}}

\def\1{\bm{1}}

\DeclareMathAlphabet{\mathsfit}{\encodingdefault}{\sfdefault}{m}{sl}
\SetMathAlphabet{\mathsfit}{bold}{\encodingdefault}{\sfdefault}{bx}{n}

\newcommand{\E}{\mathbb{E}}

\DeclareMathOperator*{\argmax}{arg\,max}

\newcommand{\sv}[1]{\ensuremath{\llbracket #1 \rrbracket}}

\theoremstyle{definition}

\theoremstyle{definition}

\usepackage{style}
\usepackage{url}

\usepackage{gb4e}
\noautomath

\usepackage[acceptedWithA]{tacl2021v1}

\usepackage{microtype}

\usepackage{inconsolata}

\usepackage{centernot}
\usepackage{wasysym}
\usepackage{marvosym}
\usepackage{cuted}

\usepackage{forest}

\usepackage{makecell}

\usepackage{adjustbox}
\usepackage{dashbox}

\definecolor{boxbg}{rgb}{0.85,0.85,0.85}

\usepackage{xspace,mfirstuc,tabulary}

\newif\iftaclinstructions
\taclinstructionsfalse %
\iftaclinstructions
\renewcommand{\confidential}{}
\renewcommand{\anonsubtext}{(No author info supplied here, for consistency with
TACL-submission anonymization requirements)}
\newcommand{\instr}
\fi

\definecolor{orange}{rgb}{1,0.5,0}
\definecolor{mdred}{rgb}{0.7,0,0}
\definecolor{mdgreen}{rgb}{0.05,0.6,0.05}
\definecolor{mdblue}{rgb}{0,0,0.7}
\definecolor{dkblue}{rgb}{0,0,0.5}
\definecolor{dkgray}{rgb}{0.3,0.3,0.3}
\definecolor{slate}{rgb}{0.25,0.25,0.4}
\definecolor{gray}{rgb}{0.5,0.5,0.5}
\definecolor{ltgray}{rgb}{0.7,0.7,0.7}
\definecolor{purple}{rgb}{0.7,0,1.0}
\definecolor{lavender}{rgb}{0.65,0.55,1.0}
\definecolor{real}{HTML}{3498DB}
\definecolor{cf}{HTML}{DB7734}

\newcommand{\papercomment}[3]{\ensuretext{\textcolor{#3}{[#1 #2]}}}
\renewcommand{\papercomment}[3]{}  %

\usepackage{xspace}
\renewcommand{\eg}{e.g.,\xspace}%

\title{Reasoning or Reciting? Exploring the Capabilities and Limitations of  Language Models Through Counterfactual Tasks}

\author{Zhaofeng Wu$^\text{\Cancer}$ \quad
    Linlu Qiu$^\text{\Cancer}$ \quad
    Alexis Ross$^\text{\Cancer}$ \quad
    Ekin Akyürek$^\text{\Cancer}$ \quad
    Boyuan Chen$^\text{\Cancer}$ \\
    \textbf{Bailin Wang}$^\text{\Cancer}$ \quad
    \textbf{Najoung Kim}$^\text{\Libra}$ \quad
    \textbf{Jacob Andreas}$^\text{\Cancer}$ \quad
    \textbf{Yoon Kim}$^\text{\Cancer}$ \quad\\
    $^\text{\Cancer}$MIT \quad $^\text{\Libra}$Boston University \\
    \texttt{zfw@csail.mit.edu}
}

\begin{document}
\maketitle

\begin{strip}  %
    \vspace{-1cm}
    \centering
    \includegraphics[width=\textwidth]{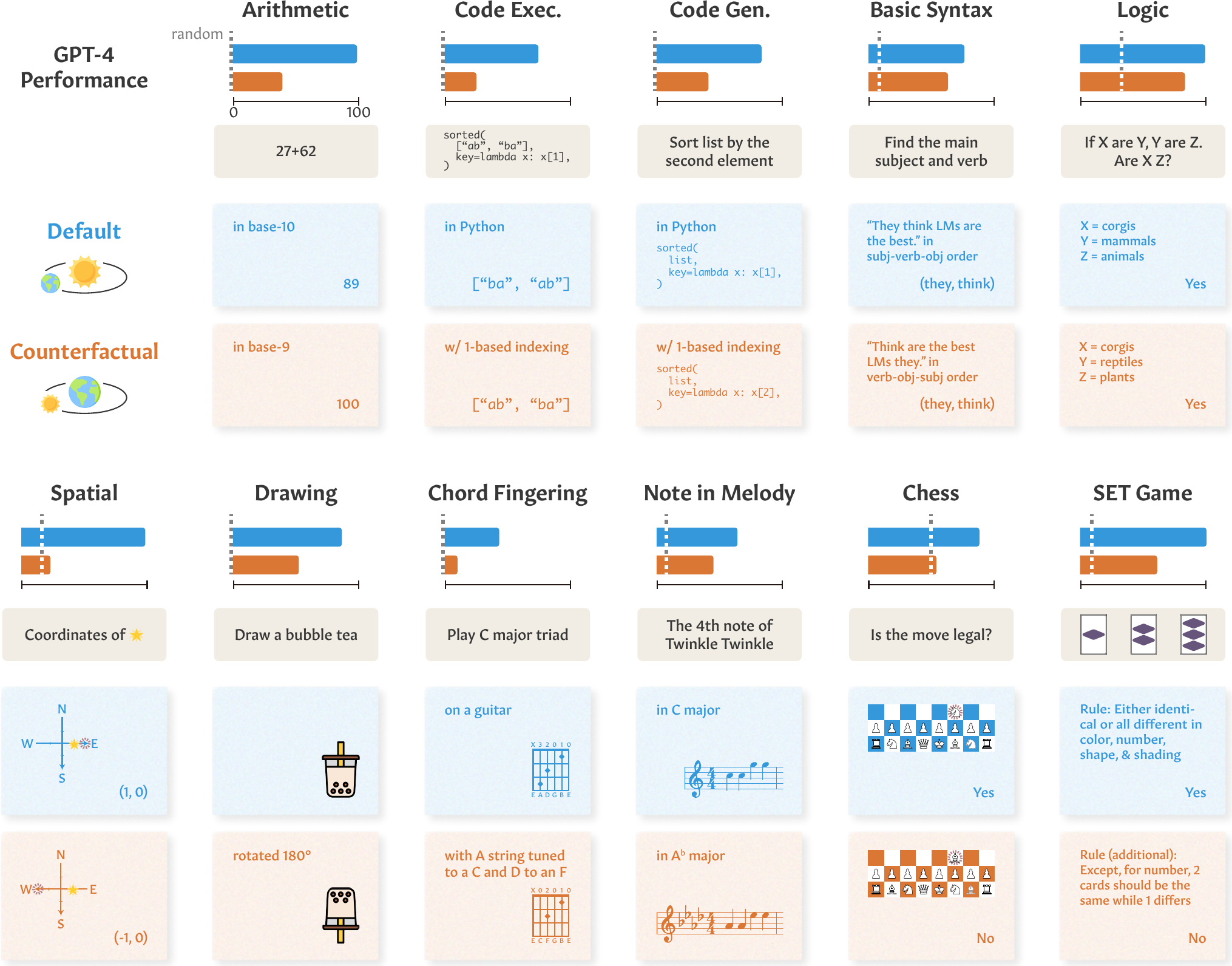}
    \captionof{figure}{GPT-4's performance on the default version of various tasks (\textcolor{real}{blue}) and counterfactual counterparts (\textcolor{cf}{orange}). The shown results use 0-shot chain-of-thought prompting (\S\ref{sec:results}; \citealp{kojima2023large}).
    GPT-4 consistently and substantially underperforms on counterfactual variants compared to default task instantiations.
    }
    \label{fig:overview}
\end{strip}

\begin{abstract}

The impressive performance of recent language models across a wide range of tasks suggests that they possess a degree of abstract reasoning skills. Are these skills general and transferable, or specialized to specific tasks seen during pretraining? To disentangle these effects, we propose an evaluation framework based on ``counterfactual'' task variants that deviate from 
the default assumptions underlying standard tasks. Across a suite of 11 tasks, we observe nontrivial performance on the counterfactual variants, but nevertheless find that performance substantially and consistently degrades compared to the default conditions.
This suggests that while current LMs may possess abstract task-solving skills to an extent, they often also rely on narrow, non-transferable procedures for task-solving.
These results motivate a more careful interpretation of language model performance that teases apart these aspects of behavior. 

\end{abstract}

\section{Introduction} \label{sec:intro}

\blfootnote{We release our code, all synthetically generated data, and LM interactions (prompts and responses) at \url{https://github.com/ZhaofengWu/counterfactual-evaluation}.}
\hspace{-.7em}
The striking empirical successes of language models (LMs) suggest 
that next-word prediction at scale
may be a viable approach for distilling the knowledge embedded in large-scale text corpora into general-purpose interactive agents. LMs obtain impressive results on various NLP benchmarks~\citepia{gpt4,palm2,claude} and display surprising abilities that suggest a nontrivial understanding of the  world~\citep{bubeck2023sparks}. They have been shown to pass professional exams~\citepia{kung2023performance,nori2023capabilities,terwiesch2023would}, exceed state-of-the-art methods on many traditional benchmarks~\citepia{sun2023chatgpt,sobania-2023-analysis,zhang2023stance,dhingra2023mind}, and surpass human performance on tasks that require seemingly nontrivial reasoning~\citepia{chowdhery2022palm,hoffmann2022training,malinka2023educational,guo2023close}. 

Ideally, we expect a general-purpose LM to be able to \emph{generalize} not only to unseen instances of known tasks, but to new tasks. Humans, for example, can transfer their knowledge to new instances and also flexibly adapt to novel tasks~\citep{singley1989transfer}. To what extent does the performance of current LMs derive from their ability to deploy task-general reasoning skills, versus their ability to recognize and recall specific tasks seen frequently in pre-training?

Past work has focused on instance-level generalization, but this is often complicated by data contamination issues~\citepia{dodge-etal-2021-documenting,magar-schwartz-2022-data}. In this work, we are interested in the models' generalizability to new task variants, which has been less systematically studied for LMs (though see \citet{li-etal-2022-quantifying}, \citet{mishra-etal-2022-cross}, and \citet{wang-etal-2022-super}). 

We propose to measure such task-level generalizability by taking tasks on which LMs perform well, and altering the conditions or rules under which these tasks are performed.
The general reasoning procedure for these tasks remains the same under the new conditions, but the specific input-output mapping functions are changed.
We call the new tasks \emph{counterfactual tasks}, as they deviate from the default, generally assumed conditions for these tasks. Figure~\ref{fig:overview} shows examples: in the top left, default arithmetic is performed in base-10, while counterfactual arithmetic is performed in base 9.
If models implement a general and transferable task-solving procedure, we expect comparable performance on counterfactual and default tasks; if they employ procedures tailored to default task conditions, we expect a drop in the counterfactual performance.

We design a suite of 11 counterfactual evaluation tasks to measure an LM's flexibility to adapt to new task variants across multiple categories and domains, as summarized in Figure~\ref{fig:overview}.
In each, the original task under the default conditions and its counterfactual variants share the same reasoning procedure but differ in their input-output mappings.
We consider  traditional NLP tasks such as deductive reasoning, non-language tasks that are nonetheless commonly evaluated such as code generation, as well as non-standard tasks such as drawing and spatial reasoning. The latter extralinguistic tasks test whether LMs are able to learn conceptual structures that  mirror the structure of the non-linguistic world, which has been suggested by recent work~\citepia{abdou-etal-2021-language,ilharco-etal-2021-probing,patel2022mapping,li2023implications,bubeck2023sparks,sogaard2023grounding}. 

We  evaluate the performance of GPT-4~\citep{gpt4}, GPT-3.5, Claude~\citep{claude}, and PaLM-2~\citep{palm2} on
tasks under both
the default and counterfactual conditions. We observe above-random counterfactual performance for most tasks, indicating some degree of task generalizability. However, the performance on counterfactual task variants consistently and substantially degrades relative to the performance on the default settings.
This suggests that these models' ability on these tasks is supported at least in part by non-transferable, default-condition-specific behaviors rather than abstract, generalizable reasoning skills.

These results also reveal several surprising relations between model behavior on default and counterfactual tasks (\S\ref{sec:analysis}), including correlations between default and counterfactual performance, varying effectiveness of zero-shot chain-of-thought prompting~\cite{kojima2023large}, and interactions between task- and instance-level frequency effects.
Overall, we find that small variations on the default instantiations of tasks are challenging for models, and thus the success of existing LMs on standard benchmarks should not be considered as sufficient evidence for their possession of full general capacity for the target task.

\section{Counterfactual Tasks} \label{sec:counterfactual-eval}

We informally conceptualize each task as a function $f_w : X \to Y $ that maps an \textbf{input} $x \in X$ under a \textbf{world model} $w \in W$ to an \textbf{output} $y \in Y$. World models encapsulate the conditions under which function evaluation takes place. For example, in Python programming, $w$ might specify assumptions of Python such as indexing and operator precedence; in arithmetic, $w$ could represent the set of conditions required for an arithmetic operation, such as the number base. 
We refer to the set of assumed default conditions, including but not limited to the base's being 10, as the \textbf{default world}, or $w^{\text{default}}$. Intuitively, for any task, $w^{\text{default}}$ corresponds to the set of conditions underlying the majority of task instances in text corpora.\footnote{This data-generating process can be described by the following generative model, $P(y \, | \, x, w)P(x \, | \, w)P(w)$. From the perspective of causal inference, our counterfactual framework can be informally seen as performing  a $\operatorname{do}$-operator on this graph \cite{pearl_2009}.}

Traditional evaluations of machine learning models assess how closely a model's learned hypothesis $h$ estimates $f_w$ by independently sampling training and test sets from the population distribution $\mathcal{D}_{f_{w}}$, and only exposing the model to the training set for learning $h$. However, 
in datasets of scraped web text, these
evaluations are subject to potential data contamination issues~\citepia{gpt3,dodge-etal-2021-documenting,magar-schwartz-2022-data}.
These issues may be more severe in recent LMs: the ever-growing pretraining datasets potentially expose the models to more evaluation instances, and the increasing sizes of recent LMs give them more ability to memorize these instances~\citep{carlini21extracting,magar-schwartz-2022-data}.

We hence consider another dimension of generalization: generalization to new task variants in \textbf{counterfactual worlds} $w^\text{cf}$, instead of new inputs $x$.
This allows us to measure the extent to which a model's $f_{w^\text{default}}$ performance is specific to $w^{\text{default}}$ or attributable to a general implementation of the task $f$.\footnote{This setup is reminiscent of intensional models of natural language semantics~(\citealp[\S12]{kratzer1998semantics}; \citealp{von2011intensional}), where $f$ is analogous to the denotation function $\sv{\cdot}$, $x$ to its input, and $y$ to its output. By default, the denotation is evaluated under the real world, extensionally, but when a different possible world is specified instead, we expect a competent system to adjust the evaluation accordingly.} 
For arithmetic, a possible $w^{\text{cf}}$ would be one that was the same as $w^{\text{default}}$ but assumed a base other than base-10. We expect a model with general arithmetic ability to perform similarly in other bases.

We emphasize that our goal is not to find counterfactual world models that are completely outside the realm of human experience. Base-9 addition, for example, is not a novel concept. Nor do we aim to guarantee that counterfactual world models are unobserved in a pretraining corpus.
Instead, counterfactuals are simply defined as variations on the \emph{default} conditions for a task. %

Concretely, we assess an LM's task performance with 0-shot prompting. We specify the task $f$, the test instance $x$, and the world model $w$ in a prompt, parse the LM's output, and compare it to the ground-truth label. We denote the LM's implementation of $f_w$ for a given instance $x$ to be, 
\begin{align*}
    h(f, w, x) = \argmax_{y'}\, P_\text{LM}\big(y' \, | \, & \operatorname{prompt}_{f}(f, x), \\[-.7em] & \operatorname{prompt}_{w}(w)\big),
\end{align*}
where the $\argmax$ is computed with an approximate decoding procedure and $\operatorname{prompt}_f$ and $\operatorname{prompt}_w$ are prompt templates that describe tasks and world models respectively.
For each task, we devise one or more $w^{\text{cf}}$ that deviate from the default world (i.e., the default task conditions).
We evaluate both $h(f, w^{\text{default}}, x)$ and $h(f, w^{\text{cf}}, x)$ via task-specific metrics. If we control $f_w(x)$ to be similarly hard between $w^{\text{default}}$ and $w^{\text{cf}}$, we can attribute the performance difference to an LM overfitting to the default instantiation of the task.

\subsection{Counterfactual Comprehension Check} \label{subsec:ccc}
One potential confounder is that an LM may be failing at a particular counterfactual task by failing to understand the prompt component that specifies the counterfactual conditions, i.e., $\operatorname{prompt}_w(w^{\text{cf}})$. That is, an LM might still be reasoning in $w^{\text{default}}$ and completely \emph{ignore} the instructions. While this would still be a failure of the LM, it does not necessarily represent a failure to perform the counterfactual task variant. We control for this by designing task-specific \textbf{counterfactual comprehension checks} (\textbf{CCC}s) that test an LM's surface understanding of the specified counterfactual world.

For each (default, counterfactual) task pair,
we introduce another control task $g_w$ with input $x'$ and output $y'$ that is much simpler than $f_w$ but still allows for the discrimination of $w^{\text{default}}$ from $w^{\text{cf}}$ (i.e., $g_{w^{\text{cf}}}(x') \ne g_{w^{\text{default}}}(x')$).
A high performance of $P_\text{LM}(y'\, |\, \operatorname{prompt}_g(g, x'), \operatorname{prompt}_w(w^{\text{cf}}))$ would indicate that  $\operatorname{prompt}_w$ is effective at making the LM perform a task in $w^{\text{cf}}$.
In the arithmetic example, for a base-9 counterfactual world, we use the same $\operatorname{prompt}_w(\texttt{base-9})$ to specify the counterfactual world, and check that it facilitates an understanding of $w=\texttt{base-9}$ by asking what the next integer after $x'$ is. If, for example, it consistently carries over digits greater than 8 and does not carry over otherwise, this would show the effectiveness of $\operatorname{prompt}_w(\texttt{base-9})$. Our CCC designs are heuristic: as with control tasks in the probing literature \citep{hewitt2019designing},
we rely on intuition to craft a $g_w$ that is ``simpler'' than $f_w$.\footnote{In this formulation, LM queries for CCC are separate from the main task queries. For some tasks, it is more natural to query about the task and CCC jointly in the same prompt, i.e., $P_{\text{LM}}(y, y' | \operatorname{prompt}_f(f, x), \operatorname{prompt}_g(g, x'), \operatorname{prompt}_w(w^{\text{cf}}))$. We use this formulation instead for those tasks.}

\section{Tasks} \label{sec:tasks}

In this section, we give a quick overview of the tasks we consider. See \S\ref{sec:full-setups} for the full description of each task and \S\ref{sec:prompts} for all the prompts used.

\subsection{Arithmetic} \label{sec:math-task}

Modern LMs have been shown to possess basic numerical reasoning abilities~\citep{lewkowycz2022solving}, with \citet{gpt3} even reporting near-perfect GPT-3 accuracy for two-digit additions. On the other hand, \citet{razeghi-etal-2022-impact} find that LMs perform significantly better on operations involving numbers that occur more frequently in the pretraining data, and \citet{li-etal-2023-bert} show that symbol replacement affects the mathematical ability of BERT~\citep{devlin2019bert}-like models; both findings point to overfitting and memorization effects. We consider the same two-digit addition task, the simplest arithmetic task in \citet{gpt3}, but inspect a model's accuracy in different bases. We use base-8, 9, 11, and 16 as the counterfactual setup which are natural generalizations to base-10 arithmetic. These bases were chosen to control for task difficulty (see \S\ref{sec:underestimation} for a discussion) and also to test for how relatively uncommon (9 \& 11) and common (8 \& 16) bases affect performance (see \S\ref{sec:analysis-world-commonness} for an analysis).
To ensure the model understands the different bases, the CCC evaluates the successor relation under each base.

\subsection{Programming} \label{sec:programming-task}

Even without explicit pretraining on large amounts of code, LMs have been found to possess  decent coding ability~\citep{gpt3}. The inclusion of large code corpora in LM pretraining~\citepia{pile,chowdhery2022palm,llama} further improves this capability in recent LMs, with  ChatGPT sometimes outperforming state-of-the-art approaches for bug fixing \citep{sobania-2023-analysis}.
Nevertheless, \citet{micelibarone2023larger} show that GPT-3 and related models are fragile under identifier swaps in programs, suggesting that these models may only possess a shallow understanding of code.
Here, we inspect an LM's programming ability through a deeper counterfactual perturbation: contrary to the traditional 0-based indexing in Python, we instruct the LM to evaluate or generate programs under a fictional language, ThonPy, that uses 1-based indexing but is otherwise identical to Python. 1-based indexing is a common assumption for other programming languages such as MATLAB and R and hence provides a fair testbed. We evaluate the LM's performance using the HumanEval dataset~\citep{chen2021codex}.
The CCC here involves the same program execution task but on much simpler inputs, such as simple list indexing, that do not involve deeper reasoning.

\subsection{Basic Syntactic Reasoning}

\citet{mahowald2023dissociating} distinguish between two types of LM capabilities: \emph{formal competence} that encompasses the knowledge of language, and \emph{functional competence} which involves using language, potentially combined with extralinguistic capacities, to interact with the world. While the other tasks we investigate in this paper assess a model's functional competence,
we also include an evaluation on formal competence. %
We revisit the attested syntactic knowledge of LMs~\citepia{yu2020word,linzen2021syntactic,ettinger2020bert,pimentel-cotterell-2021-bayesian,belinkov2022probing,lasri2022probing} by considering a meta-linguistic task~\citepia{beguvs2023large,hu2023prompt}: evaluating LMs in synthetic versions of English with  
different word orders from English's subject-verb-object (SVO) ordering. We ask the LM to identify the main subject and the main verb of a sentence under both the original and counterfactual orders, where the latter is obtained from manipulating dependency trees~\citep{ravfogel2019studying}. 
The CCC requires the model to revert simple reordered sentences to the original SVO ordering, equivalent to identifying these elements in a sentence.

\subsection{Natural Language Reasoning with First-Order Logic} \label{sec:folio-task}

We next consider a deductive reasoning task that is still based on natural language.
Logical reasoning is a prerequisite ability for many complex tasks~\citep{McCarthy_Programs59} and has been the focus of much recent work~\citepia{clark2020transformers,tafjord-etal-2021-proofwriter,saparov-mitchell-2022-towards,saparov2023language}. Nevertheless, LMs struggle with reasoning with premises that are inconsistent with common sense~\citep{dasgupta2022language,yu2023ifqa,tang2023large}. Here, we undertake a similar study from the perspective of counterfactual analysis to disentangle the effect of common sense from  a model's actual logical reasoning capability.

Following prior work, we evaluate in an entailment format and ask LMs if a series of premises entails a conclusion. We use the FOLIO dataset~\citep{han2022folio} most of whose premises are consistent with common sense, and manually rewrite them to violate common sense. We study if LM performance is affected by the truthfulness of the premises under which they operate.
The CCC directly asks the model if the original 
or post-rewrite premise is true, when presented both as options.

\subsection{Spatial Reasoning} \label{sec:spatial-task}

A major debate around LMs is whether grounded representations of meaning can be learned from form alone~\citep{bender-koller-2020-climbing, piantadosi2022meaning,mollo2023vector}. Studies have shown that LMs can learn meaningful world representations through text-only training~\citep{abdou-etal-2021-language, li2023emergent, jin2023evidence}.
In particular, \citet{patel2022mapping} find that LMs learn representations of spatial relations and cardinal directions that can be aligned to grounded conceptual spaces with few-shot demonstrations.

We similarly investigate an understanding of cardinal directions, but instead of evaluating whether a model can \emph{induce} structured conceptual spaces, we ask if it can \emph{apply}  conceptual spaces to reason about the locations of objects. Specifically, we ask an LM for the coordinates of objects whose positions are described using cardinal directions, under a conventional 2D coordinate system (e.g., where \texttt{east} corresponds to $(1, 0)$)
versus coordinate systems with swapped, rotated, and randomly permuted axes. We expect a robust representation to not be sensitive to such transformations.
The CCC involves asking the model to directly output the counterfactual cardinal directions.

\subsection{Drawing}\label{sec:drawing-task}

Despite being trained on only textual data, LMs have been shown to be able to structure their representations of perceptual concepts such as size and color~\citepia{abdou-etal-2021-language, patel2022mapping, zhang-etal-2020-language-embeddings, ilharco-etal-2021-probing}  in a way that credibly mirrors the physical world. Recent LMs can even generate plausible drawings of objects using code such  as TikZ and SVG~\cite{bubeck2023sparks,zhang2023controllable}. We evaluate the visual understanding of LMs by asking them to generate code for drawing various objects in the Processing language, which \citet{sharma2024vision} found the LMs to be more adept in. Psychological studies have shown that humans have the ability to rotate mental representations of objects~\cite{shepard1971mental, vandenberg1978mental}. For the counterfactual settings, we similarly ask the LM to generate code that draws the same object, but rotated or vertically flipped. We disallow the use of functions such as \texttt{rotate} to prevent  shortcut solutions  (see~\S\ref{sec:overestimation} for further discussion).
As with the spatial reasoning task (\S\ref{sec:spatial-task}), an ideal model should be robust to these settings.
For the CCC, we ask the model to draw a straight  line at the top of the canvas in addition to the object; a flipped/rotated line thus signifies an understanding of the transformations.

\vspace{-1mm}
\subsection{Music} \label{sec:music-task}
\vspace{-1mm}

Recent work has shown the potential of large-scale models for music infilling \citep{DBLP:journals/corr/abs-1903-07227,huang2019bach} and generation \citep{agostinelli2023musiclm, copet2023simple,ren2020popmag}. \citet{bubeck2023sparks} show that even a text-only LM with no music-specific pretraining exhibits some musical abilities, including understanding musical structure and manipulating melodies. We investigate the extent of LMs' musical abilities
through two tasks.

In the \emph{chord placement} task, we evaluate whether LMs can provide the correct chord fret placements for string instruments with standard or altered string tunings. The altered tunings, known as \emph{scordatura}, are typical in music and are used to evoke a specific sound or effect (\eg enabling heavier, deeper sound in metal music). We evaluate LMs using an existing database\footnote{\url{https://github.com/tombatossals/chords-db}} that includes chords for guitar and ukulele. In the counterfactual setting, we instruct  LMs to provide fret placements for a special guitar/ukulele where one or two of the strings are altered.
For guitar, we include drop-D tuning, a popular alternative guitar tuning that allows us to investigate whether the frequency of counterfactual tunings affects results (see \S\ref{sec:analysis-world-commonness}).
To check whether the model has understood the tunings, we ask for the first three notes on each string (including open string) as the CCC.

In the \emph{note retreival} task, we evaluate whether LMs can retrieve notes from famous melodies (\eg ``Twinkle Twinkle Little Star''). The process of re-writing melodies in different keys, referred to as ``transposition,'' is common in music (\eg to accommodate the ranges of different singers or instruments).
We evaluate LMs' musical abilities under transpositions by prompting them to retrieve the $n$-th note in a melody in either its canonical key (default setting) or a different key (counterfactual setting). 
We ask the LMs to retrieve the $n$-th note of the scale of the given key as the CCC.

\vspace{-1mm}
\subsection{Chess}
\vspace{-1mm}
Chess playing has long been regarded as a testbed for AI~\cite{DBLP:journals/corr/abs-1712-01815,DBLP:journals/corr/abs-2009-04374}, and modern LMs have exhibited abilities that imply an understanding of chess rules~\cite{srivastava2023imitation,du2023improving}.
We test this understanding by asking for the legality of a 4-move opening. In the counterfactual setting, we swap the initial positions of knights and bishops---a setup present in a real-world chess variant ``Chess 960''---and similarly ask LMs for opening legality under this new starting configuration.\footnote{A conceptually similar analysis was performed in \citet{li2023emergent} for the game of Othello.} 
We ask for the starting positions of the knights and the bishops as the CCC.

\vspace{-1mm}
\subsection{SET Game} \label{sec:set-setup}
\vspace{-1mm}
SET is a popular card game where each card has 4 attributes with 3 different values for each  attribute:
\begin{itemize}
    \item \emph{color}: (\texttt{red, blue, green})
    \item \emph{shape}: (\texttt{diamond, oval, squiggle})
    \item \emph{shading}: (\texttt{solid, shaded, open})
    \item \emph{number}: (\texttt{1, 2, 3})
\end{itemize}
In each round, a player finds a SET of 3 cards in a 12-card board whose values for each attribute are \textbf{either} \emph{all the same} \textbf{or} \emph{all unique}. This game has been thoroughly studied in computer science, from the perspective of coding theory and combinatorics~\citep{davis2003card}, linear algebra~\citep{coleman2012game}, and complexity theory~\citep{Chaudhuri2003ONTC}. We suspect this popularity makes it susceptible to overfitting by LMs and investigate this possibility.
We ask the LM to identify the card on a board that completes a 3-card SET with two given cards.
In the counterfactual setup, we invert the rule for the \emph{number} attribute, requiring its value to be mixed, in other words, \textbf{neither} \emph{all the same} \textbf{nor} \emph{all unique}.
For the CCC, we ask the model for the validity of a SET under the original rule and the counterfactual rule.

\begin{figure*}
    \centering
    \vspace{-0.6cm}
    \includegraphics[width=\textwidth]{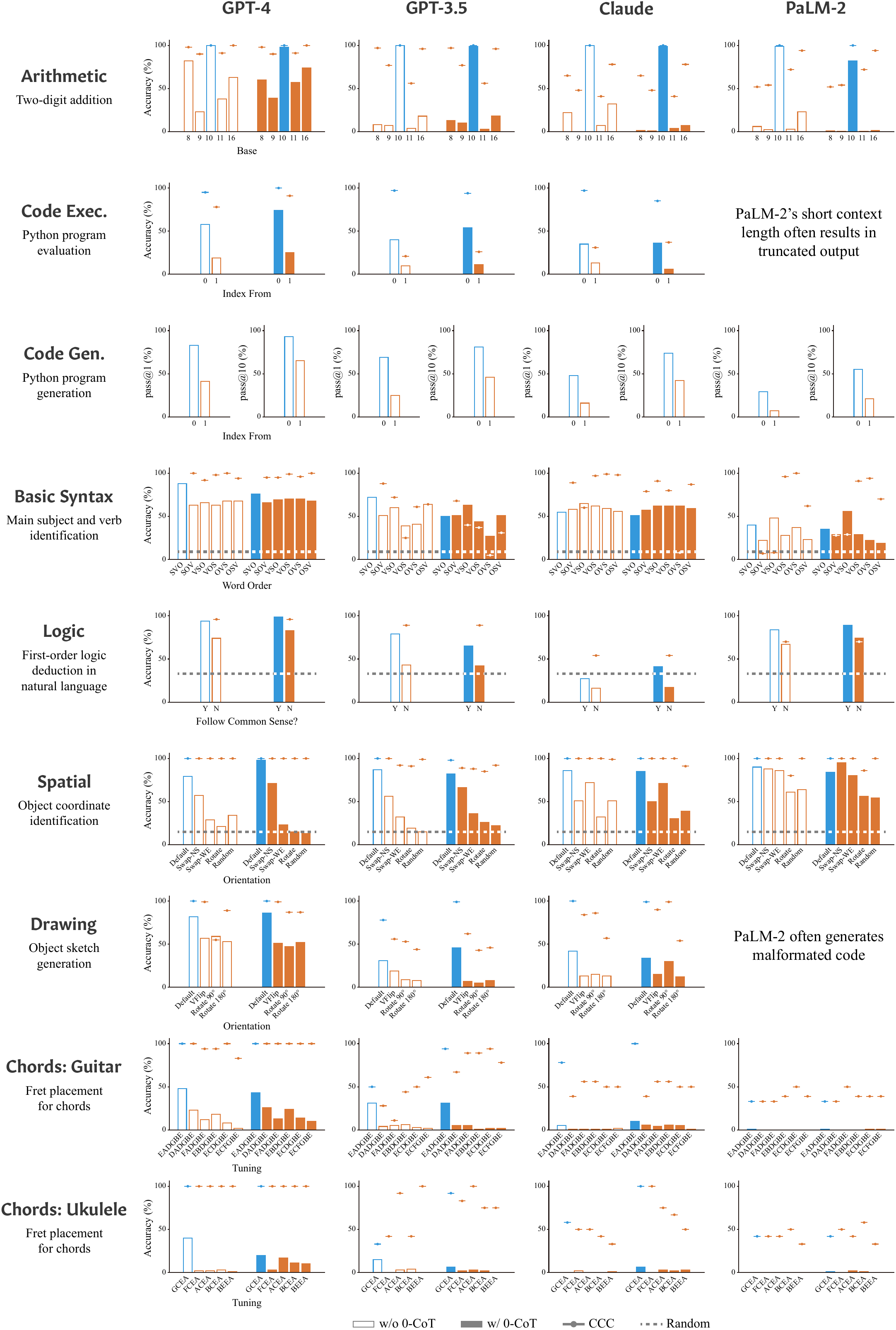}
    \caption{
    Main results. The \textcolor{real}{blue} and \textcolor{cf}{orange} bars represent the \textcolor{real}{default} and \textcolor{cf}{counterfactual} conditions respectively, either  with or without 0-shot chain-of-thought (0-CoT) (except code generation; see \S\ref{sec:programming-setup}). CCC is the counterfactual comprehension check (\S\ref{subsec:ccc}), but when applicable, we report it for the default setting too.
    Random performance is marked whenever nontrivial.
    PaLM-2 here is not the largest version (\S\ref{sec:results}).
    The CCC for code execution/generation are identical. For spatial reasoning, we average the results from all rotation degrees.
    Counterfactual performance is consistently lower than the default task performance, while CCC is usually high.
    \S\ref{sec:raw-results} reports numeric results.}
    \vspace{-1cm}
    \label{fig:results-1}
\end{figure*}

\begin{figure*}
    \centering
    \vspace{-4mm}
    \includegraphics[width=\textwidth]{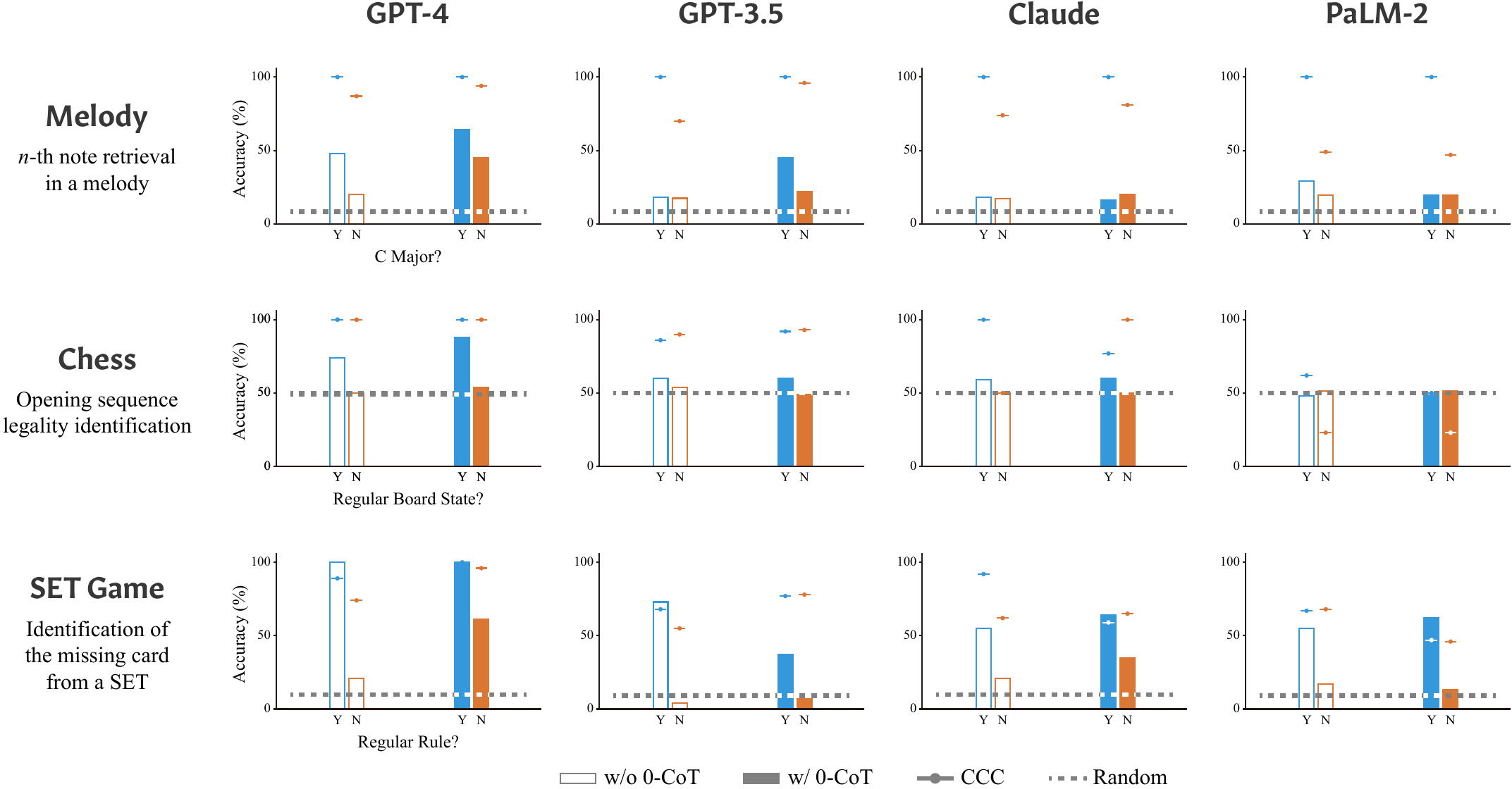}
    \caption{
    Main results (continued).
    The \textcolor{real}{blue} and \textcolor{cf}{orange} bars represent the \textcolor{real}{default} and \textcolor{cf}{counterfactual} conditions respectively, either  with or without 0-shot chain-of-thought (0-CoT). CCC is the counterfactual comprehension check (\S\ref{subsec:ccc}), but when applicable, we report it for the default setting too.
    Random performance is marked whenever nontrivial.
    PaLM-2 here is not the largest version (\S\ref{sec:results}).
    Counterfactual performance is consistently lower than the default task performance, while CCC is usually high.
    \S\ref{sec:raw-results} reports numeric results.}
    \label{fig:results-2}
\end{figure*}

\vspace{-1mm}
\section{Results} \label{sec:results}
\vspace{-1mm}
For each task, we evaluate GPT-4 (\texttt{gpt-4-0314}; \citealp{gpt4}), GPT-3.5 (\texttt{gpt-3.5-turbo-0301}), Claude (\texttt{claude-v1.3}; \citealp{claude}), and PaLM-2 (\texttt{text-bison-001}; \citealp{palm2}). As these are closed-source models, we do not have any information regarding their size, architecture, and pretaining details.\footnote{We also explored open-source models in preliminary experiments, but found that they possess unsatisfactory instruction-following ability, to the point that often their output cannot be meaningfully parsed into a prediction. We therefore do not include these models.} We note that the largest PaLM-2 model is not publicly accessible, and we can only test the second-largest version.
For each task, we experiment both with and without encouraging the model to reason step by step, by adding the phrase ``\texttt{Let's think step by step.}'' in our prompts~\citep{kojima2023large,Reynolds2021prompt}. Following \citet{kojima2023large}, we refer to this step-by-step setup as zero-shot chain-of-thought prompting (\textbf{0-CoT}; \citealp{nye2021work,wei2022chain}).
We include all  prompts in \S\ref{sec:prompts}.

Figures~\ref{fig:results-1} and \ref{fig:results-2} show our results. 
\S\ref{sec:raw-results} contains the numeric version.
We see a consistent pattern where LMs perform substantially worse on the counterfactual task variants, both with and without 0-shot CoT. For most cases, LMs exhibit an above-random counterfactual performance, suggesting some degree of the targeted ability. However, when the CCC accuracy is high, as is usually the case for GPT-4 and in select settings for other models too, the gaps in default vs. counterfactual task performance demonstrate limitations in their abstract capacity to solve the target task.
When the CCC accuracy is lower, the failure of counterfactual world comprehension would be a confounder to this conclusion, but often the gaps are so large (sometimes even dropping from near-perfect to near-zero, such as for arithmetic) that they are nonetheless strongly indicative of non-transferable, default-condition-specific implementations of the original task. The fact that the LMs sometimes cannot evaluate the CCC well under the counterfactual conditions, but can do so under the default conditions (e.g., for arithmetic, programming, drawing, etc.) itself also points to overfitting to the latter.

\section{Analysis} \label{sec:analysis}

We now investigate how a variety of factors affect the default and counterfactual performance trends that we observed in \S\ref{sec:results}.
Unless otherwise specified, we only consider GPT-4 with 0-shot CoT, which has the strongest performance in our results above.

\subsection{``Commonness'' of Counterfactual Conditions} \label{sec:analysis-world-commonness}
Our counterfactual worlds are  not designed to be completely alien to the LMs but only less common than the assumed default case.
In this sense, the counterfactual-ness of these worlds is relative, and here we take a more nuanced look at how the commonness of these counterfactual conditions affects the default-counterfactual performance gap.
For example, in the arithmetic task, all models perform better in bases 8 and 16, likely due to their relative abundance compared to  bases 9 and 11.
In spatial reasoning, the smallest counterfactual performance degradation is usually from when the north and south directions are swapped---even exceeding the default task performance for PaLM-2---potentially because some programming libraries use an inverted $y$-axis, such as matplotlib (Python), ggplot (R), and D3 (JavaScript) (see \S\ref{sec:spatial-setup}).
For chord fingering, the common alternative drop-D tuning of guitars (DADGBE) leads to the highest counterfactual performance for GPT-4.
These correlations between the counterfactual performance and the commonness of the counterfactual worlds paint a more fine-grained picture than a binary default versus counterfactual distinction and point to a memorization-like
effect where the models perform better under more common conditions.

\begin{figure*}
     \centering
     \vspace{-2mm}
     \begin{subfigure}[b]{0.24\textwidth}
         \centering
        \includegraphics[width=\textwidth]{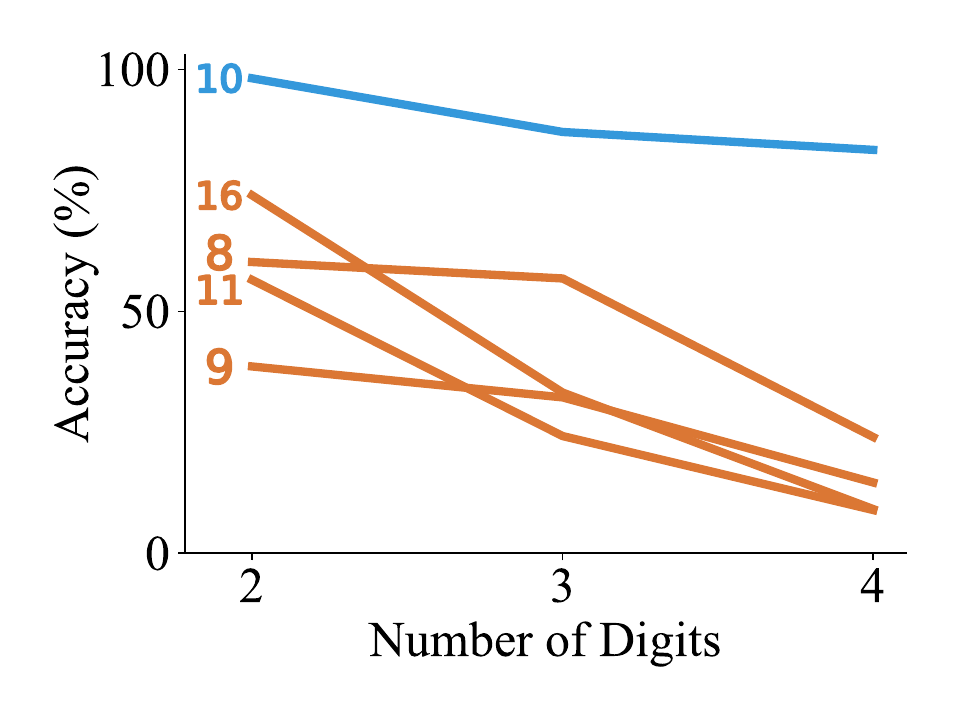}
        \caption{Addition accuracy with varying numbers of digits in the operands.}
        \label{fig:math-ndigits}
     \end{subfigure}
     \hfill
     \begin{subfigure}[b]{0.24\textwidth}
         \centering
        \includegraphics[width=\textwidth]{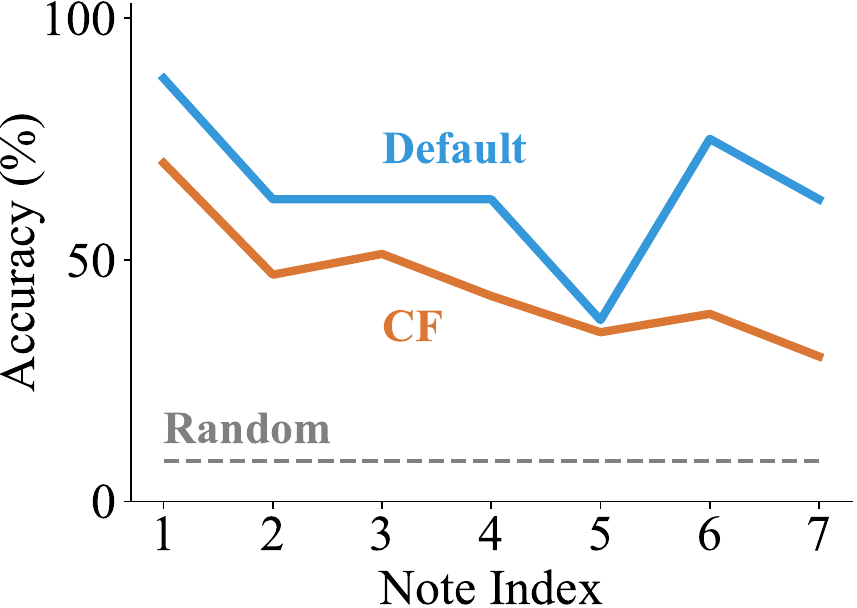}
        \caption{Melody note recall accuracy broken down by the index of the note.}
        \label{fig:songs_by_n}
     \end{subfigure}
     \hfill
     \begin{subfigure}[b]{0.24\textwidth}
         \centering
        \includegraphics[width=\textwidth]{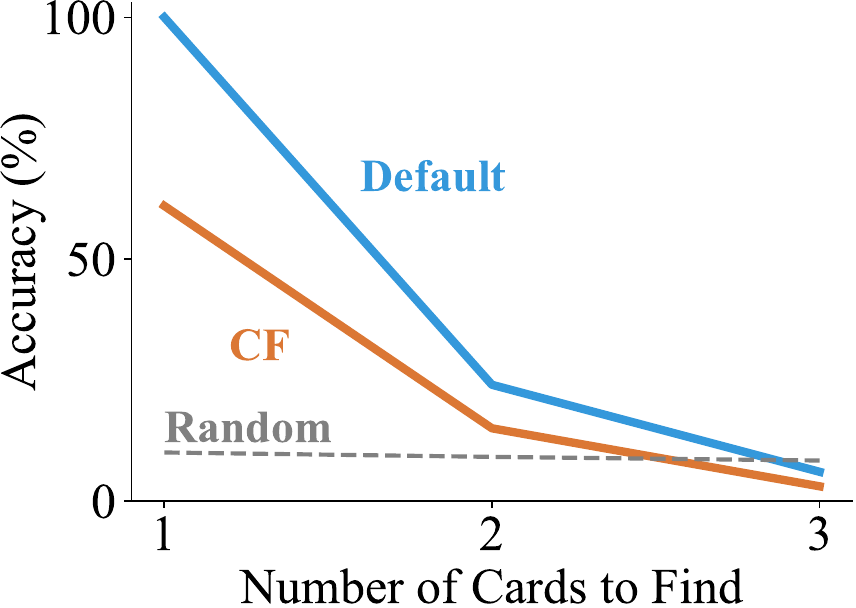}
        \caption{SET identification accuracy when needing to find different numbers of cards in a SET.}
        \label{fig:set-nhints}
     \end{subfigure}
     \hfill
     \begin{subfigure}[b]{0.24\textwidth}
         \centering
        \includegraphics[width=\textwidth]
        {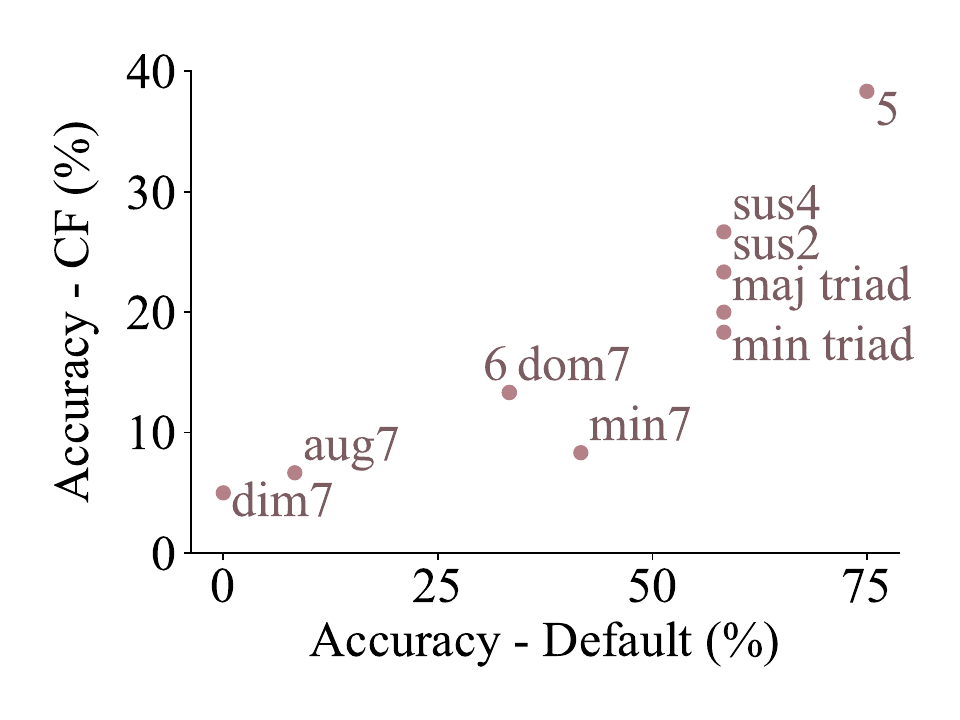}
        \caption{Fret placement accuracy by chord type. The $y$-axis averages over all altered tunings.}
        \label{fig:chords_by_type}
     \end{subfigure}
     
    \caption{Investigating the relationship between the default task performance and counterfactual performance, broken down by different factors. Only GPT-4 with 0-shot CoT results are shown. There is a consistent default-counterfactual correlation across task variants when varying different factors.}
    \vspace{-2mm}
    \label{fig:rc-relationship}
\end{figure*}

\begin{figure}[t!]
    \centering
    \includegraphics[width=0.48\textwidth]{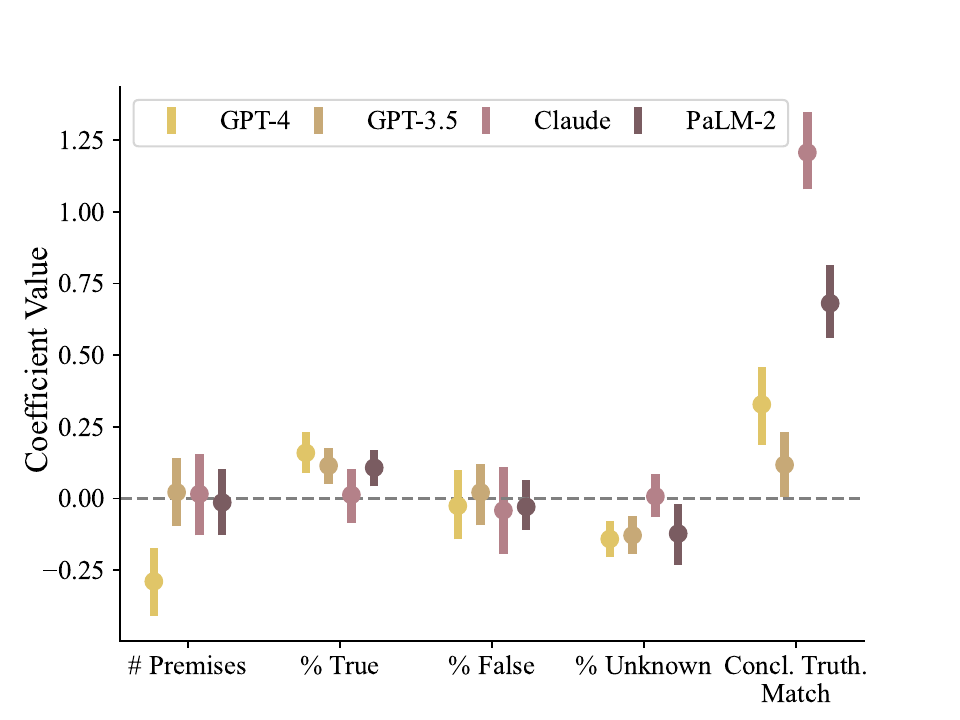}
    \caption{Logistic regression coefficients of features that predict whether an LM correctly predicts the label of an instance. ``Concl. Truth. Match'' is a binary feature that is 1 iff the instance label matches the (LM-believed) truthfulness of the conclusion. The 95\% confidence intervals are also shown. LMs tend to predict more correctly when there are more true premises, when the instance label matches the conclusion truthfulness, but less correctly with more false and unknown premises.}
    \label{fig:folio-coefs}
    \vspace{-2mm}
\end{figure}

\subsection{Proximity between Default and Counterfactual Conditions} \label{sec:analysis-world-realness}

Another axis along which the counterfactual worlds differ is in their proximity to the default conditions.
For example, for the different arithmetic bases, bases 9 and 11 are \emph{closer} to base 10, but \emph{less common} than bases 8 and 16.
While the default-counterfactual gap is most affected by commonness for the arithmetic task,
for the guitar and ukulele tunings (other than the drop-D tuning), the LM performance  generally decreases monotonically with increasing distance from the original tunings.

The FOLIO dataset~\citep{han2022folio} enables another analysis of how proximity to the default conditions affects LM performance, \emph{without} counterfactual perturbations. This dataset was constructed to mostly follow common sense, with premises and conclusions deemed true in the real world. But this is not always the case, with premises like ``John can make meals which are popular at the party,'' whose factuality cannot be determined alone. 

We evaluate how the distance between the (LM-believed) real world and the world state described by the premises (occasionally counterfactual to the LM) influences the LM's performance by training a predictive model given features approximating this distance.
For each test instance, we ask the LMs whether the premises and conclusion are true, false, or uncertain. We train a logistic regression model to predict LM correctness on each test instance, using as features the total number of premises in an input, the proportion of the premises that are true/false/uncertain, as encoded by the LM, as well as whether the LM-predicted truthfulness of the conclusion matches the label of the instance. %

Figure~\ref{fig:folio-coefs} shows the learned coefficients of these features, as well as their 95\% confidence interval bootstrapping with 1,000 iterations~\citep{EfroTibs93}.
Ideally, a robust model should predict solely based on symbolic deduction and extralinguistic truthfulness information should not affect its accuracy. In other words, these features should all have coefficients 0 and have no predictive power with respect to the model's correctness. However, all LMs predict more correctly with more realistic (true) premises, and when the conclusion's LM-predicted truthfulness matches the label (indicating a tendency to predict the label solely based on the conclusion, ignoring premises). On the other hand, they perform worse when there are more false or uncertain premises.
Most of these trends are statistically significant. 
This means that the reasoning ability of LMs is affected by the distance between the (LM-believed) real world and the world state under which the LMs are expected to reason.

Overall, these results show that LMs tend to perform better on task variants that are closer to the default instantiation of a task.

\subsection{Relationship between Default vs. Counterfactual Performance} \label{sec:rc-relationship}

Recalling our formalization $h_{\text{LM}}(f, w, x)$ in \S\ref{sec:counterfactual-eval}, the previous two subsections analyzed how  the commonness of $w$ and its proximity to $w^\text{default}$ affect the observed patterns. We now explore how the counterfactual performance correlates with the default task performance by varying the other three elements: the task $f$, the input $x$, and the LM.

We first consider different task variants with various difficulties.
For arithmetic, beyond 2-digit addition, we also measure GPT-4's 3- and 4-digit addition performance (Figure~\ref{fig:math-ndigits}).\footnote{See \citet{dziri2023faith} for a related analysis but for multiplications.}
For note retrieval from melodies, we use the index of the inquired note as the proxy for  difficulty (Figure~\ref{fig:songs_by_n}).\footnote{This is not a new task variant as compared to the setup in \S\ref{sec:music-task}, but rather a decomposition of our original results.}
For SET, while our original task shows two cards and asks a model to find the missing one from a 3-card SET, we change the task to instead show one or none of the cards in a SET, while still requiring the model to identify the SET (Figure~\ref{fig:set-nhints}).
For all these task variants, we see a strong correlation between the original and counterfactual world performance.

We also see this effect when breaking down results by test instances $x$. In Figure~\ref{fig:chords_by_type}, we separate the chord types, and observe that the default task performance correlates with the counterfactual performance. Similarly, reexamining our main results in Figures~\ref{fig:results-1} and \ref{fig:results-2}, for most tasks, stronger models under default conditions are also stronger models under counterfactual conditions, and vice versa.
Overall, these correlations mean that the default task performance can be a good indicator of its counterfactual performance, and hence we should not discount the utility of traditional evaluations.\footnote{Though these correlations are not necessarily causal.}

Furthermore, despite our evidence of LMs' overfitting to the default task conditions, 
these correlations also signify some degree of reasoning that is transferable between the default and counterfactual worlds.
This highlights that the question in our title, ``Reasoning or Reciting?'', is not a dichotomy, but rather they can co-exist in a continuum.
For example, revisiting the arithmetic results with more digits (Figure~\ref{fig:math-ndigits}), in addition to the default-counterfactual correlation, we also see an effect of memorization:
the base-10 performance decreases much more slowly than the other bases. When the input-output mappings are memorized, increased complexity would not affect the default task accuracy much; but when the counterfactual instances are not memorized, the task complexity should inversely correlate with model performance.

Occasionally, this default-counterfactual correlation trend is reversed.
In the spatial reasoning task, for example, GPT-4 achieves the best accuracy under default conditions with 0-shot CoT, but it also suffers from the largest counterfactual performance degradation. PaLM-2 performs worse under default conditions, but is the most robust to counterfactual perturbations.
An obvious possible explanation is that these models could be trained on different data, and are hence familiar with different conditions. Nevertheless, \citet{mckenzie2023inverse}, who found a similar trend but with respect to pretraining FLOPs and termed it ``inverse scaling,'' also provided a memorization-based explanation: they observed that when a task contradicts with pretraining texts, similar to how our counterfactual conditions deviate from the default conditions in pretraining, larger LMs tend to rely on the pretraining text and, in turn, fail at the contradictory task.

\subsection{0-Shot Chain-of-Thought Prompting}

Consistent with prior findings~\citepia{chen2022program,dasgupta2022language}, we generally observe 0-shot CoT to be helpful for most cases.
There are, however, exceptions. For example, 0-shot CoT substantially hurts PaLM-2's addition performance in base-10 and 16, and consistently degrades GPT-4 and GPT-3.5's chord-playing performance for the default tuning. This may be due to a model pragmatically inferring that a task is more difficult than it actually is when  explicitly asked to ``\texttt{think step by step}'', and this ``overthinking'' on simple tasks could lead to mistakes~\citep{kojima2023large}. It is also possible that these are due to memorization: the model could have memorized the specific input-output mapping of a task, without understanding how to derive the output from the input, and when explicitly instructed to spell out that process, it makes more errors~\citep{zhang2023language}.

\begin{figure}[t!]
    \centering
    \includegraphics[width=0.45\textwidth]{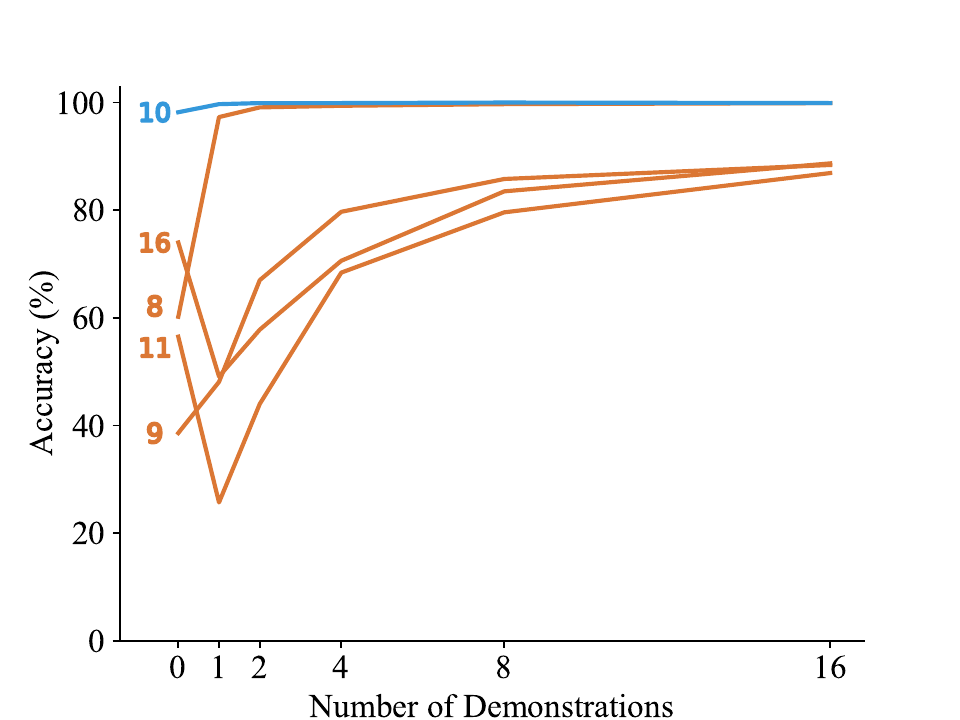}
    \caption{Two-digit addition accuracy when given different numbers of demonstration examples. The default-counterfactual gap reduces, but is not eliminated.}
    \vspace{-4mm}
    \label{fig:math-icl}
\end{figure}

\subsection{Few-shot Demonstrations} \label{sec:analysis-icl}
We study if additional demonstration examples using in-context learning~\citep{gpt3} bridges the default-counterfactual gap. For the arithmetic task, we construct few-shot CoT prompts~\citep{nye2021work,wei2022chain} and prepend up to 16 samples. As shown in Figure~\ref{fig:math-icl}, while the gap is reduced, it is still substantial for bases 9, 11, and 16. Moreover, the accuracy improvement with more demonstrations plateaus towards 16-shot, suggesting that the default-counterfactual gap  is unlikely to  be eliminated by simply adding more demonstrations (at least for arithmetic).\footnote{An interesting pattern is that bases 11 and 16 \emph{suffer} from 1-shot demonstration than 0-shot. We hypothesize that this may be due to these being the two bases with letter digits.}

\subsection{Qualitative Analysis of Drawing Results}

\begin{figure}[t!]
    \centering
    \includegraphics[width=0.48\textwidth]{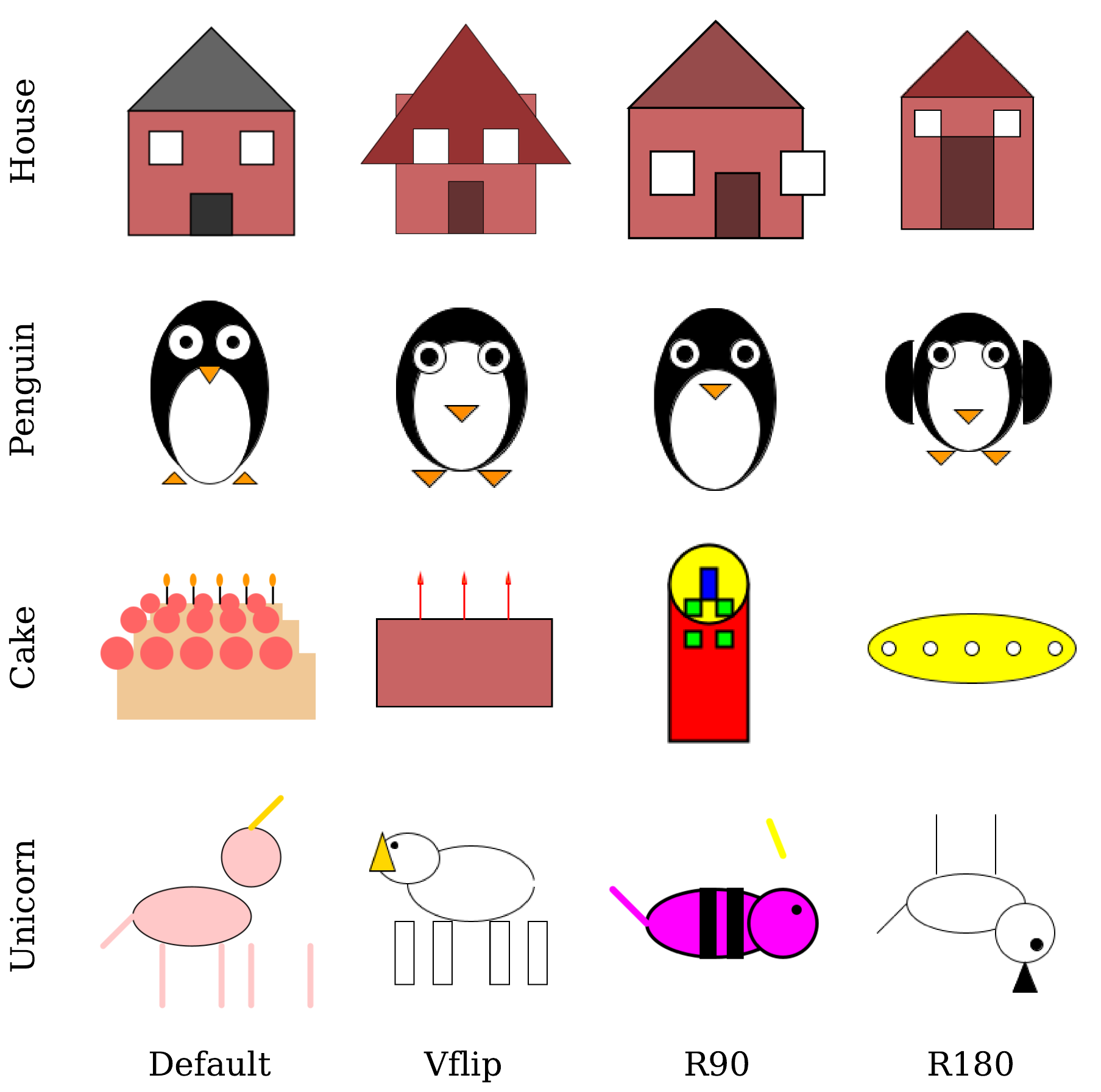}
    \caption{Visualizations of objects drawn by GPT-4 under the default (upright) and counterfactual conditions: vertical flip (Vflip, i.e. upside-down), rotates 90 degrees (R90), and 180 degrees (R180). In all cases, the CCC (not shown) passes.
    We show the original output, without flipping/rotating back as in our quantitative evaluation (\S\ref{sec:drawing-setup}).
    For the counterfactual settings, GPT-4 either does not transform the objects as instructed (e.g., house and penguin) or struggles to draw meaningful objects (e.g., cake and unicorn).}
    \vspace{-2mm}
    \label{fig:drawing_vis} 
\end{figure}

We conduct a qualitative error analysis on the drawing task and show some examples in Figure~\ref{fig:drawing_vis}.  We first note that GPT-4 successfully passes the CCC for these cases (see \S\ref{sec:drawing-task}; but not displayed here), indicating that it understands the flip/rotation instructions.  However, the objects in the counterfactual worlds are often not flipped or rotated. Even when they are transformed appropriately, the resulting drawing is often simplified or of worse quality (e.g., \texttt{Unicorn}, \texttt{Cake}). We also observed much more syntactically invalid programs in the counterfactual cases for GPT-3.5.\footnote{On average, the number of parseable programs generated by GPT-3.5 drops from 99\% in the default condition to 62\%, 71\%, and 75\% for the vertically flipped, 90\textdegree\ rotated, and 180\textdegree\ rotated settings, respectively.} These results indicate that even when a model can perform a task in the counterfactual setup, its capabilities are reduced.

\section{Discussion} \label{sec:discussion}

\textbf{Do humans also perform worse with unfamiliar counterfactual conditions?}
It is possible that humans may have lower performance under the counterfactual conditions with a fixed time budget, but not necessarily when given ample time to reason and revise. Analogous to the classic competence/performance distinction in linguistics \cite[\S1.1]{chomsky1965}, we hypothesize that humans have the \emph{competence} to generalize to new task conditions, even though it may sometimes require sufficient execution budget to realize it as robust \emph{performance}.\footnote{It is arguable if our evaluation setting provides sufficient execution budget~\citep{lampinen2023language}. Our in-context learning experiment (\S\ref{sec:analysis-icl}) may be thought of as increasing this budget, and yet the default-counterfactual gap is still sizeable there.}
In fact, there is increasing evidence from cognitive science that human reasoning is scaffolded by rich causal models of the world~\citep{pearl1988,Lake2017-LAKBMT,ullman2020bayesian,wong2023word}, and that humans can intervene on these models to perform rapid and flexible counterfactual simulations~\citep{Lagnado2013CausalRA,Gerstenberg2017EyeTrackingC,Gerstenberg2021ACS}.
However, stepping back, replicating or modeling human intelligence need not be a main goal of LMs in the first place, and human behavior is largely orthogonal to the desiderata we set for these models.

\textbf{Is task-specific reasoning bad?}
It is not necessarily bad when solving familiar tasks, but an ideal system should also possess general reasoning abilities that, when prompted, can be used to generalize to novel situations. Our point is that memorization is an often-overlooked confounding factor in interpreting LMs' reasoning abilities.

\textbf{Why do we care about counterfactual worlds? Wouldn't a model for only the default task instantiation be nonetheless useful?}
It is certainly true that such a model would  still be useful. However, many of the counterfactual worlds that we investigate are not very distant so that model performance under them still bears utility. For example, addition in different bases is certainly useful for many applications. More generally, we are necessarily interested in the counterfactual tasks themselves; we are only interested in them insofar as performance on these tasks can serve as a measurable proxy for the generalizability of these models and their underlying reasoning capabilities. %

\textbf{Aren't the observed trends trivial? The default task variant is likely the most frequent during pretraining, so of course an LM performs better under it.}
Indeed,  our results parallel the classic train-test gap in machine learning. However, an ideal learner with the right inductive biases should be able to structure their internal parameters and representations to implement general-purpose abstractions (e.g., the concept of addition), and use these abstractions to generalize to counterfactual conditions, analogous to physicists using mathematical abstractions to make predictions about universes that are substantially different from our own, or more generally to humans who can generalize to new stimuli in cognitive science studies~\citep{Lagnado2013CausalRA,Gerstenberg2017EyeTrackingC,Gerstenberg2021ACS}. Our study indicates that LMs trained on large text corpora,
remarkable as they may be, are still quite susceptible to overfitting with frequency effects.

\textbf{Can some more carefully designed prompts eliminate the default-counterfactual gap?}
This is always a possibility, and one that we can never tractably rule out. Nevertheless, given the consistency of the gap across our tasks (which use different prompts) and the 0-shot CoT setting, we believe that a prompt that \emph{completely} bridges the default-counterfactual gap is unlikely.   Our in-context learning experiment (\S\ref{sec:analysis-icl}) further shows that while this gap could be reduced by more informative prompts,  it is not fully removed. It would be interesting to apply more advanced prompting techniques~\citepia{wang2022self,wang-etal-2022-iteratively,yao2023tree,sordoni2023deep} to our counterfactual tasks. We considered 0-shot chain-of-thought in this work, which did not fully bridge the default-counterfactual gap, but we leave the exploration of these more recent prompting techniques to future work.

\section{Limitations} \label{sec:limitations}

Despite our attempt to devise novel counterfactual conditions to gauge an LM's ``true'' reasoning ability, it may not be precisely reflected by the counterfactual performance due to several factors.

\subsection{Underestimation} \label{sec:underestimation}

For our main evaluations, we aim to construct counterfactual tasks that have the same difficulty as the default variants so that task difficulty does not confound our comparisons. This is not always possible---in fact, an objective difficulty measure may not even exist. One could, for example, argue that base-11 addition is harder than base-10 because it requires reasoning with one additional digit, or base-9 is harder than base-10 because on average the sums would consist of more digits.

Retrieving notes in melodies in different keys faces a similar issue.
We expect similar retrieval difficulty under different keys
if the model recalls a melody as a series of abstract relations in a scale and directly maps them onto notes in a target key.
However, an alternative strategy would be to first retrieve the note in a canonical key and then \emph{transpose} it to the desired uncommon key. This 2-step process is a natural one that is often employed by musicians. And with this strategy, the counterfactual task consists of 2 steps and is harder than (and requires first) completing the 1-step original task. The counterfactual setup thus introduces a confounder: low performance may be driven by the increased difficulty of the counterfactual task, rather than overfitting to melodies in their canonical keys, \emph{if} models are employing two-step strategy. However, since both strategies are available to models and we do not prompt them to use a particular one, reliance on this two-step strategy may itself be indicative of overfitting to the original canonical keys.

\subsection{Overestimation} \label{sec:overestimation}

We can never be certain of how rarely particular counterfactual conditions are encountered during pretraining. It is quite likely that there is text online that, for example, draws rotated versions of various objects used in our study. Consequently, the effect of overfitting could also manifest in our counterfactual conditions, and the default-counterfactual gap could actually be larger for some genuinely unseen conditions.

We also distinguish between two types of counterfactual perturbations. One type fundamentally affects the operation of the world model and necessitates an understanding of the counterfactual world to perform the task in it (e.g., arithmetic base or 1-based indexing\footnote{It may be tempting to consider a simple replacement strategy \texttt{[i]}\textrightarrow\texttt{[i-1]} to map back to 0-based indexing. But this does not work for the dictionary type. There are other complications; see Table~\ref{tab:prompts-programming}.}). On the other hand, some perturbations are more superficial and may admit a shortcut where the model first figures out a simple mapping of the input back to the default conditions and performs the task (potentially leveraging instance-level memorization) under those.
In some of our tasks, this mapping may be simple, such as the word replacements in the natural language logical reasoning task\footnote{To be more concrete, imagine that a model memorizes an instance with nine premises on dogs involving complex logical relationships, and that it entails a given conclusion. For the counterfactual instance, we replace the word ``dogs'' with another object, say ``headphones,'' to make the premises no longer factually true. Instead of performing the reasoning over premises with headphones such as how they are, counterfactually, the cutest creatures, a model could identify the mapping ``dogs'' $\to$ ``headphones'', revert it (i.e., replace all ``headphones'' back to ``dogs''), and perform the task under the default common-sense-complying conditions.} (\S\ref{sec:folio-task}) and the transformation functions for the drawing task (\S\ref{sec:drawing-task}), which could potentially be exploited by the models. We explicitly disallow this in our prompt for the drawing task (Table~\ref{tab:prompts-drawing}) but did not identify a good way to forbid this for logical reasoning, potentially accounting for its generally high counterfactual performance.

Finally, we reiterate from \S\ref{sec:results} that a non-perfect CCC accuracy does not allow us to perfectly tease apart counterfactual performance and a failure of counterfactual condition comprehension. But often the default-counterfactual gap is so prominent that it is still strongly suggestive of overfitting to the default conditions. Also, recall from \S\ref{sec:counterfactual-eval} that the CCC itself is also a nontrivial task. For ThonPy, for example, the CCC also involves program evaluation, albeit with simpler statements that involve less reasoning, such as \texttt{print("qrstu"[4])}. We do not see an easy way to introduce ThonPy CCC that is entirely disentangled from program evaluation. This conflation would result in the CCC accuracy's being lower than what would reflect the model's understanding of the counterfactual conditions.

\section{Related Work}

\paragraph{Evaluating Conceptual Structures in LMs.}
Much prior work has investigated the extent to which LMs acquire a grounded understanding of the world through text-only training~\citepia{piantadosi2022meaning,zhang-etal-2020-language-embeddings, ilharco-etal-2021-probing, li-etal-2021-implicit}. These studies have generally found that conceptual structures of certain concepts (e.g., color, size)  often plausibly mirror those of the grounded world~\citep{abdou-etal-2021-language, patel2022mapping,mollo2023vector}. As in our study, these studies are a test of  generalization---such structures would not manifest if the concepts were memorized in a one-hot-like manner. But our evaluation differs in that it targets the \emph{reasoning} process instead of the generalization to new concepts or conceptual structures~\citep{kondo2023probing}. While prior work identified that the latter is embedded in LMs, we found that they do not fully learn the former.

\paragraph{Causal Analysis.}
Our counterfactual perturbations can be informally viewed as interventions under a causal inference framework~\citep{pearl_2009}. This relationship has been explored in machine learning and NLP for commonsense reasoning~\citep{kıcıman2023causal}, interpretability~\citep{elazar-etal-2021-amnesic,geiger2021causal,geiger22a}, spurious correlation detection~\citep{veitch2021counterfactual,eisenstein-2022-informativeness}, fairness~\citep{kusner2017counterfactual,nabi18fair}, etc.
Under this perspective, the failure  of generalization to counterfactual worlds that we observe in LMs can be viewed as a failure to robustly learn the causal effects of world states on our evaluated tasks.

\paragraph{Counterfactual Evaluation.}
``Counterfactuals'' is an informally-used term in NLP and has been used to refer to different types of perturbations. One line of work concerns counterfactuals to a certain event or situation that is still licensed in a default world model~\citepia{qin-etal-2019-counterfactual,qin-etal-2020-back,yang-etal-2020-semeval,frohberg-binder-2022-crass}, in contrast to our counterfactual world states that deviate from the default. 
\citet{qin-etal-2019-counterfactual} and \citet{frohberg-binder-2022-crass} found that GPT-3 and earlier models struggle with consistently reasoning under this type of counterfactual conditions, while \citet{kıcıman2023causal} observed more recent LMs to achieve higher counterfactual reasoning accuracy.
Another body of work examines the robustness of model predictions using counterfactual data~\citep{Kaushik2020Learning,kaushik2021explaining,gardner-etal-2020-evaluating}.
More similar to our study, \citet{li2023counterfactual} showed that while the LMs they investigated seem to be able to perform some reasoning in counterfactual worlds, this is largely affected by superficial lexical cues. Our results reveal that more recent LMs still exhibit such difficulties.

\section{Conclusion}
Through our counterfactual evaluation on 11 tasks, we identified consistent and substantial degradation of LM performance under counterfactual conditions. We attribute this gap to overfitting to the default task variants, and thus encourage future LM analyses to explicitly consider abstract task ability as detached from observed task performance, especially when these evaluated task variants might exist in abundance in the LM pretraining corpora.
Furthermore, insofar as this degradation is a result of the LMs' being trained only on surface form text, it would also be interesting future work to see if more grounded LMs (grounded in the ``real'' world, or some semantic representation, etc.) are more robust to task variations.

\iftaclpubformat
\section*{Acknowledgments}
We thank, alphabetically, Alex Gu, Alisa Liu, Belinda Li, Chenghao Yang, Han Guo, Hao Peng, Heyun Li, Jesse Dodge, Pratyusha Sharma, Tiwa Eisape, and Yizhong Wang for helpful discussions and feedback for this work. We are also grateful to  Simeng Han for providing us with an updated version of the FOLIO dataset.
Our drawing evaluation would not have been possible without our annotators Alex Hu, Ananya Harsh Jha, Belinda Li, Erjia Cao, Ha-na Park, Huirong Wen, Jiangjie Chen, Kabir Swain, Ka Wai Chan, Lucy Li, Simran Swain, Tejas Srinivasan, Tianyu Liu, Yue Bai, Yutaro Yamada, and Ziwei Wei. 
Zhaofeng would like to thank Jiamin Zhang for the guitar lessons, which were short but helpful for the relevant components of this paper.
Figure~\ref{fig:overview} uses icons from \url{flaticon.com}. This study was supported by funds from the MIT--IBM Watson AI Lab, the MIT Quest for Intelligence, and the National Science Foundation under grants IIS-2212310 and IIS-2238240.
\fi

\bibliography{custom}

\begin{thebibliography}{128}
\expandafter\ifx\csname natexlab\endcsname\relax\def\natexlab#1{#1}\fi

\bibitem[{Abdou et~al.(2021)Abdou, Kulmizev, Hershcovich, Frank, Pavlick, and
  S{\o}gaard}]{abdou-etal-2021-language}
Mostafa Abdou, Artur Kulmizev, Daniel Hershcovich, Stella Frank, Ellie Pavlick,
  and Anders S{\o}gaard. 2021.
\newblock \href {https://doi.org/10.18653/v1/2021.conll-1.9} {Can language
  models encode perceptual structure without grounding? a case study in color}.
\newblock In \emph{Proceedings of the 25th Conference on Computational Natural
  Language Learning}, pages 109--132, Online. Association for Computational
  Linguistics.

\bibitem[{Agostinelli et~al.(2023)Agostinelli, Denk, Borsos, Engel, Verzetti,
  Caillon, Huang, Jansen, Roberts, Tagliasacchi, Sharifi, Zeghidour, and
  Frank}]{agostinelli2023musiclm}
Andrea Agostinelli, Timo~I. Denk, Zalán Borsos, Jesse Engel, Mauro Verzetti,
  Antoine Caillon, Qingqing Huang, Aren Jansen, Adam Roberts, Marco
  Tagliasacchi, Matt Sharifi, Neil Zeghidour, and Christian Frank. 2023.
\newblock \href {http://arxiv.org/abs/2301.11325} {{MusicLM}: Generating music
  from text}.

\bibitem[{Anil et~al.(2023)Anil, Dai, Firat, Johnson, Lepikhin, Passos,
  Shakeri, Taropa, Bailey, Chen, Chu, Clark, Shafey, Huang, Meier-Hellstern,
  Mishra, Moreira, Omernick, Robinson, Ruder, Tay, Xiao, Xu, Zhang, Abrego,
  Ahn, Austin, Barham, Botha, Bradbury, Brahma, Brooks, Catasta, Cheng, Cherry,
  Choquette-Choo, Chowdhery, Crepy, Dave, Dehghani, Dev, Devlin, Díaz, Du,
  Dyer, Feinberg, Feng, Fienber, Freitag, Garcia, Gehrmann, Gonzalez, Gur-Ari,
  Hand, Hashemi, Hou, Howland, Hu, Hui, Hurwitz, Isard, Ittycheriah, Jagielski,
  Jia, Kenealy, Krikun, Kudugunta, Lan, Lee, Lee, Li, Li, Li, Li, Li, Lim, Lin,
  Liu, Liu, Maggioni, Mahendru, Maynez, Misra, Moussalem, Nado, Nham, Ni,
  Nystrom, Parrish, Pellat, Polacek, Polozov, Pope, Qiao, Reif, Richter, Riley,
  Ros, Roy, Saeta, Samuel, Shelby, Slone, Smilkov, So, Sohn, Tokumine, Valter,
  Vasudevan, Vodrahalli, Wang, Wang, Wang, Wang, Wieting, Wu, Xu, Xu, Xue, Yin,
  Yu, Zhang, Zheng, Zheng, Zhou, Zhou, Petrov, and Wu}]{palm2}
Rohan Anil, Andrew~M. Dai, Orhan Firat, Melvin Johnson, Dmitry Lepikhin,
  Alexandre Passos, Siamak Shakeri, Emanuel Taropa, Paige Bailey, Zhifeng Chen,
  Eric Chu, Jonathan~H. Clark, Laurent~El Shafey, Yanping Huang, Kathy
  Meier-Hellstern, Gaurav Mishra, Erica Moreira, Mark Omernick, Kevin Robinson,
  Sebastian Ruder, Yi~Tay, Kefan Xiao, Yuanzhong Xu, Yujing Zhang,
  Gustavo~Hernandez Abrego, Junwhan Ahn, Jacob Austin, Paul Barham, Jan Botha,
  James Bradbury, Siddhartha Brahma, Kevin Brooks, Michele Catasta, Yong Cheng,
  Colin Cherry, Christopher~A. Choquette-Choo, Aakanksha Chowdhery, Clément
  Crepy, Shachi Dave, Mostafa Dehghani, Sunipa Dev, Jacob Devlin, Mark Díaz,
  Nan Du, Ethan Dyer, Vlad Feinberg, Fangxiaoyu Feng, Vlad Fienber, Markus
  Freitag, Xavier Garcia, Sebastian Gehrmann, Lucas Gonzalez, Guy Gur-Ari,
  Steven Hand, Hadi Hashemi, Le~Hou, Joshua Howland, Andrea Hu, Jeffrey Hui,
  Jeremy Hurwitz, Michael Isard, Abe Ittycheriah, Matthew Jagielski, Wenhao
  Jia, Kathleen Kenealy, Maxim Krikun, Sneha Kudugunta, Chang Lan, Katherine
  Lee, Benjamin Lee, Eric Li, Music Li, Wei Li, YaGuang Li, Jian Li, Hyeontaek
  Lim, Hanzhao Lin, Zhongtao Liu, Frederick Liu, Marcello Maggioni, Aroma
  Mahendru, Joshua Maynez, Vedant Misra, Maysam Moussalem, Zachary Nado, John
  Nham, Eric Ni, Andrew Nystrom, Alicia Parrish, Marie Pellat, Martin Polacek,
  Alex Polozov, Reiner Pope, Siyuan Qiao, Emily Reif, Bryan Richter, Parker
  Riley, Alex~Castro Ros, Aurko Roy, Brennan Saeta, Rajkumar Samuel, Renee
  Shelby, Ambrose Slone, Daniel Smilkov, David~R. So, Daniel Sohn, Simon
  Tokumine, Dasha Valter, Vijay Vasudevan, Kiran Vodrahalli, Xuezhi Wang,
  Pidong Wang, Zirui Wang, Tao Wang, John Wieting, Yuhuai Wu, Kelvin Xu, Yunhan
  Xu, Linting Xue, Pengcheng Yin, Jiahui Yu, Qiao Zhang, Steven Zheng,
  Ce~Zheng, Weikang Zhou, Denny Zhou, Slav Petrov, and Yonghui Wu. 2023.
\newblock \href {http://arxiv.org/abs/2305.10403} {{PaLM} 2 technical report}.

\bibitem[{Anthropic(2023)}]{claude}
Anthropic. 2023.
\newblock \href {https://www.anthropic.com/index/introducing-claude}
  {Introducing {Claude}}.

\bibitem[{Bai et~al.(2022)Bai, Jones, Ndousse, Askell, Chen, DasSarma, Drain,
  Fort, Ganguli, Henighan, Joseph, Kadavath, Kernion, Conerly, El-Showk,
  Elhage, Hatfield-Dodds, Hernandez, Hume, Johnston, Kravec, Lovitt, Nanda,
  Olsson, Amodei, Brown, Clark, McCandlish, Olah, Mann, and
  Kaplan}]{bai2022training}
Yuntao Bai, Andy Jones, Kamal Ndousse, Amanda Askell, Anna Chen, Nova DasSarma,
  Dawn Drain, Stanislav Fort, Deep Ganguli, Tom Henighan, Nicholas Joseph,
  Saurav Kadavath, Jackson Kernion, Tom Conerly, Sheer El-Showk, Nelson Elhage,
  Zac Hatfield-Dodds, Danny Hernandez, Tristan Hume, Scott Johnston, Shauna
  Kravec, Liane Lovitt, Neel Nanda, Catherine Olsson, Dario Amodei, Tom Brown,
  Jack Clark, Sam McCandlish, Chris Olah, Ben Mann, and Jared Kaplan. 2022.
\newblock \href {http://arxiv.org/abs/2204.05862} {Training a helpful and
  harmless assistant with reinforcement learning from human feedback}.

\bibitem[{Begu{\v{s}} et~al.(2023)Begu{\v{s}}, D{\k{a}}bkowski, and
  Rhodes}]{beguvs2023large}
Ga{\v{s}}per Begu{\v{s}}, Maksymilian D{\k{a}}bkowski, and Ryan Rhodes. 2023.
\newblock \href {https://arxiv.org/abs/2305.00948} {Large linguistic models:
  Analyzing theoretical linguistic abilities of {LLMs}}.
\newblock \emph{ArXiv preprint}, abs/2305.00948.

\bibitem[{Belinkov(2022)}]{belinkov2022probing}
Yonatan Belinkov. 2022.
\newblock \href {https://doi.org/10.1162/coli_a_00422} {Probing classifiers:
  Promises, shortcomings, and advances}.
\newblock \emph{Computational Linguistics}, 48(1):207--219.

\bibitem[{Bender and Koller(2020)}]{bender-koller-2020-climbing}
Emily~M. Bender and Alexander Koller. 2020.
\newblock \href {https://doi.org/10.18653/v1/2020.acl-main.463} {Climbing
  towards {NLU}: {On} meaning, form, and understanding in the age of data}.
\newblock In \emph{Proceedings of the 58th Annual Meeting of the Association
  for Computational Linguistics}, pages 5185--5198, Online. Association for
  Computational Linguistics.

\bibitem[{Brown et~al.(2020)Brown, Mann, Ryder, Subbiah, Kaplan, Dhariwal,
  Neelakantan, Shyam, Sastry, Askell, Agarwal, Herbert{-}Voss, Krueger,
  Henighan, Child, Ramesh, Ziegler, Wu, Winter, Hesse, Chen, Sigler, Litwin,
  Gray, Chess, Clark, Berner, McCandlish, Radford, Sutskever, and
  Amodei}]{gpt3}
Tom~B. Brown, Benjamin Mann, Nick Ryder, Melanie Subbiah, Jared Kaplan,
  Prafulla Dhariwal, Arvind Neelakantan, Pranav Shyam, Girish Sastry, Amanda
  Askell, Sandhini Agarwal, Ariel Herbert{-}Voss, Gretchen Krueger, Tom
  Henighan, Rewon Child, Aditya Ramesh, Daniel~M. Ziegler, Jeffrey Wu, Clemens
  Winter, Christopher Hesse, Mark Chen, Eric Sigler, Mateusz Litwin, Scott
  Gray, Benjamin Chess, Jack Clark, Christopher Berner, Sam McCandlish, Alec
  Radford, Ilya Sutskever, and Dario Amodei. 2020.
\newblock \href
  {https://proceedings.neurips.cc/paper/2020/hash/1457c0d6bfcb4967418bfb8ac142f64a-Abstract.html}
  {Language models are few-shot learners}.
\newblock In \emph{Advances in Neural Information Processing Systems 33: Annual
  Conference on Neural Information Processing Systems 2020, NeurIPS 2020,
  December 6-12, 2020, virtual}.

\bibitem[{Bubeck et~al.(2023)Bubeck, Chandrasekaran, Eldan, Gehrke, Horvitz,
  Kamar, Lee, Lee, Li, Lundberg, Nori, Palangi, Ribeiro, and
  Zhang}]{bubeck2023sparks}
Sébastien Bubeck, Varun Chandrasekaran, Ronen Eldan, Johannes Gehrke, Eric
  Horvitz, Ece Kamar, Peter Lee, Yin~Tat Lee, Yuanzhi Li, Scott Lundberg,
  Harsha Nori, Hamid Palangi, Marco~Tulio Ribeiro, and Yi~Zhang. 2023.
\newblock \href {http://arxiv.org/abs/2303.12712} {Sparks of artificial general
  intelligence: Early experiments with {GPT-4}}.

\bibitem[{Carlini et~al.(2020)Carlini, Tramer, Wallace, Jagielski,
  Herbert-Voss, Lee, Roberts, Brown, Song, Erlingsson, Oprea, and
  Raffel}]{carlini21extracting}
Nicholas Carlini, Florian Tramer, Eric Wallace, Matthew Jagielski, Ariel
  Herbert-Voss, Katherine Lee, Adam Roberts, Tom Brown, Dawn Song, Ulfar
  Erlingsson, Alina Oprea, and Colin Raffel. 2020.
\newblock \href {https://arxiv.org/abs/2012.07805} {Extracting training data
  from large language models}.
\newblock \emph{ArXiv preprint}, abs/2012.07805.

\bibitem[{Chaudhuri et~al.(2003)Chaudhuri, Godfrey, and
  Ratajczak}]{Chaudhuri2003ONTC}
Kamalika Chaudhuri, Brighten Godfrey, and David Ratajczak. 2003.
\newblock On the complexity of the game of {SET}.

\bibitem[{Chen et~al.(2021)Chen, Tworek, Jun, Yuan, de~Oliveira~Pinto, Kaplan,
  Edwards, Burda, Joseph, Brockman, Ray, Puri, Krueger, Petrov, Khlaaf, Sastry,
  Mishkin, Chan, Gray, Ryder, Pavlov, Power, Kaiser, Bavarian, Winter, Tillet,
  Such, Cummings, Plappert, Chantzis, Barnes, Herbert-Voss, Guss, Nichol,
  Paino, Tezak, Tang, Babuschkin, Balaji, Jain, Saunders, Hesse, Carr, Leike,
  Achiam, Misra, Morikawa, Radford, Knight, Brundage, Murati, Mayer, Welinder,
  McGrew, Amodei, McCandlish, Sutskever, and Zaremba}]{chen2021codex}
Mark Chen, Jerry Tworek, Heewoo Jun, Qiming Yuan, Henrique~Ponde
  de~Oliveira~Pinto, Jared Kaplan, Harri Edwards, Yuri Burda, Nicholas Joseph,
  Greg Brockman, Alex Ray, Raul Puri, Gretchen Krueger, Michael Petrov, Heidy
  Khlaaf, Girish Sastry, Pamela Mishkin, Brooke Chan, Scott Gray, Nick Ryder,
  Mikhail Pavlov, Alethea Power, Lukasz Kaiser, Mohammad Bavarian, Clemens
  Winter, Philippe Tillet, Felipe~Petroski Such, Dave Cummings, Matthias
  Plappert, Fotios Chantzis, Elizabeth Barnes, Ariel Herbert-Voss,
  William~Hebgen Guss, Alex Nichol, Alex Paino, Nikolas Tezak, Jie Tang, Igor
  Babuschkin, Suchir Balaji, Shantanu Jain, William Saunders, Christopher
  Hesse, Andrew~N. Carr, Jan Leike, Josh Achiam, Vedant Misra, Evan Morikawa,
  Alec Radford, Matthew Knight, Miles Brundage, Mira Murati, Katie Mayer, Peter
  Welinder, Bob McGrew, Dario Amodei, Sam McCandlish, Ilya Sutskever, and
  Wojciech Zaremba. 2021.
\newblock \href {http://arxiv.org/abs/2107.03374} {Evaluating large language
  models trained on code}.

\bibitem[{Chen et~al.(2022)Chen, Ma, Wang, and Cohen}]{chen2022program}
Wenhu Chen, Xueguang Ma, Xinyi Wang, and William~W Cohen. 2022.
\newblock \href {https://arxiv.org/abs/2211.12588} {Program of thoughts
  prompting: Disentangling computation from reasoning for numerical reasoning
  tasks}.
\newblock \emph{ArXiv preprint}, abs/2211.12588.

\bibitem[{Chomsky(1965)}]{chomsky1965}
Noam Chomsky. 1965.
\newblock \emph{Aspects of the Theory of Syntax}.
\newblock The MIT Press, Cambridge.

\bibitem[{Chowdhery et~al.(2022)Chowdhery, Narang, Devlin, Bosma, Mishra,
  Roberts, Barham, Chung, Sutton, Gehrmann, Schuh, Shi, Tsvyashchenko, Maynez,
  Rao, Barnes, Tay, Shazeer, Prabhakaran, Reif, Du, Hutchinson, Pope, Bradbury,
  Austin, Isard, Gur-Ari, Yin, Duke, Levskaya, Ghemawat, Dev, Michalewski,
  Garcia, Misra, Robinson, Fedus, Zhou, Ippolito, Luan, Lim, Zoph, Spiridonov,
  Sepassi, Dohan, Agrawal, Omernick, Dai, Pillai, Pellat, Lewkowycz, Moreira,
  Child, Polozov, Lee, Zhou, Wang, Saeta, Diaz, Firat, Catasta, Wei,
  Meier-Hellstern, Eck, Dean, Petrov, and Fiedel}]{chowdhery2022palm}
Aakanksha Chowdhery, Sharan Narang, Jacob Devlin, Maarten Bosma, Gaurav Mishra,
  Adam Roberts, Paul Barham, Hyung~Won Chung, Charles Sutton, Sebastian
  Gehrmann, Parker Schuh, Kensen Shi, Sasha Tsvyashchenko, Joshua Maynez,
  Abhishek Rao, Parker Barnes, Yi~Tay, Noam Shazeer, Vinodkumar Prabhakaran,
  Emily Reif, Nan Du, Ben Hutchinson, Reiner Pope, James Bradbury, Jacob
  Austin, Michael Isard, Guy Gur-Ari, Pengcheng Yin, Toju Duke, Anselm
  Levskaya, Sanjay Ghemawat, Sunipa Dev, Henryk Michalewski, Xavier Garcia,
  Vedant Misra, Kevin Robinson, Liam Fedus, Denny Zhou, Daphne Ippolito, David
  Luan, Hyeontaek Lim, Barret Zoph, Alexander Spiridonov, Ryan Sepassi, David
  Dohan, Shivani Agrawal, Mark Omernick, Andrew~M. Dai,
  Thanumalayan~Sankaranarayana Pillai, Marie Pellat, Aitor Lewkowycz, Erica
  Moreira, Rewon Child, Oleksandr Polozov, Katherine Lee, Zongwei Zhou, Xuezhi
  Wang, Brennan Saeta, Mark Diaz, Orhan Firat, Michele Catasta, Jason Wei,
  Kathy Meier-Hellstern, Douglas Eck, Jeff Dean, Slav Petrov, and Noah Fiedel.
  2022.
\newblock \href {http://arxiv.org/abs/2204.02311} {{PaLM}: Scaling language
  modeling with pathways}.

\bibitem[{Clark et~al.(2020)Clark, Tafjord, and
  Richardson}]{clark2020transformers}
Peter Clark, Oyvind Tafjord, and Kyle Richardson. 2020.
\newblock \href {https://doi.org/10.24963/ijcai.2020/537} {Transformers as soft
  reasoners over language}.
\newblock In \emph{Proceedings of the Twenty-Ninth International Joint
  Conference on Artificial Intelligence, {IJCAI} 2020}, pages 3882--3890.
  ijcai.org.

\bibitem[{Coleman and Hartshorn(2012)}]{coleman2012game}
Ben Coleman and Kevin Hartshorn. 2012.
\newblock \href {https://doi.org/10.4169/math.mag.85.2.083} {Game, set, math}.
\newblock \emph{Mathematics Magazine}, 85(2):83--96.

\bibitem[{Copet et~al.(2023)Copet, Kreuk, Gat, Remez, Kant, Synnaeve, Adi, and
  Défossez}]{copet2023simple}
Jade Copet, Felix Kreuk, Itai Gat, Tal Remez, David Kant, Gabriel Synnaeve,
  Yossi Adi, and Alexandre Défossez. 2023.
\newblock \href {http://arxiv.org/abs/2306.05284} {Simple and controllable
  music generation}.

\bibitem[{Dasgupta et~al.(2022)Dasgupta, Lampinen, Chan, Creswell, Kumaran,
  McClelland, and Hill}]{dasgupta2022language}
Ishita Dasgupta, Andrew~K. Lampinen, Stephanie C.~Y. Chan, Antonia Creswell,
  Dharshan Kumaran, James~L. McClelland, and Felix Hill. 2022.
\newblock \href {http://arxiv.org/abs/2207.07051} {Language models show
  human-like content effects on reasoning}.

\bibitem[{Davis and Maclagan(2003)}]{davis2003card}
Benjamin~Lent Davis and Diane Maclagan. 2003.
\newblock \href {https://doi.org/10.1007/BF02984846} {The card game {SET}}.
\newblock \emph{The Mathematical Intelligencer}, 25:33--40.

\bibitem[{Devlin et~al.(2019)Devlin, Chang, Lee, and
  Toutanova}]{devlin2019bert}
Jacob Devlin, Ming-Wei Chang, Kenton Lee, and Kristina Toutanova. 2019.
\newblock \href {https://doi.org/10.18653/v1/N19-1423} {{BERT}: Pre-training of
  deep bidirectional transformers for language understanding}.
\newblock In \emph{Proceedings of the 2019 Conference of the North {A}merican
  Chapter of the Association for Computational Linguistics: Human Language
  Technologies, Volume 1 (Long and Short Papers)}, pages 4171--4186,
  Minneapolis, Minnesota. Association for Computational Linguistics.

\bibitem[{Dhingra et~al.(2023)Dhingra, Singh, SB, Malviya, and
  Gill}]{dhingra2023mind}
Sifatkaur Dhingra, Manmeet Singh, Vaisakh SB, Neetiraj Malviya, and
  Sukhpal~Singh Gill. 2023.
\newblock \href {http://arxiv.org/abs/2303.11436} {Mind meets machine:
  Unravelling {GPT}-4's cognitive psychology}.

\bibitem[{Dodge et~al.(2021)Dodge, Sap, Marasovi{\'c}, Agnew, Ilharco,
  Groeneveld, Mitchell, and Gardner}]{dodge-etal-2021-documenting}
Jesse Dodge, Maarten Sap, Ana Marasovi{\'c}, William Agnew, Gabriel Ilharco,
  Dirk Groeneveld, Margaret Mitchell, and Matt Gardner. 2021.
\newblock \href {https://doi.org/10.18653/v1/2021.emnlp-main.98} {Documenting
  large webtext corpora: A case study on the colossal clean crawled corpus}.
\newblock In \emph{Proceedings of the 2021 Conference on Empirical Methods in
  Natural Language Processing}, pages 1286--1305, Online and Punta Cana,
  Dominican Republic. Association for Computational Linguistics.

\bibitem[{Du et~al.(2023)Du, Li, Torralba, Tenenbaum, and
  Mordatch}]{du2023improving}
Yilun Du, Shuang Li, Antonio Torralba, Joshua~B Tenenbaum, and Igor Mordatch.
  2023.
\newblock \href {https://arxiv.org/abs/2305.14325} {Improving factuality and
  reasoning in language models through multiagent debate}.
\newblock \emph{ArXiv preprint}, abs/2305.14325.

\bibitem[{Dziri et~al.(2023)Dziri, Lu, Sclar, Li, Jiang, Lin, West,
  Bhagavatula, Bras, Hwang, Sanyal, Welleck, Ren, Ettinger, Harchaoui, and
  Choi}]{dziri2023faith}
Nouha Dziri, Ximing Lu, Melanie Sclar, Xiang~Lorraine Li, Liwei Jiang,
  Bill~Yuchen Lin, Peter West, Chandra Bhagavatula, Ronan~Le Bras, Jena~D.
  Hwang, Soumya Sanyal, Sean Welleck, Xiang Ren, Allyson Ettinger, Zaid
  Harchaoui, and Yejin Choi. 2023.
\newblock \href {http://arxiv.org/abs/2305.18654} {Faith and fate: Limits of
  transformers on compositionality}.

\bibitem[{Efron and Tibshirani(1993)}]{EfroTibs93}
Bradley Efron and Robert~J. Tibshirani. 1993.
\newblock \emph{An Introduction to the Bootstrap}.
\newblock Number~57 in Monographs on Statistics and Applied Probability.
  Chapman \& Hall/CRC, Boca Raton, Florida, USA.

\bibitem[{Eisenstein(2022)}]{eisenstein-2022-informativeness}
Jacob Eisenstein. 2022.
\newblock \href {https://doi.org/10.18653/v1/2022.naacl-main.321}
  {Informativeness and invariance: Two perspectives on spurious correlations in
  natural language}.
\newblock In \emph{Proceedings of the 2022 Conference of the North American
  Chapter of the Association for Computational Linguistics: Human Language
  Technologies}, pages 4326--4331, Seattle, United States. Association for
  Computational Linguistics.

\bibitem[{Elazar et~al.(2021)Elazar, Ravfogel, Jacovi, and
  Goldberg}]{elazar-etal-2021-amnesic}
Yanai Elazar, Shauli Ravfogel, Alon Jacovi, and Yoav Goldberg. 2021.
\newblock \href {https://doi.org/10.1162/tacl_a_00359} {Amnesic probing:
  Behavioral explanation with amnesic counterfactuals}.
\newblock \emph{Transactions of the Association for Computational Linguistics},
  9:160--175.

\bibitem[{Ettinger(2020)}]{ettinger2020bert}
Allyson Ettinger. 2020.
\newblock \href {https://doi.org/10.1162/tacl_a_00298} {What {BERT} is not:
  Lessons from a new suite of psycholinguistic diagnostics for language
  models}.
\newblock \emph{Transactions of the Association for Computational Linguistics},
  8:34--48.

\bibitem[{Frohberg and Binder(2022)}]{frohberg-binder-2022-crass}
J{\"o}rg Frohberg and Frank Binder. 2022.
\newblock \href {https://aclanthology.org/2022.lrec-1.229} {{CRASS}: A novel
  data set and benchmark to test counterfactual reasoning of large language
  models}.
\newblock In \emph{Proceedings of the Thirteenth Language Resources and
  Evaluation Conference}, pages 2126--2140, Marseille, France. European
  Language Resources Association.

\bibitem[{Gao et~al.(2021)Gao, Biderman, Black, Golding, Hoppe, Foster, Phang,
  He, Thite, Nabeshima, Presser, and Leahy}]{pile}
Leo Gao, Stella Biderman, Sid Black, Laurence Golding, Travis Hoppe, Charles
  Foster, Jason Phang, Horace He, Anish Thite, Noa Nabeshima, Shawn Presser,
  and Connor Leahy. 2021.
\newblock \href {https://arxiv.org/abs/2101.00027} {The {P}ile: An 800{GB}
  dataset of diverse text for language modeling}.
\newblock \emph{ArXiv preprint}, abs/2101.00027.

\bibitem[{Gardner et~al.(2020)Gardner, Artzi, Basmov, Berant, Bogin, Chen,
  Dasigi, Dua, Elazar, Gottumukkala, Gupta, Hajishirzi, Ilharco, Khashabi, Lin,
  Liu, Liu, Mulcaire, Ning, Singh, Smith, Subramanian, Tsarfaty, Wallace,
  Zhang, and Zhou}]{gardner-etal-2020-evaluating}
Matt Gardner, Yoav Artzi, Victoria Basmov, Jonathan Berant, Ben Bogin, Sihao
  Chen, Pradeep Dasigi, Dheeru Dua, Yanai Elazar, Ananth Gottumukkala, Nitish
  Gupta, Hannaneh Hajishirzi, Gabriel Ilharco, Daniel Khashabi, Kevin Lin,
  Jiangming Liu, Nelson~F. Liu, Phoebe Mulcaire, Qiang Ning, Sameer Singh,
  Noah~A. Smith, Sanjay Subramanian, Reut Tsarfaty, Eric Wallace, Ally Zhang,
  and Ben Zhou. 2020.
\newblock \href {https://doi.org/10.18653/v1/2020.findings-emnlp.117}
  {Evaluating models{'} local decision boundaries via contrast sets}.
\newblock In \emph{Findings of the Association for Computational Linguistics:
  EMNLP 2020}, pages 1307--1323, Online. Association for Computational
  Linguistics.

\bibitem[{Geiger et~al.(2021)Geiger, Lu, Icard, and Potts}]{geiger2021causal}
Atticus Geiger, Hanson Lu, Thomas Icard, and Christopher Potts. 2021.
\newblock \href
  {https://proceedings.neurips.cc/paper/2021/hash/4f5c422f4d49a5a807eda27434231040-Abstract.html}
  {Causal abstractions of neural networks}.
\newblock In \emph{Advances in Neural Information Processing Systems 34: Annual
  Conference on Neural Information Processing Systems 2021, NeurIPS 2021,
  December 6-14, 2021, virtual}, pages 9574--9586.

\bibitem[{Geiger et~al.(2022)Geiger, Wu, Lu, Rozner, Kreiss, Icard, Goodman,
  and Potts}]{geiger22a}
Atticus Geiger, Zhengxuan Wu, Hanson Lu, Josh Rozner, Elisa Kreiss, Thomas
  Icard, Noah~D. Goodman, and Christopher Potts. 2022.
\newblock \href {https://proceedings.mlr.press/v162/geiger22a.html} {Inducing
  causal structure for interpretable neural networks}.
\newblock In \emph{International Conference on Machine Learning, {ICML} 2022,
  17-23 July 2022, Baltimore, Maryland, {USA}}, volume 162 of \emph{Proceedings
  of Machine Learning Research}, pages 7324--7338. {PMLR}.

\bibitem[{Gerstenberg et~al.(2021)Gerstenberg, Goodman, Lagnado, and
  Tenenbaum}]{Gerstenberg2021ACS}
Tobias Gerstenberg, Noah~D. Goodman, David~A. Lagnado, and Joshua~B. Tenenbaum.
  2021.
\newblock A counterfactual simulation model of causal judgments for physical
  events.
\newblock \emph{Psychological review}.

\bibitem[{Gerstenberg et~al.(2017)Gerstenberg, Peterson, Goodman, Lagnado, and
  Tenenbaum}]{Gerstenberg2017EyeTrackingC}
Tobias Gerstenberg, Matthew Peterson, Noah~D. Goodman, David~A. Lagnado, and
  Joshua~B. Tenenbaum. 2017.
\newblock Eye-tracking causality.
\newblock \emph{Psychological Science}, 28:1731 -- 1744.

\bibitem[{Guo et~al.(2023)Guo, Zhang, Wang, Jiang, Nie, Ding, Yue, and
  Wu}]{guo2023close}
Biyang Guo, Xin Zhang, Ziyuan Wang, Minqi Jiang, Jinran Nie, Yuxuan Ding,
  Jianwei Yue, and Yupeng Wu. 2023.
\newblock \href {http://arxiv.org/abs/2301.07597} {How close is {ChatGPT} to
  human experts? {Comparison} corpus, evaluation, and detection}.

\bibitem[{Han et~al.(2022)Han, Schoelkopf, Zhao, Qi, Riddell, Benson, Sun,
  Zubova, Qiao, Burtell, Peng, Fan, Liu, Wong, Sailor, Ni, Nan, Kasai, Yu,
  Zhang, Joty, Fabbri, Kryscinski, Lin, Xiong, and Radev}]{han2022folio}
Simeng Han, Hailey Schoelkopf, Yilun Zhao, Zhenting Qi, Martin Riddell, Luke
  Benson, Lucy Sun, Ekaterina Zubova, Yujie Qiao, Matthew Burtell, David Peng,
  Jonathan Fan, Yixin Liu, Brian Wong, Malcolm Sailor, Ansong Ni, Linyong Nan,
  Jungo Kasai, Tao Yu, Rui Zhang, Shafiq Joty, Alexander~R. Fabbri, Wojciech
  Kryscinski, Xi~Victoria Lin, Caiming Xiong, and Dragomir Radev. 2022.
\newblock \href {http://arxiv.org/abs/2209.00840} {{FOLIO}: Natural language
  reasoning with first-order logic}.

\bibitem[{Heim and Kratzer(1998)}]{kratzer1998semantics}
Irene Heim and Angelika Kratzer. 1998.
\newblock \emph{Semantics in Generative Grammar}.
\newblock Blackwell.

\bibitem[{Hewitt and Liang(2019)}]{hewitt2019designing}
John Hewitt and Percy Liang. 2019.
\newblock \href {https://doi.org/10.18653/v1/D19-1275} {Designing and
  interpreting probes with control tasks}.
\newblock In \emph{Proceedings of the 2019 Conference on Empirical Methods in
  Natural Language Processing and the 9th International Joint Conference on
  Natural Language Processing (EMNLP-IJCNLP)}, pages 2733--2743, Hong Kong,
  China. Association for Computational Linguistics.

\bibitem[{Hoffmann et~al.(2022)Hoffmann, Borgeaud, Mensch, Buchatskaya, Cai,
  Rutherford, de~Las~Casas, Hendricks, Welbl, Clark, Hennigan, Noland,
  Millican, van~den Driessche, Damoc, Guy, Osindero, Simonyan, Elsen, Rae,
  Vinyals, and Sifre}]{hoffmann2022training}
Jordan Hoffmann, Sebastian Borgeaud, Arthur Mensch, Elena Buchatskaya, Trevor
  Cai, Eliza Rutherford, Diego de~Las~Casas, Lisa~Anne Hendricks, Johannes
  Welbl, Aidan Clark, Tom Hennigan, Eric Noland, Katie Millican, George van~den
  Driessche, Bogdan Damoc, Aurelia Guy, Simon Osindero, Karen Simonyan, Erich
  Elsen, Jack~W. Rae, Oriol Vinyals, and Laurent Sifre. 2022.
\newblock \href {http://arxiv.org/abs/2203.15556} {Training compute-optimal
  large language models}.

\bibitem[{Hu and Levy(2023)}]{hu2023prompt}
Jennifer Hu and Roger Levy. 2023.
\newblock \href {https://arxiv.org/abs/2305.13264} {Prompt-based methods may
  underestimate large language models' linguistic generalizations}.
\newblock \emph{ArXiv preprint}, abs/2305.13264.

\bibitem[{Huang et~al.(2019{\natexlab{a}})Huang, Cooijmans, Roberts, Courville,
  and Eck}]{DBLP:journals/corr/abs-1903-07227}
Cheng{-}Zhi~Anna Huang, Tim Cooijmans, Adam Roberts, Aaron~C. Courville, and
  Douglas Eck. 2019{\natexlab{a}}.
\newblock \href {https://arxiv.org/abs/1903.07227} {Counterpoint by
  convolution}.
\newblock \emph{ArXiv preprint}, abs/1903.07227.

\bibitem[{Huang et~al.(2019{\natexlab{b}})Huang, Hawthorne, Roberts,
  Dinculescu, Wexler, Hong, and Howcroft}]{huang2019bach}
Cheng-Zhi~Anna Huang, Curtis Hawthorne, Adam Roberts, Monica Dinculescu, James
  Wexler, Leon Hong, and Jacob Howcroft. 2019{\natexlab{b}}.
\newblock \href {http://arxiv.org/abs/1907.06637} {The bach doodle:
  Approachable music composition with machine learning at scale}.

\bibitem[{Ilharco et~al.(2021)Ilharco, Zellers, Farhadi, and
  Hajishirzi}]{ilharco-etal-2021-probing}
Gabriel Ilharco, Rowan Zellers, Ali Farhadi, and Hannaneh Hajishirzi. 2021.
\newblock \href {https://doi.org/10.18653/v1/2021.naacl-main.422} {Probing
  contextual language models for common ground with visual representations}.
\newblock In \emph{Proceedings of the 2021 Conference of the North American
  Chapter of the Association for Computational Linguistics: Human Language
  Technologies}, pages 5367--5377, Online. Association for Computational
  Linguistics.

\bibitem[{Jin and Rinard(2023)}]{jin2023evidence}
Charles Jin and Martin Rinard. 2023.
\newblock \href {https://arxiv.org/abs/2305.11169} {Evidence of meaning in
  language models trained on programs}.
\newblock \emph{ArXiv preprint}, abs/2305.11169.

\bibitem[{Kaushik et~al.(2020)Kaushik, Hovy, and Lipton}]{Kaushik2020Learning}
Divyansh Kaushik, Eduard~H. Hovy, and Zachary~Chase Lipton. 2020.
\newblock \href {https://openreview.net/forum?id=Sklgs0NFvr} {Learning the
  difference that makes a difference with counterfactually-augmented data}.
\newblock In \emph{8th International Conference on Learning Representations,
  {ICLR} 2020, Addis Ababa, Ethiopia, April 26-30, 2020}. OpenReview.net.

\bibitem[{Kaushik et~al.(2021)Kaushik, Setlur, Hovy, and
  Lipton}]{kaushik2021explaining}
Divyansh Kaushik, Amrith Setlur, Eduard~H. Hovy, and Zachary~Chase Lipton.
  2021.
\newblock \href {https://openreview.net/forum?id=HHiiQKWsOcV} {Explaining the
  efficacy of counterfactually augmented data}.
\newblock In \emph{9th International Conference on Learning Representations,
  {ICLR} 2021, Virtual Event, Austria, May 3-7, 2021}. OpenReview.net.

\bibitem[{Kojima et~al.(2023)Kojima, Gu, Reid, Matsuo, and
  Iwasawa}]{kojima2023large}
Takeshi Kojima, Shixiang~Shane Gu, Machel Reid, Yutaka Matsuo, and Yusuke
  Iwasawa. 2023.
\newblock \href {http://arxiv.org/abs/2205.11916} {Large language models are
  zero-shot reasoners}.

\bibitem[{Kondo et~al.(2023)Kondo, Sugawara, and Aizawa}]{kondo2023probing}
Kazushi Kondo, Saku Sugawara, and Akiko Aizawa. 2023.
\newblock \href {https://arxiv.org/abs/2306.02258} {Probing physical reasoning
  with counter-commonsense context}.
\newblock \emph{ArXiv preprint}, abs/2306.02258.

\bibitem[{Kung et~al.(2023)Kung, Cheatham, Medenilla, Sillos, De~Leon,
  Elepaño, Madriaga, Aggabao, Diaz-Candido, Maningo, and
  Tseng}]{kung2023performance}
Tiffany~H. Kung, Morgan Cheatham, Arielle Medenilla, Czarina Sillos, Lorie
  De~Leon, Camille Elepaño, Maria Madriaga, Rimel Aggabao, Giezel
  Diaz-Candido, James Maningo, and Victor Tseng. 2023.
\newblock \href {https://doi.org/10.1371/journal.pdig.0000198} {Performance of
  {ChatGPT} on {USMLE}: Potential for {AI}-assisted medical education using
  large language models}.
\newblock \emph{PLOS Digital Health}, 2(2):1--12.

\bibitem[{Kusner et~al.(2017)Kusner, Loftus, Russell, and
  Silva}]{kusner2017counterfactual}
Matt~J. Kusner, Joshua~R. Loftus, Chris Russell, and Ricardo Silva. 2017.
\newblock \href
  {https://proceedings.neurips.cc/paper/2017/hash/a486cd07e4ac3d270571622f4f316ec5-Abstract.html}
  {Counterfactual fairness}.
\newblock In \emph{Advances in Neural Information Processing Systems 30: Annual
  Conference on Neural Information Processing Systems 2017, December 4-9, 2017,
  Long Beach, CA, {USA}}, pages 4066--4076.

\bibitem[{Kıcıman et~al.(2023)Kıcıman, Ness, Sharma, and
  Tan}]{kıcıman2023causal}
Emre Kıcıman, Robert Ness, Amit Sharma, and Chenhao Tan. 2023.
\newblock \href {http://arxiv.org/abs/2305.00050} {Causal reasoning and large
  language models: Opening a new frontier for causality}.

\bibitem[{Lagnado et~al.(2013)Lagnado, Gerstenberg, and
  Zultan}]{Lagnado2013CausalRA}
David~A. Lagnado, Tobias Gerstenberg, and Ro’i Zultan. 2013.
\newblock Causal responsibility and counterfactuals.
\newblock \emph{Cognitive Science}, 37:1036 -- 1073.

\bibitem[{Lake et~al.(2017)Lake, Ullman, Tenenbaum, and
  Gershman}]{Lake2017-LAKBMT}
Brenden~M. Lake, Tomer~D. Ullman, Joshua~B. Tenenbaum, and Samuel~J. Gershman.
  2017.
\newblock \href {https://doi.org/10.1017/s0140525x16001837} {Building machines
  that learn and think like people}.
\newblock \emph{Behavioral and Brain Sciences}, 40.

\bibitem[{Lampinen(2023)}]{lampinen2023language}
Andrew~Kyle Lampinen. 2023.
\newblock \href {http://arxiv.org/abs/2210.15303} {Can language models handle
  recursively nested grammatical structures? {A} case study on comparing models
  and humans}.

\bibitem[{Lasri et~al.(2022)Lasri, Pimentel, Lenci, Poibeau, and
  Cotterell}]{lasri2022probing}
Karim Lasri, Tiago Pimentel, Alessandro Lenci, Thierry Poibeau, and Ryan
  Cotterell. 2022.
\newblock \href {https://doi.org/10.18653/v1/2022.acl-long.603} {Probing for
  the usage of grammatical number}.
\newblock In \emph{Proceedings of the 60th Annual Meeting of the Association
  for Computational Linguistics (Volume 1: Long Papers)}, pages 8818--8831,
  Dublin, Ireland. Association for Computational Linguistics.

\bibitem[{Lewkowycz et~al.(2022)Lewkowycz, Andreassen, Dohan, Dyer,
  Michalewski, Ramasesh, Slone, Anil, Schlag, Gutman-Solo, Wu, Neyshabur,
  Gur-Ari, and Misra}]{lewkowycz2022solving}
Aitor Lewkowycz, Anders Andreassen, David Dohan, Ethan Dyer, Henryk
  Michalewski, Vinay Ramasesh, Ambrose Slone, Cem Anil, Imanol Schlag, Theo
  Gutman-Solo, Yuhuai Wu, Behnam Neyshabur, Guy Gur-Ari, and Vedant Misra.
  2022.
\newblock \href {http://arxiv.org/abs/2206.14858} {Solving quantitative
  reasoning problems with language models}.

\bibitem[{Li et~al.(2022)Li, Yu, Khabsa, Zettlemoyer, Halevy, and
  Andreas}]{li-etal-2022-quantifying}
Belinda Li, Jane Yu, Madian Khabsa, Luke Zettlemoyer, Alon Halevy, and Jacob
  Andreas. 2022.
\newblock \href {https://doi.org/10.18653/v1/2022.naacl-main.346} {Quantifying
  adaptability in pre-trained language models with 500 tasks}.
\newblock In \emph{Proceedings of the 2022 Conference of the North American
  Chapter of the Association for Computational Linguistics: Human Language
  Technologies}, pages 4696--4715, Seattle, United States. Association for
  Computational Linguistics.

\bibitem[{Li et~al.(2021)Li, Nye, and Andreas}]{li-etal-2021-implicit}
Belinda~Z. Li, Maxwell Nye, and Jacob Andreas. 2021.
\newblock \href {https://doi.org/10.18653/v1/2021.acl-long.143} {Implicit
  representations of meaning in neural language models}.
\newblock In \emph{Proceedings of the 59th Annual Meeting of the Association
  for Computational Linguistics and the 11th International Joint Conference on
  Natural Language Processing (Volume 1: Long Papers)}, pages 1813--1827,
  Online. Association for Computational Linguistics.

\bibitem[{Li et~al.(2023{\natexlab{a}})Li, Kementchedjhieva, and
  S{\o}gaard}]{li2023implications}
Jiaang Li, Yova Kementchedjhieva, and Anders S{\o}gaard. 2023{\natexlab{a}}.
\newblock \href {https://arxiv.org/abs/2302.06555} {Implications of the
  convergence of language and vision model geometries}.
\newblock \emph{ArXiv preprint}, abs/2302.06555.

\bibitem[{Li et~al.(2023{\natexlab{b}})Li, Yu, and
  Ettinger}]{li2023counterfactual}
Jiaxuan Li, Lang Yu, and Allyson Ettinger. 2023{\natexlab{b}}.
\newblock \href {http://arxiv.org/abs/2305.16572} {Counterfactual reasoning:
  Testing language models' understanding of hypothetical scenarios}.

\bibitem[{Li et~al.(2023{\natexlab{c}})Li, Hopkins, Bau, Vi{\'e}gas, Pfister,
  and Wattenberg}]{li2023emergent}
Kenneth Li, Aspen~K Hopkins, David Bau, Fernanda Vi{\'e}gas, Hanspeter Pfister,
  and Martin Wattenberg. 2023{\natexlab{c}}.
\newblock \href {https://openreview.net/forum?id=DeG07_TcZvT} {Emergent world
  representations: Exploring a sequence model trained on a synthetic task}.
\newblock In \emph{The Eleventh International Conference on Learning
  Representations}.

\bibitem[{Li et~al.(2023{\natexlab{d}})Li, Ziser, Coavoux, and
  Cohen}]{li-etal-2023-bert}
Weixian~Waylon Li, Yftah Ziser, Maximin Coavoux, and Shay~B. Cohen.
  2023{\natexlab{d}}.
\newblock \href {https://aclanthology.org/2023.eacl-main.260} {{BERT} is not
  the count: Learning to match mathematical statements with proofs}.
\newblock In \emph{Proceedings of the 17th Conference of the European Chapter
  of the Association for Computational Linguistics}, pages 3581--3593,
  Dubrovnik, Croatia. Association for Computational Linguistics.

\bibitem[{Linzen and Baroni(2021)}]{linzen2021syntactic}
Tal Linzen and Marco Baroni. 2021.
\newblock Syntactic structure from deep learning.
\newblock \emph{Annual Review of Linguistics}, 7:195--212.

\bibitem[{Magar and Schwartz(2022)}]{magar-schwartz-2022-data}
Inbal Magar and Roy Schwartz. 2022.
\newblock \href {https://doi.org/10.18653/v1/2022.acl-short.18} {Data
  contamination: From memorization to exploitation}.
\newblock In \emph{Proceedings of the 60th Annual Meeting of the Association
  for Computational Linguistics (Volume 2: Short Papers)}, pages 157--165,
  Dublin, Ireland. Association for Computational Linguistics.

\bibitem[{Mahowald et~al.(2023)Mahowald, Ivanova, Blank, Kanwisher, Tenenbaum,
  and Fedorenko}]{mahowald2023dissociating}
Kyle Mahowald, Anna~A Ivanova, Idan~A Blank, Nancy Kanwisher, Joshua~B
  Tenenbaum, and Evelina Fedorenko. 2023.
\newblock \href {https://arxiv.org/abs/2301.06627} {Dissociating language and
  thought in large language models: a cognitive perspective}.
\newblock \emph{ArXiv preprint}, abs/2301.06627.

\bibitem[{Malinka et~al.(2023)Malinka, Perešíni, Firc, Hujňák, and
  Januš}]{malinka2023educational}
Kamil Malinka, Martin Perešíni, Anton Firc, Ondřej Hujňák, and Filip
  Januš. 2023.
\newblock \href {http://arxiv.org/abs/2303.11146} {On the educational impact of
  {ChatGPT}: {Is} artificial intelligence ready to obtain a university degree?}

\bibitem[{Marcus et~al.(1993)Marcus, Santorini, and
  Marcinkiewicz}]{marcus1993building}
Mitchell~P. Marcus, Beatrice Santorini, and Mary~Ann Marcinkiewicz. 1993.
\newblock \href {https://aclanthology.org/J93-2004} {Building a large annotated
  corpus of {E}nglish: The {P}enn {T}reebank}.
\newblock \emph{Computational Linguistics}, 19(2):313--330.

\bibitem[{McCarthy(1959)}]{McCarthy_Programs59}
John McCarthy. 1959.
\newblock \href {http://www-formal.stanford.edu/jmc/mcc59.html} {Programs with
  common sense}.
\newblock In \emph{Proceedings of the {T}eddington Conference on the
  Mechanization of Thought Processes}, pages 75--91.

\bibitem[{McKenzie et~al.(2023)McKenzie, Lyzhov, Pieler, Parrish, Mueller,
  Prabhu, McLean, Kirtland, Ross, Liu, Gritsevskiy, Wurgaft, Kauffman, Recchia,
  Liu, Cavanagh, Weiss, Huang, Droid, Tseng, Korbak, Shen, Zhang, Zhou, Kim,
  Bowman, and Perez}]{mckenzie2023inverse}
Ian~R. McKenzie, Alexander Lyzhov, Michael Pieler, Alicia Parrish, Aaron
  Mueller, Ameya Prabhu, Euan McLean, Aaron Kirtland, Alexis Ross, Alisa Liu,
  Andrew Gritsevskiy, Daniel Wurgaft, Derik Kauffman, Gabriel Recchia, Jiacheng
  Liu, Joe Cavanagh, Max Weiss, Sicong Huang, The~Floating Droid, Tom Tseng,
  Tomasz Korbak, Xudong Shen, Yuhui Zhang, Zhengping Zhou, Najoung Kim,
  Samuel~R. Bowman, and Ethan Perez. 2023.
\newblock \href {http://arxiv.org/abs/2306.09479} {Inverse scaling: When bigger
  isn't better}.

\bibitem[{Miceli-Barone et~al.(2023)Miceli-Barone, Barez, Konstas, and
  Cohen}]{micelibarone2023larger}
Antonio~Valerio Miceli-Barone, Fazl Barez, Ioannis Konstas, and Shay~B. Cohen.
  2023.
\newblock \href {http://arxiv.org/abs/2305.15507} {The larger they are, the
  harder they fail: Language models do not recognize identifier swaps in
  {Python}}.

\bibitem[{Mishra et~al.(2022)Mishra, Khashabi, Baral, and
  Hajishirzi}]{mishra-etal-2022-cross}
Swaroop Mishra, Daniel Khashabi, Chitta Baral, and Hannaneh Hajishirzi. 2022.
\newblock \href {https://doi.org/10.18653/v1/2022.acl-long.244} {Cross-task
  generalization via natural language crowdsourcing instructions}.
\newblock In \emph{Proceedings of the 60th Annual Meeting of the Association
  for Computational Linguistics (Volume 1: Long Papers)}, pages 3470--3487,
  Dublin, Ireland. Association for Computational Linguistics.

\bibitem[{Mollo and Milli{\`e}re(2023)}]{mollo2023vector}
Dimitri~Coelho Mollo and Rapha{\"e}l Milli{\`e}re. 2023.
\newblock \href {https://arxiv.org/abs/2304.01481} {The vector grounding
  problem}.
\newblock \emph{ArXiv preprint}, abs/2304.01481.

\bibitem[{Nabi and Shpitser(2018)}]{nabi18fair}
Razieh Nabi and Ilya Shpitser. 2018.
\newblock \href
  {https://www.aaai.org/ocs/index.php/AAAI/AAAI18/paper/view/16683} {Fair
  inference on outcomes}.
\newblock In \emph{Proceedings of the Thirty-Second {AAAI} Conference on
  Artificial Intelligence, (AAAI-18), the 30th innovative Applications of
  Artificial Intelligence (IAAI-18), and the 8th {AAAI} Symposium on
  Educational Advances in Artificial Intelligence (EAAI-18), New Orleans,
  Louisiana, USA, February 2-7, 2018}, pages 1931--1940. {AAAI} Press.

\bibitem[{Nivre et~al.(2016)Nivre, de~Marneffe, Ginter, Goldberg, Haji{\v{c}},
  Manning, McDonald, Petrov, Pyysalo, Silveira, Tsarfaty, and
  Zeman}]{nivre2016universal}
Joakim Nivre, Marie-Catherine de~Marneffe, Filip Ginter, Yoav Goldberg, Jan
  Haji{\v{c}}, Christopher~D. Manning, Ryan McDonald, Slav Petrov, Sampo
  Pyysalo, Natalia Silveira, Reut Tsarfaty, and Daniel Zeman. 2016.
\newblock \href {https://aclanthology.org/L16-1262} {{U}niversal {D}ependencies
  v1: A multilingual treebank collection}.
\newblock In \emph{Proceedings of the Tenth International Conference on
  Language Resources and Evaluation ({LREC}'16)}, pages 1659--1666,
  Portoro{\v{z}}, Slovenia. European Language Resources Association (ELRA).

\bibitem[{Nori et~al.(2023)Nori, King, McKinney, Carignan, and
  Horvitz}]{nori2023capabilities}
Harsha Nori, Nicholas King, Scott~Mayer McKinney, Dean Carignan, and Eric
  Horvitz. 2023.
\newblock \href {http://arxiv.org/abs/2303.13375} {Capabilities of {GPT-4} on
  medical challenge problems}.

\bibitem[{Nye et~al.(2021)Nye, Andreassen, Gur-Ari, Michalewski, Austin,
  Bieber, Dohan, Lewkowycz, Bosma, Luan, Sutton, and Odena}]{nye2021work}
Maxwell Nye, Anders~Johan Andreassen, Guy Gur-Ari, Henryk Michalewski, Jacob
  Austin, David Bieber, David Dohan, Aitor Lewkowycz, Maarten Bosma, David
  Luan, Charles Sutton, and Augustus Odena. 2021.
\newblock \href {http://arxiv.org/abs/2112.00114} {Show your work: Scratchpads
  for intermediate computation with language models}.

\bibitem[{OpenAI(2023)}]{gpt4}
OpenAI. 2023.
\newblock \href {http://arxiv.org/abs/2303.08774} {{GPT-4} technical report}.

\bibitem[{Patel and Pavlick(2022)}]{patel2022mapping}
Roma Patel and Ellie Pavlick. 2022.
\newblock \href {https://openreview.net/forum?id=gJcEM8sxHK} {Mapping language
  models to grounded conceptual spaces}.
\newblock In \emph{The Tenth International Conference on Learning
  Representations, {ICLR} 2022, Virtual Event, April 25-29, 2022}.
  OpenReview.net.

\bibitem[{Pearl(1988)}]{pearl1988}
Judea Pearl. 1988.
\newblock \emph{Probabilistic Reasoning in Intelligent Systems: Networks of
  Plausible Inference}.
\newblock Morgan Kaufmann Publishers Inc., San Francisco, CA, USA.

\bibitem[{Pearl(2009)}]{pearl_2009}
Judea Pearl. 2009.
\newblock \href {https://doi.org/10.1017/CBO9780511803161} {\emph{Causality}},
  2nd edition.
\newblock Cambridge University Press.

\bibitem[{Piantadosi and Hill(2022)}]{piantadosi2022meaning}
Steven Piantadosi and Felix Hill. 2022.
\newblock \href {https://openreview.net/forum?id=nRkJEwmZnM} {Meaning without
  reference in large language models}.
\newblock In \emph{NeurIPS 2022 Workshop on Neuro Causal and Symbolic AI
  (nCSI)}.

\bibitem[{Pimentel and Cotterell(2021)}]{pimentel-cotterell-2021-bayesian}
Tiago Pimentel and Ryan Cotterell. 2021.
\newblock \href {https://doi.org/10.18653/v1/2021.emnlp-main.229} {A {B}ayesian
  framework for information-theoretic probing}.
\newblock In \emph{Proceedings of the 2021 Conference on Empirical Methods in
  Natural Language Processing}, pages 2869--2887, Online and Punta Cana,
  Dominican Republic. Association for Computational Linguistics.

\bibitem[{Qin et~al.(2019)Qin, Bosselut, Holtzman, Bhagavatula, Clark, and
  Choi}]{qin-etal-2019-counterfactual}
Lianhui Qin, Antoine Bosselut, Ari Holtzman, Chandra Bhagavatula, Elizabeth
  Clark, and Yejin Choi. 2019.
\newblock \href {https://doi.org/10.18653/v1/D19-1509} {Counterfactual story
  reasoning and generation}.
\newblock In \emph{Proceedings of the 2019 Conference on Empirical Methods in
  Natural Language Processing and the 9th International Joint Conference on
  Natural Language Processing (EMNLP-IJCNLP)}, pages 5043--5053, Hong Kong,
  China. Association for Computational Linguistics.

\bibitem[{Qin et~al.(2020)Qin, Shwartz, West, Bhagavatula, Hwang, Le~Bras,
  Bosselut, and Choi}]{qin-etal-2020-back}
Lianhui Qin, Vered Shwartz, Peter West, Chandra Bhagavatula, Jena~D. Hwang,
  Ronan Le~Bras, Antoine Bosselut, and Yejin Choi. 2020.
\newblock \href {https://doi.org/10.18653/v1/2020.emnlp-main.58} {Back to the
  future: Unsupervised backprop-based decoding for counterfactual and abductive
  commonsense reasoning}.
\newblock In \emph{Proceedings of the 2020 Conference on Empirical Methods in
  Natural Language Processing (EMNLP)}, pages 794--805, Online. Association for
  Computational Linguistics.

\bibitem[{Radford et~al.(2021)Radford, Kim, Hallacy, Ramesh, Goh, Agarwal,
  Sastry, Askell, Mishkin, Clark, Krueger, and Sutskever}]{clip}
Alec Radford, Jong~Wook Kim, Chris Hallacy, Aditya Ramesh, Gabriel Goh,
  Sandhini Agarwal, Girish Sastry, Amanda Askell, Pamela Mishkin, Jack Clark,
  Gretchen Krueger, and Ilya Sutskever. 2021.
\newblock \href {http://proceedings.mlr.press/v139/radford21a.html} {Learning
  transferable visual models from natural language supervision}.
\newblock In \emph{Proceedings of the 38th International Conference on Machine
  Learning, {ICML} 2021, 18-24 July 2021, Virtual Event}, volume 139 of
  \emph{Proceedings of Machine Learning Research}, pages 8748--8763. {PMLR}.

\bibitem[{Ravfogel et~al.(2019)Ravfogel, Goldberg, and
  Linzen}]{ravfogel2019studying}
Shauli Ravfogel, Yoav Goldberg, and Tal Linzen. 2019.
\newblock \href {https://doi.org/10.18653/v1/N19-1356} {Studying the inductive
  biases of {RNN}s with synthetic variations of natural languages}.
\newblock In \emph{Proceedings of the 2019 Conference of the North {A}merican
  Chapter of the Association for Computational Linguistics: Human Language
  Technologies, Volume 1 (Long and Short Papers)}, pages 3532--3542,
  Minneapolis, Minnesota. Association for Computational Linguistics.

\bibitem[{Razeghi et~al.(2022)Razeghi, Logan~IV, Gardner, and
  Singh}]{razeghi-etal-2022-impact}
Yasaman Razeghi, Robert~L Logan~IV, Matt Gardner, and Sameer Singh. 2022.
\newblock \href {https://aclanthology.org/2022.findings-emnlp.59} {Impact of
  pretraining term frequencies on few-shot numerical reasoning}.
\newblock In \emph{Findings of the Association for Computational Linguistics:
  EMNLP 2022}, pages 840--854, Abu Dhabi, United Arab Emirates. Association for
  Computational Linguistics.

\bibitem[{Ren et~al.(2020)Ren, He, Tan, Qin, Zhao, and Liu}]{ren2020popmag}
Yi~Ren, Jinzheng He, Xu~Tan, Tao Qin, Zhou Zhao, and Tie{-}Yan Liu. 2020.
\newblock \href {https://doi.org/10.1145/3394171.3413721} {Popmag: Pop music
  accompaniment generation}.
\newblock In \emph{{MM} '20: The 28th {ACM} International Conference on
  Multimedia, Virtual Event / Seattle, WA, USA, October 12-16, 2020}, pages
  1198--1206.

\bibitem[{Reynolds and McDonell(2021)}]{Reynolds2021prompt}
Laria Reynolds and Kyle McDonell. 2021.
\newblock \href {https://doi.org/10.1145/3411763.3451760} {Prompt programming
  for large language models: ddddd the few-shot paradigm}.
\newblock In \emph{Extended Abstracts of the 2021 CHI Conference on Human
  Factors in Computing Systems}, CHI EA '21, New York, NY, USA. Association for
  Computing Machinery.

\bibitem[{Saparov and He(2023)}]{saparov2023language}
Abulhair Saparov and He~He. 2023.
\newblock Language models are greedy reasoners: A systematic formal analysis of
  chain-of-thought.
\newblock In \emph{Proceedings of ICLR}.

\bibitem[{Saparov and Mitchell(2022)}]{saparov-mitchell-2022-towards}
Abulhair Saparov and Tom~M. Mitchell. 2022.
\newblock \href {https://doi.org/10.1162/tacl_a_00463} {Towards general natural
  language understanding with probabilistic worldbuilding}.
\newblock \emph{Transactions of the Association for Computational Linguistics},
  10:325--342.

\bibitem[{Schuster and Manning(2016)}]{schuster2016enhanced}
Sebastian Schuster and Christopher~D. Manning. 2016.
\newblock \href {https://aclanthology.org/L16-1376} {Enhanced {E}nglish
  {U}niversal {D}ependencies: An improved representation for natural language
  understanding tasks}.
\newblock In \emph{Proceedings of the Tenth International Conference on
  Language Resources and Evaluation ({LREC}'16)}, pages 2371--2378,
  Portoro{\v{z}}, Slovenia. European Language Resources Association (ELRA).

\bibitem[{Sharma et~al.(2024)Sharma, Shaham, Baradad, Fu, Rodriguez-Munoz,
  Duggal, Isola, and Torralba}]{sharma2024vision}
Pratyusha Sharma, Tamar~Rott Shaham, Manel Baradad, Stephanie Fu, Adrian
  Rodriguez-Munoz, Shivam Duggal, Phillip Isola, and Antonio Torralba. 2024.
\newblock \href {http://arxiv.org/abs/2401.01862} {A vision check-up for
  language models}.

\bibitem[{Shepard and Metzler(1971)}]{shepard1971mental}
Roger~N Shepard and Jacqueline Metzler. 1971.
\newblock Mental rotation of three-dimensional objects.
\newblock \emph{Science}, 171(3972):701--703.

\bibitem[{Silver et~al.(2017)Silver, Hubert, Schrittwieser, Antonoglou, Lai,
  Guez, Lanctot, Sifre, Kumaran, Graepel, Lillicrap, Simonyan, and
  Hassabis}]{DBLP:journals/corr/abs-1712-01815}
David Silver, Thomas Hubert, Julian Schrittwieser, Ioannis Antonoglou, Matthew
  Lai, Arthur Guez, Marc Lanctot, Laurent Sifre, Dharshan Kumaran, Thore
  Graepel, Timothy~P. Lillicrap, Karen Simonyan, and Demis Hassabis. 2017.
\newblock \href {https://arxiv.org/abs/1712.01815} {Mastering chess and {Shogi}
  by self-play with a general reinforcement learning algorithm}.
\newblock \emph{ArXiv preprint}, abs/1712.01815.

\bibitem[{Singley and Anderson(1989)}]{singley1989transfer}
Mark~K Singley and John~Robert Anderson. 1989.
\newblock \emph{The transfer of cognitive skill}.
\newblock 9. Harvard University Press.

\bibitem[{Sobania et~al.(2023)Sobania, Briesch, Hanna, and
  Petke}]{sobania-2023-analysis}
Dominik Sobania, Martin Briesch, Carol Hanna, and Justyna Petke. 2023.
\newblock An analysis of the automatic bug fixing performance of {ChatGPT}.
\newblock In \emph{Proceedings of the 45th International Conference on Software
  Engineering}.

\bibitem[{S{\o}gaard(2023)}]{sogaard2023grounding}
Anders S{\o}gaard. 2023.
\newblock Grounding the vector space of an octopus: Word meaning from raw text.
\newblock \emph{Minds and Machines}, 33(1):33--54.

\bibitem[{Sordoni et~al.(2023)Sordoni, Yuan, C{\^o}t{\'e}, Pereira, Trischler,
  Xiao, Hosseini, Niedtner, and Roux}]{sordoni2023deep}
Alessandro Sordoni, Xingdi Yuan, Marc-Alexandre C{\^o}t{\'e}, Matheus Pereira,
  Adam Trischler, Ziang Xiao, Arian Hosseini, Friederike Niedtner, and
  Nicolas~Le Roux. 2023.
\newblock \href {https://arxiv.org/abs/2306.12509} {Deep language networks:
  Joint prompt training of stacked llms using variational inference}.
\newblock \emph{ArXiv preprint}, abs/2306.12509.

\bibitem[{Srivastava et~al.(2023)Srivastava, Rastogi, Rao, Shoeb, Abid, Fisch,
  Brown, Santoro, Gupta, Garriga-Alonso, Kluska, Lewkowycz, Agarwal, Power,
  Ray, Warstadt, Kocurek, Safaya, Tazarv, Xiang, Parrish, Nie, Hussain, Askell,
  Dsouza, Slone, Rahane, Iyer, Andreassen, Madotto, Santilli, Stuhlmüller,
  Dai, La, Lampinen, Zou, Jiang, Chen, Vuong, Gupta, Gottardi, Norelli,
  Venkatesh, Gholamidavoodi, Tabassum, Menezes, Kirubarajan, Mullokandov,
  Sabharwal, Herrick, Efrat, Erdem, Karakaş, Roberts, Loe, Zoph, Bojanowski,
  Özyurt, Hedayatnia, Neyshabur, Inden, Stein, Ekmekci, Lin, Howald, Orinion,
  Diao, Dour, Stinson, Argueta, Ramírez, Singh, Rathkopf, Meng, Baral, Wu,
  Callison-Burch, Waites, Voigt, Manning, Potts, Ramirez, Rivera, Siro, Raffel,
  Ashcraft, Garbacea, Sileo, Garrette, Hendrycks, Kilman, Roth, Freeman,
  Khashabi, Levy, González, Perszyk, Hernandez, Chen, Ippolito, Gilboa, Dohan,
  Drakard, Jurgens, Datta, Ganguli, Emelin, Kleyko, Yuret, Chen, Tam, Hupkes,
  Misra, Buzan, Mollo, Yang, Lee, Schrader, Shutova, Cubuk, Segal, Hagerman,
  Barnes, Donoway, Pavlick, Rodola, Lam, Chu, Tang, Erdem, Chang, Chi, Dyer,
  Jerzak, Kim, Manyasi, Zheltonozhskii, Xia, Siar, Martínez-Plumed, Happé,
  Chollet, Rong, Mishra, Winata, de~Melo, Kruszewski, Parascandolo, Mariani,
  Wang, Jaimovitch-López, Betz, Gur-Ari, Galijasevic, Kim, Rashkin,
  Hajishirzi, Mehta, Bogar, Shevlin, Schütze, Yakura, Zhang, Wong, Ng, Noble,
  Jumelet, Geissinger, Kernion, Hilton, Lee, Fisac, Simon, Koppel, Zheng, Zou,
  Kocoń, Thompson, Wingfield, Kaplan, Radom, Sohl-Dickstein, Phang, Wei,
  Yosinski, Novikova, Bosscher, Marsh, Kim, Taal, Engel, Alabi, Xu, Song, Tang,
  Waweru, Burden, Miller, Balis, Batchelder, Berant, Frohberg, Rozen,
  Hernandez-Orallo, Boudeman, Guerr, Jones, Tenenbaum, Rule, Chua, Kanclerz,
  Livescu, Krauth, Gopalakrishnan, Ignatyeva, Markert, Dhole, Gimpel, Omondi,
  Mathewson, Chiafullo, Shkaruta, Shridhar, McDonell, Richardson, Reynolds,
  Gao, Zhang, Dugan, Qin, Contreras-Ochando, Morency, Moschella, Lam, Noble,
  Schmidt, He, Colón, Metz, Şenel, Bosma, Sap, ter Hoeve, Farooqi, Faruqui,
  Mazeika, Baturan, Marelli, Maru, Quintana, Tolkiehn, Giulianelli, Lewis,
  Potthast, Leavitt, Hagen, Schubert, Baitemirova, Arnaud, McElrath, Yee,
  Cohen, Gu, Ivanitskiy, Starritt, Strube, Swędrowski, Bevilacqua, Yasunaga,
  Kale, Cain, Xu, Suzgun, Walker, Tiwari, Bansal, Aminnaseri, Geva, Gheini, T,
  Peng, Chi, Lee, Krakover, Cameron, Roberts, Doiron, Martinez, Nangia,
  Deckers, Muennighoff, Keskar, Iyer, Constant, Fiedel, Wen, Zhang, Agha,
  Elbaghdadi, Levy, Evans, Casares, Doshi, Fung, Liang, Vicol,
  Alipoormolabashi, Liao, Liang, Chang, Eckersley, Htut, Hwang, Miłkowski,
  Patil, Pezeshkpour, Oli, Mei, Lyu, Chen, Banjade, Rudolph, Gabriel, Habacker,
  Risco, Millière, Garg, Barnes, Saurous, Arakawa, Raymaekers, Frank, Sikand,
  Novak, Sitelew, LeBras, Liu, Jacobs, Zhang, Salakhutdinov, Chi, Lee, Stovall,
  Teehan, Yang, Singh, Mohammad, Anand, Dillavou, Shleifer, Wiseman, Gruetter,
  Bowman, Schoenholz, Han, Kwatra, Rous, Ghazarian, Ghosh, Casey, Bischoff,
  Gehrmann, Schuster, Sadeghi, Hamdan, Zhou, Srivastava, Shi, Singh, Asaadi,
  Gu, Pachchigar, Toshniwal, Upadhyay, Shyamolima, Debnath, Shakeri, Thormeyer,
  Melzi, Reddy, Makini, Lee, Torene, Hatwar, Dehaene, Divic, Ermon, Biderman,
  Lin, Prasad, Piantadosi, Shieber, Misherghi, Kiritchenko, Mishra, Linzen,
  Schuster, Li, Yu, Ali, Hashimoto, Wu, Desbordes, Rothschild, Phan, Wang,
  Nkinyili, Schick, Kornev, Tunduny, Gerstenberg, Chang, Neeraj, Khot, Shultz,
  Shaham, Misra, Demberg, Nyamai, Raunak, Ramasesh, Prabhu, Padmakumar,
  Srikumar, Fedus, Saunders, Zhang, Vossen, Ren, Tong, Zhao, Wu, Shen,
  Yaghoobzadeh, Lakretz, Song, Bahri, Choi, Yang, Hao, Chen, Belinkov, Hou,
  Hou, Bai, Seid, Zhao, Wang, Wang, Wang, and Wu}]{srivastava2023imitation}
Aarohi Srivastava, Abhinav Rastogi, Abhishek Rao, Abu Awal~Md Shoeb, Abubakar
  Abid, Adam Fisch, Adam~R. Brown, Adam Santoro, Aditya Gupta, Adrià
  Garriga-Alonso, Agnieszka Kluska, Aitor Lewkowycz, Akshat Agarwal, Alethea
  Power, Alex Ray, Alex Warstadt, Alexander~W. Kocurek, Ali Safaya, Ali Tazarv,
  Alice Xiang, Alicia Parrish, Allen Nie, Aman Hussain, Amanda Askell, Amanda
  Dsouza, Ambrose Slone, Ameet Rahane, Anantharaman~S. Iyer, Anders Andreassen,
  Andrea Madotto, Andrea Santilli, Andreas Stuhlmüller, Andrew Dai, Andrew La,
  Andrew Lampinen, Andy Zou, Angela Jiang, Angelica Chen, Anh Vuong, Animesh
  Gupta, Anna Gottardi, Antonio Norelli, Anu Venkatesh, Arash Gholamidavoodi,
  Arfa Tabassum, Arul Menezes, Arun Kirubarajan, Asher Mullokandov, Ashish
  Sabharwal, Austin Herrick, Avia Efrat, Aykut Erdem, Ayla Karakaş, B.~Ryan
  Roberts, Bao~Sheng Loe, Barret Zoph, Bartłomiej Bojanowski, Batuhan Özyurt,
  Behnam Hedayatnia, Behnam Neyshabur, Benjamin Inden, Benno Stein, Berk
  Ekmekci, Bill~Yuchen Lin, Blake Howald, Bryan Orinion, Cameron Diao, Cameron
  Dour, Catherine Stinson, Cedrick Argueta, César~Ferri Ramírez, Chandan
  Singh, Charles Rathkopf, Chenlin Meng, Chitta Baral, Chiyu Wu, Chris
  Callison-Burch, Chris Waites, Christian Voigt, Christopher~D. Manning,
  Christopher Potts, Cindy Ramirez, Clara~E. Rivera, Clemencia Siro, Colin
  Raffel, Courtney Ashcraft, Cristina Garbacea, Damien Sileo, Dan Garrette, Dan
  Hendrycks, Dan Kilman, Dan Roth, Daniel Freeman, Daniel Khashabi, Daniel
  Levy, Daniel~Moseguí González, Danielle Perszyk, Danny Hernandez, Danqi
  Chen, Daphne Ippolito, Dar Gilboa, David Dohan, David Drakard, David Jurgens,
  Debajyoti Datta, Deep Ganguli, Denis Emelin, Denis Kleyko, Deniz Yuret, Derek
  Chen, Derek Tam, Dieuwke Hupkes, Diganta Misra, Dilyar Buzan, Dimitri~Coelho
  Mollo, Diyi Yang, Dong-Ho Lee, Dylan Schrader, Ekaterina Shutova, Ekin~Dogus
  Cubuk, Elad Segal, Eleanor Hagerman, Elizabeth Barnes, Elizabeth Donoway,
  Ellie Pavlick, Emanuele Rodola, Emma Lam, Eric Chu, Eric Tang, Erkut Erdem,
  Ernie Chang, Ethan~A. Chi, Ethan Dyer, Ethan Jerzak, Ethan Kim, Eunice~Engefu
  Manyasi, Evgenii Zheltonozhskii, Fanyue Xia, Fatemeh Siar, Fernando
  Martínez-Plumed, Francesca Happé, Francois Chollet, Frieda Rong, Gaurav
  Mishra, Genta~Indra Winata, Gerard de~Melo, Germán Kruszewski, Giambattista
  Parascandolo, Giorgio Mariani, Gloria Wang, Gonzalo Jaimovitch-López, Gregor
  Betz, Guy Gur-Ari, Hana Galijasevic, Hannah Kim, Hannah Rashkin, Hannaneh
  Hajishirzi, Harsh Mehta, Hayden Bogar, Henry Shevlin, Hinrich Schütze,
  Hiromu Yakura, Hongming Zhang, Hugh~Mee Wong, Ian Ng, Isaac Noble, Jaap
  Jumelet, Jack Geissinger, Jackson Kernion, Jacob Hilton, Jaehoon Lee,
  Jaime~Fernández Fisac, James~B. Simon, James Koppel, James Zheng, James Zou,
  Jan Kocoń, Jana Thompson, Janelle Wingfield, Jared Kaplan, Jarema Radom,
  Jascha Sohl-Dickstein, Jason Phang, Jason Wei, Jason Yosinski, Jekaterina
  Novikova, Jelle Bosscher, Jennifer Marsh, Jeremy Kim, Jeroen Taal, Jesse
  Engel, Jesujoba Alabi, Jiacheng Xu, Jiaming Song, Jillian Tang, Joan Waweru,
  John Burden, John Miller, John~U. Balis, Jonathan Batchelder, Jonathan
  Berant, Jörg Frohberg, Jos Rozen, Jose Hernandez-Orallo, Joseph Boudeman,
  Joseph Guerr, Joseph Jones, Joshua~B. Tenenbaum, Joshua~S. Rule, Joyce Chua,
  Kamil Kanclerz, Karen Livescu, Karl Krauth, Karthik Gopalakrishnan, Katerina
  Ignatyeva, Katja Markert, Kaustubh~D. Dhole, Kevin Gimpel, Kevin Omondi, Kory
  Mathewson, Kristen Chiafullo, Ksenia Shkaruta, Kumar Shridhar, Kyle McDonell,
  Kyle Richardson, Laria Reynolds, Leo Gao, Li~Zhang, Liam Dugan, Lianhui Qin,
  Lidia Contreras-Ochando, Louis-Philippe Morency, Luca Moschella, Lucas Lam,
  Lucy Noble, Ludwig Schmidt, Luheng He, Luis~Oliveros Colón, Luke Metz,
  Lütfi~Kerem Şenel, Maarten Bosma, Maarten Sap, Maartje ter Hoeve, Maheen
  Farooqi, Manaal Faruqui, Mantas Mazeika, Marco Baturan, Marco Marelli, Marco
  Maru, Maria Jose~Ramírez Quintana, Marie Tolkiehn, Mario Giulianelli, Martha
  Lewis, Martin Potthast, Matthew~L. Leavitt, Matthias Hagen, Mátyás
  Schubert, Medina~Orduna Baitemirova, Melody Arnaud, Melvin McElrath,
  Michael~A. Yee, Michael Cohen, Michael Gu, Michael Ivanitskiy, Michael
  Starritt, Michael Strube, Michał Swędrowski, Michele Bevilacqua, Michihiro
  Yasunaga, Mihir Kale, Mike Cain, Mimee Xu, Mirac Suzgun, Mitch Walker,
  Mo~Tiwari, Mohit Bansal, Moin Aminnaseri, Mor Geva, Mozhdeh Gheini,
  Mukund~Varma T, Nanyun Peng, Nathan~A. Chi, Nayeon Lee, Neta Gur-Ari
  Krakover, Nicholas Cameron, Nicholas Roberts, Nick Doiron, Nicole Martinez,
  Nikita Nangia, Niklas Deckers, Niklas Muennighoff, Nitish~Shirish Keskar,
  Niveditha~S. Iyer, Noah Constant, Noah Fiedel, Nuan Wen, Oliver Zhang, Omar
  Agha, Omar Elbaghdadi, Omer Levy, Owain Evans, Pablo Antonio~Moreno Casares,
  Parth Doshi, Pascale Fung, Paul~Pu Liang, Paul Vicol, Pegah Alipoormolabashi,
  Peiyuan Liao, Percy Liang, Peter Chang, Peter Eckersley, Phu~Mon Htut, Pinyu
  Hwang, Piotr Miłkowski, Piyush Patil, Pouya Pezeshkpour, Priti Oli, Qiaozhu
  Mei, Qing Lyu, Qinlang Chen, Rabin Banjade, Rachel~Etta Rudolph, Raefer
  Gabriel, Rahel Habacker, Ramon Risco, Raphaël Millière, Rhythm Garg,
  Richard Barnes, Rif~A. Saurous, Riku Arakawa, Robbe Raymaekers, Robert Frank,
  Rohan Sikand, Roman Novak, Roman Sitelew, Ronan LeBras, Rosanne Liu, Rowan
  Jacobs, Rui Zhang, Ruslan Salakhutdinov, Ryan Chi, Ryan Lee, Ryan Stovall,
  Ryan Teehan, Rylan Yang, Sahib Singh, Saif~M. Mohammad, Sajant Anand, Sam
  Dillavou, Sam Shleifer, Sam Wiseman, Samuel Gruetter, Samuel~R. Bowman,
  Samuel~S. Schoenholz, Sanghyun Han, Sanjeev Kwatra, Sarah~A. Rous, Sarik
  Ghazarian, Sayan Ghosh, Sean Casey, Sebastian Bischoff, Sebastian Gehrmann,
  Sebastian Schuster, Sepideh Sadeghi, Shadi Hamdan, Sharon Zhou, Shashank
  Srivastava, Sherry Shi, Shikhar Singh, Shima Asaadi, Shixiang~Shane Gu, Shubh
  Pachchigar, Shubham Toshniwal, Shyam Upadhyay, Shyamolima, Debnath, Siamak
  Shakeri, Simon Thormeyer, Simone Melzi, Siva Reddy, Sneha~Priscilla Makini,
  Soo-Hwan Lee, Spencer Torene, Sriharsha Hatwar, Stanislas Dehaene, Stefan
  Divic, Stefano Ermon, Stella Biderman, Stephanie Lin, Stephen Prasad,
  Steven~T. Piantadosi, Stuart~M. Shieber, Summer Misherghi, Svetlana
  Kiritchenko, Swaroop Mishra, Tal Linzen, Tal Schuster, Tao Li, Tao Yu, Tariq
  Ali, Tatsu Hashimoto, Te-Lin Wu, Théo Desbordes, Theodore Rothschild, Thomas
  Phan, Tianle Wang, Tiberius Nkinyili, Timo Schick, Timofei Kornev, Titus
  Tunduny, Tobias Gerstenberg, Trenton Chang, Trishala Neeraj, Tushar Khot,
  Tyler Shultz, Uri Shaham, Vedant Misra, Vera Demberg, Victoria Nyamai, Vikas
  Raunak, Vinay Ramasesh, Vinay~Uday Prabhu, Vishakh Padmakumar, Vivek
  Srikumar, William Fedus, William Saunders, William Zhang, Wout Vossen, Xiang
  Ren, Xiaoyu Tong, Xinran Zhao, Xinyi Wu, Xudong Shen, Yadollah Yaghoobzadeh,
  Yair Lakretz, Yangqiu Song, Yasaman Bahri, Yejin Choi, Yichi Yang, Yiding
  Hao, Yifu Chen, Yonatan Belinkov, Yu~Hou, Yufang Hou, Yuntao Bai, Zachary
  Seid, Zhuoye Zhao, Zijian Wang, Zijie~J. Wang, Zirui Wang, and Ziyi Wu. 2023.
\newblock \href {http://arxiv.org/abs/2206.04615} {Beyond the imitation game:
  Quantifying and extrapolating the capabilities of language models}.

\bibitem[{Sun et~al.(2023)Sun, Yan, Ma, Ren, Yin, and Ren}]{sun2023chatgpt}
Weiwei Sun, Lingyong Yan, Xinyu Ma, Pengjie Ren, Dawei Yin, and Zhaochun Ren.
  2023.
\newblock \href {http://arxiv.org/abs/2304.09542} {Is {ChatGPT} good at search?
  {Investigating} large language models as re-ranking agent}.

\bibitem[{Tafjord et~al.(2021)Tafjord, Dalvi, and
  Clark}]{tafjord-etal-2021-proofwriter}
Oyvind Tafjord, Bhavana Dalvi, and Peter Clark. 2021.
\newblock \href {https://doi.org/10.18653/v1/2021.findings-acl.317}
  {{P}roof{W}riter: Generating implications, proofs, and abductive statements
  over natural language}.
\newblock In \emph{Findings of the Association for Computational Linguistics:
  ACL-IJCNLP 2021}, pages 3621--3634, Online. Association for Computational
  Linguistics.

\bibitem[{Tang et~al.(2023)Tang, Zheng, Li, Meng, Zhu, Liang, and
  Zhang}]{tang2023large}
Xiaojuan Tang, Zilong Zheng, Jiaqi Li, Fanxu Meng, Song-Chun Zhu, Yitao Liang,
  and Muhan Zhang. 2023.
\newblock \href {http://arxiv.org/abs/2305.14825} {Large language models are
  in-context semantic reasoners rather than symbolic reasoners}.

\bibitem[{Terwiesch(2023)}]{terwiesch2023would}
Christian Terwiesch. 2023.
\newblock Would {Chat} {GPT3} get a {Wharton} {MBA}? {A} prediction based on
  its performance in the operations management course.

\bibitem[{Tomasev et~al.(2020)Tomasev, Paquet, Hassabis, and
  Kramnik}]{DBLP:journals/corr/abs-2009-04374}
Nenad Tomasev, Ulrich Paquet, Demis Hassabis, and Vladimir Kramnik. 2020.
\newblock \href {https://arxiv.org/abs/2009.04374} {Assessing game balance with
  {AlphaZero}: Exploring alternative rule sets in chess}.
\newblock \emph{ArXiv preprint}, abs/2009.04374.

\bibitem[{Touvron et~al.(2023)Touvron, Lavril, Izacard, Martinet, Lachaux,
  Lacroix, Rozière, Goyal, Hambro, Azhar, Rodriguez, Joulin, Grave, and
  Lample}]{llama}
Hugo Touvron, Thibaut Lavril, Gautier Izacard, Xavier Martinet, Marie-Anne
  Lachaux, Timothée Lacroix, Baptiste Rozière, Naman Goyal, Eric Hambro,
  Faisal Azhar, Aurelien Rodriguez, Armand Joulin, Edouard Grave, and Guillaume
  Lample. 2023.
\newblock \href {http://arxiv.org/abs/2302.13971} {{LLaMA}: Open and efficient
  foundation language models}.

\bibitem[{Ullman and Tenenbaum(2020)}]{ullman2020bayesian}
Tomer~D. Ullman and Joshua~B. Tenenbaum. 2020.
\newblock \href {https://doi.org/10.1146/annurev-devpsych-121318-084833}
  {Bayesian models of conceptual development: Learning as building models of
  the world}.
\newblock \emph{Annual Review of Developmental Psychology}, 2(1):533--558.

\bibitem[{Vandenberg and Kuse(1978)}]{vandenberg1978mental}
Steven~G Vandenberg and Allan~R Kuse. 1978.
\newblock Mental rotations, a group test of three-dimensional spatial
  visualization.
\newblock \emph{Perceptual and motor skills}, 47(2):599--604.

\bibitem[{Veitch et~al.(2021)Veitch, D'Amour, Yadlowsky, and
  Eisenstein}]{veitch2021counterfactual}
Victor Veitch, Alexander D'Amour, Steve Yadlowsky, and Jacob Eisenstein. 2021.
\newblock \href
  {https://proceedings.neurips.cc/paper/2021/hash/8710ef761bbb29a6f9d12e4ef8e4379c-Abstract.html}
  {Counterfactual invariance to spurious correlations in text classification}.
\newblock In \emph{Advances in Neural Information Processing Systems 34: Annual
  Conference on Neural Information Processing Systems 2021, NeurIPS 2021,
  December 6-14, 2021, virtual}, pages 16196--16208.

\bibitem[{Von~Fintel and Heim(2011)}]{von2011intensional}
Kai Von~Fintel and Irene Heim. 2011.
\newblock \href
  {https://github.com/fintelkai/fintel-heim-intensional-notes/blob/master/fintel-heim-2011-intensional.pdf}
  {Intensional semantics}.
\newblock \emph{Unpublished Lecture Notes}.

\bibitem[{Wang et~al.(2022{\natexlab{a}})Wang, Deng, and
  Sun}]{wang-etal-2022-iteratively}
Boshi Wang, Xiang Deng, and Huan Sun. 2022{\natexlab{a}}.
\newblock \href {https://aclanthology.org/2022.emnlp-main.174} {Iteratively
  prompt pre-trained language models for chain of thought}.
\newblock In \emph{Proceedings of the 2022 Conference on Empirical Methods in
  Natural Language Processing}, pages 2714--2730, Abu Dhabi, United Arab
  Emirates. Association for Computational Linguistics.

\bibitem[{Wang et~al.(2023{\natexlab{a}})Wang, Wei, Schuurmans, Le, Chi,
  Narang, Chowdhery, and Zhou}]{wang2022self}
Xuezhi Wang, Jason Wei, Dale Schuurmans, Quoc Le, Ed~Chi, Sharan Narang,
  Aakanksha Chowdhery, and Denny Zhou. 2023{\natexlab{a}}.
\newblock Self-consistency improves chain of thought reasoning in language
  models.
\newblock In \emph{Proceedings of ICLR}.

\bibitem[{Wang et~al.(2023{\natexlab{b}})Wang, Ivison, Dasigi, Hessel, Khot,
  Chandu, Wadden, MacMillan, Smith, Beltagy, and Hajishirzi}]{wang2023far}
Yizhong Wang, Hamish Ivison, Pradeep Dasigi, Jack Hessel, Tushar Khot,
  Khyathi~Raghavi Chandu, David Wadden, Kelsey MacMillan, Noah~A. Smith,
  Iz~Beltagy, and Hannaneh Hajishirzi. 2023{\natexlab{b}}.
\newblock \href {http://arxiv.org/abs/2306.04751} {How far can camels go?
  {Exploring} the state of instruction tuning on open resources}.

\bibitem[{Wang et~al.(2022{\natexlab{b}})Wang, Mishra, Alipoormolabashi, Kordi,
  Mirzaei, Naik, Ashok, Dhanasekaran, Arunkumar, Stap, Pathak, Karamanolakis,
  Lai, Purohit, Mondal, Anderson, Kuznia, Doshi, Pal, Patel, Moradshahi,
  Parmar, Purohit, Varshney, Kaza, Verma, Puri, Karia, Doshi, Sampat, Mishra,
  Reddy~A, Patro, Dixit, and Shen}]{wang-etal-2022-super}
Yizhong Wang, Swaroop Mishra, Pegah Alipoormolabashi, Yeganeh Kordi, Amirreza
  Mirzaei, Atharva Naik, Arjun Ashok, Arut~Selvan Dhanasekaran, Anjana
  Arunkumar, David Stap, Eshaan Pathak, Giannis Karamanolakis, Haizhi Lai,
  Ishan Purohit, Ishani Mondal, Jacob Anderson, Kirby Kuznia, Krima Doshi,
  Kuntal~Kumar Pal, Maitreya Patel, Mehrad Moradshahi, Mihir Parmar, Mirali
  Purohit, Neeraj Varshney, Phani~Rohitha Kaza, Pulkit Verma, Ravsehaj~Singh
  Puri, Rushang Karia, Savan Doshi, Shailaja~Keyur Sampat, Siddhartha Mishra,
  Sujan Reddy~A, Sumanta Patro, Tanay Dixit, and Xudong Shen.
  2022{\natexlab{b}}.
\newblock \href {https://aclanthology.org/2022.emnlp-main.340}
  {Super-{N}atural{I}nstructions: Generalization via declarative instructions
  on 1600+ {NLP} tasks}.
\newblock In \emph{Proceedings of the 2022 Conference on Empirical Methods in
  Natural Language Processing}, pages 5085--5109, Abu Dhabi, United Arab
  Emirates. Association for Computational Linguistics.

\bibitem[{Wei et~al.(2022)Wei, Wang, Schuurmans, Bosma, brian ichter, Xia, Chi,
  Le, and Zhou}]{wei2022chain}
Jason Wei, Xuezhi Wang, Dale Schuurmans, Maarten Bosma, brian ichter, Fei Xia,
  Ed~H. Chi, Quoc~V Le, and Denny Zhou. 2022.
\newblock \href {https://openreview.net/forum?id=_VjQlMeSB_J} {Chain of thought
  prompting elicits reasoning in large language models}.
\newblock In \emph{Advances in Neural Information Processing Systems}.

\bibitem[{Wong et~al.(2023)Wong, Grand, Lew, Goodman, Mansinghka, Andreas, and
  Tenenbaum}]{wong2023word}
Lionel Wong, Gabriel Grand, Alexander~K. Lew, Noah~D. Goodman, Vikash~K.
  Mansinghka, Jacob Andreas, and Joshua~B. Tenenbaum. 2023.
\newblock \href {http://arxiv.org/abs/2306.12672} {From word models to world
  models: Translating from natural language to the probabilistic language of
  thought}.

\bibitem[{Xu et~al.(2022)Xu, Alon, Neubig, and Hellendoorn}]{xu2022systematic}
Frank~F. Xu, Uri Alon, Graham Neubig, and Vincent~Josua Hellendoorn. 2022.
\newblock \href {https://doi.org/10.1145/3520312.3534862} {A systematic
  evaluation of large language models of code}.
\newblock In \emph{Proceedings of the 6th ACM SIGPLAN International Symposium
  on Machine Programming}, MAPS 2022, page 1–10, New York, NY, USA.
  Association for Computing Machinery.

\bibitem[{Yang et~al.(2020)Yang, Obadinma, Zhao, Zhang, Matwin, and
  Zhu}]{yang-etal-2020-semeval}
Xiaoyu Yang, Stephen Obadinma, Huasha Zhao, Qiong Zhang, Stan Matwin, and
  Xiaodan Zhu. 2020.
\newblock \href {https://doi.org/10.18653/v1/2020.semeval-1.40}
  {{S}em{E}val-2020 task 5: Counterfactual recognition}.
\newblock In \emph{Proceedings of the Fourteenth Workshop on Semantic
  Evaluation}, pages 322--335, Barcelona (online). International Committee for
  Computational Linguistics.

\bibitem[{Yao et~al.(2023)Yao, Yu, Zhao, Shafran, Griffiths, Cao, and
  Narasimhan}]{yao2023tree}
Shunyu Yao, Dian Yu, Jeffrey Zhao, Izhak Shafran, Thomas~L Griffiths, Yuan Cao,
  and Karthik Narasimhan. 2023.
\newblock \href {https://arxiv.org/abs/2305.10601} {Tree of thoughts:
  Deliberate problem solving with large language models}.
\newblock \emph{ArXiv preprint}, abs/2305.10601.

\bibitem[{Yu et~al.(2020)Yu, Sie, Tedeschi, and Bergen}]{yu2020word}
Charles Yu, Ryan Sie, Nicolas Tedeschi, and Leon Bergen. 2020.
\newblock \href {https://doi.org/10.18653/v1/2020.emnlp-main.331} {Word
  frequency does not predict grammatical knowledge in language models}.
\newblock In \emph{Proceedings of the 2020 Conference on Empirical Methods in
  Natural Language Processing (EMNLP)}, pages 4040--4054, Online. Association
  for Computational Linguistics.

\bibitem[{Yu et~al.(2023)Yu, Jiang, Clark, and Sabharwal}]{yu2023ifqa}
Wenhao Yu, Meng Jiang, Peter Clark, and Ashish Sabharwal. 2023.
\newblock \href {http://arxiv.org/abs/2305.14010} {{IfQA}: A dataset for
  open-domain question answering under counterfactual presuppositions}.

\bibitem[{Zhang et~al.(2023{\natexlab{a}})Zhang, Ding, and
  Jing}]{zhang2023stance}
Bowen Zhang, Daijun Ding, and Liwen Jing. 2023{\natexlab{a}}.
\newblock \href {http://arxiv.org/abs/2212.14548} {How would stance detection
  techniques evolve after the launch of {ChatGPT}?}

\bibitem[{Zhang et~al.(2023{\natexlab{b}})Zhang, Press, Merrill, Liu, and
  Smith}]{zhang2023language}
Muru Zhang, Ofir Press, William Merrill, Alisa Liu, and Noah~A. Smith.
  2023{\natexlab{b}}.
\newblock \href {http://arxiv.org/abs/2305.13534} {How language model
  hallucinations can snowball}.

\bibitem[{Zhang et~al.(2023{\natexlab{c}})Zhang, Zhang, Vineet, Joshi, and
  Wang}]{zhang2023controllable}
Tianjun Zhang, Yi~Zhang, Vibhav Vineet, Neel Joshi, and Xin Wang.
  2023{\natexlab{c}}.
\newblock \href {https://arxiv.org/abs/2305.18583} {Controllable text-to-image
  generation with {GPT-4}}.
\newblock \emph{ArXiv preprint}, abs/2305.18583.

\bibitem[{Zhang et~al.(2020)Zhang, Ramachandran, Tenney, Elazar, and
  Roth}]{zhang-etal-2020-language-embeddings}
Xikun Zhang, Deepak Ramachandran, Ian Tenney, Yanai Elazar, and Dan Roth. 2020.
\newblock \href {https://doi.org/10.18653/v1/2020.blackboxnlp-1.27} {Do
  language embeddings capture scales?}
\newblock In \emph{Proceedings of the Third BlackboxNLP Workshop on Analyzing
  and Interpreting Neural Networks for NLP}, pages 292--299, Online.
  Association for Computational Linguistics.

\end{thebibliography}
\bibliographystyle{acl_natbib}

\clearpage
\appendix

\section{Full Setups} \label{sec:full-setups}

Unless otherwise specified, we use temperature=0 when sampling from the LMs.

\subsection{Arithmetic}

We randomly sample 1,000 two-digit addition expressions and evaluate them in bases 8, 9, 10, 11, and 16. Each base is sampled separately---for bases other than base-10, we make sure all expressions evaluate to a different result in that base compared to base-10 so that these expressions discriminate between the bases.
To ensure the LMs understand these bases, we design the CCC to ask the model what the number following a given number is. We want the model to know when to carry over and when not to, so we take the 100 smallest numbers in the given basis that ends with the maximum digit in that base, and 100 that end with 0.

\subsection{Programming} \label{sec:programming-setup}

We use the HumanEval dataset~\citep{chen2021codex} which has short Python programs and is commonly used to assess the coding ability of LMs~\citepia{bai2022training,xu2022systematic,wang2023far}. It was designed as a code-\emph{generation} dataset, where a model writes a function from a specification and is evaluated against test cases with input-output pairs. Different from our other tasks, we follow prior work~\citep{llama,wang2023far} and (1) use temperature 0.1 when evaluating pass@1 and 0.8 for pass@10, (2) sample 50 responses, and (3) only evaluate without 0-shot CoT. While the original work \cite{chen2021codex} recommended sampling 200 responses, this is very expensive, and we follow \citet{wang2023far} and only sample 50.
In Figure~\ref{fig:results-1}, we only show the performance on the subset of HumanEval where a 1-based execution of the ground-truth program fails the unit tests. These are the instances that distinguish between 0- and 1-based indexing. We also report results on the full HumanEval dataset in Table~\ref{tab:programming-gen-results}.

We also consider another setup---code \emph{execution}, where we give the LM the ground-truth program and ask the LM for the output of the test cases given the input. We remove four programs in HumanEval that are not compatible with this format (ID: 32, 38, 50, and 53), only for this execution task.
Because the program would have a different functionality under 1-based indexing, we remove the docstring that is the function description, and also rename the function to the uninformative \texttt{function}, to avoid confusing the LM.
Some programs also become invalid under 1-based indexing, specifically, those that perform any indexing using \texttt{0}. We remove all test cases that involve indexing with \texttt{0} and programs that do not have test cases left after this removal. 150 programs and 969 test cases remain.
Some of these test cases may not distinguish between 0- and 1-based indexing.
So for our main task (i.e., not CCC), we only consider test cases whose outputs are different under 0- vs. 1-based indexing, and there are 113 of them.

Because we use the same prompt to indicate the counterfactual conditions for both code generation and execution, and because we want to maintain comparability with prior work for the former, we only include CCC in the execution setup. We believe they reflect the LMs' understanding of 1-based indexing in the generation setup too.
We ask the LM for the output of simple tests about 1-based indexing such as \texttt{"qrstu"[4]} and \texttt{"qrs"[:2]}. They do not require sophisticated reasoning under the counterfactual conditions and yet are sufficient to discriminate between the default and the counterfactual conditions.
We append  5 such checks after each of the 150 programs, totaling 750 CCC.

For the execution task, we do not consider PaLM-2, because it only has a maximum of 1,024 output context length and leads to truncated, unparseable results for most test instances, especially under 0-shot CoT.

\subsection{Basic Syntactic Reasoning}

We follow \citet{ravfogel2019studying} and create synthetic variants of English with all six orderings of the subject, verb, and object. 
Given a dependency tree of a regular English sentence, we alter the order of subject and object nodes with respect to the corresponding verb. The subtrees rooted at subject or object nodes are moved as a whole, whereas other non-core dependent nodes (e.g., prepositional phrases) are kept in the original positions. We use 100 sentences from English Penn Treebank~\cite{marcus1993building}, and convert the original phrase-structure trees into Universal Dependencies~\cite{nivre2016universal} using the Stanford converter~\cite{schuster2016enhanced}. 

Our task is to identify the main verb and the main subject of a sentence. We only choose sentences where the main subject contains a single word. 
\citet{ravfogel2019studying}'s data generation procedure sometimes results in sentences in the SVO order to be unnatural English sentences. To eliminate this complexity, we retain only sentences whose SVO variant according to \citet{ravfogel2019studying}'s data generation procedure is identical to the original English sentence.

We designed the CCC to assess how well LMs understand the instruction that explains the difference of word orders in the counterfactual settings. We synthetically generate 100 simple three-word sentences (e.g., ``\texttt{anna saw john}'') in five counterfactual word orders (e.g., ``\texttt{anna john saw}'' in SOV), and ask LMs to reconstruct the original English sentences in SVO order. Conceptually, this is equivalent to asking the model to identify the subject, verb, and object in the perturbed order, but using a format that is more familiar to the LM. %

To generate the simple sentences for the CCC, we designed a simple context-free grammar where the subject and the object are sampled from the vocabulary of person names, and the verb is sampled from the set $\{$\texttt{saw, loves, calls, knows, sees}$\}$. A key feature of the sentences generated from this approach is their retained plausibility when the subject and object are interchanged. This means that given a counterfactual sentence (e.g., ``\texttt{anna john saw}''), there are two natural English sentences as candidates for reconstruction (i.e., ``\texttt{anna saw john}'' and ``\texttt{john saw anna}'').  Due to this inherent ambiguity,  LMs cannot default to the heuristic of treating the synthetic sentence as bag-of-words and then reconstructing the most natural ordering of those words in real English.
The random baseline chooses a random noun as the main subject and a random verb as the main verb.

\paragraph{A note on CCC results.}
The results for this task are shown in Table~\ref{tab:word_order_raw}. Generally, the models pass our crafted CCC challenge with decent accuracy, but we observed that, in a few cases, the LMs are confused by the reconstruction ambiguity explained above. GPT-3.5 and Claude fail in the OVS settings where they often directly copy the original sentence---e.g., instead of reconstructing ``\texttt{anna saw john}'' to ``\texttt{john saw anna}'', they simply copy the original sentence ``\texttt{anna saw john}'' as the output. Similarly, PaLM-2 often incorrectly reverses the subject and object in the SOV and VSO settings---e.g., instead of reconstructing ``\texttt{calls tom lucas}'' to ``\texttt{tom calls lucas}'', it outputs ``\texttt{lucas calls tom}''.

\subsection{Natural Language Reasoning with First-Order Logic} \label{sec:setup-folio}

We use the FOLIO dataset~\citep{han2022folio} that contains premises most of which are consistent with common sense and are hence amenable to our counterfactual study. We use the full dataset, combining the training and development sets for a total of 1,204 instances, for the logistic regression analysis in \S\ref{sec:analysis-world-commonness}. But for our counterfactual study, automatically altering the premises to violate common sense is not trivial, so one author manually rewrote the premises of a subset of 81 instances to be counterfactual, and another author verified the rewrite. Considering the analysis in \S\ref{sec:analysis-world-commonness}, we chose this subset by including every instance with premises all of which GPT-4 believes to be true and whose conclusion whose GPT-4-believed truth value matches the entailment label.

We explicitly instruct the model to use no common sense or world knowledge (\S\ref{sec:prompts}), thereby requiring symbolic reasoning. For the CCC, we ask the model if the unaltered or the altered premise is true, when both are presented as options, and expect the latter.

While the FOLIO dataset has a public release, the authors have made subsequent updates which, at the time of this paper, have not been made public. We hence do not release the LM interaction data for this task, and use a fictional example in Table~\ref{tab:prompts-folio}.

\subsection{Spatial Reasoning} \label{sec:spatial-setup}

We ask the LM for the coordinates of objects in a room.
We randomly sample 100 rooms, each with 3 different objects placed in 3 different cardinal directions specified using unit vectors (out of north $(0, -1)$, south $(0, 1)$, east $(1, 0)$, and west $(-1, 0)$ as the default conditions). Though using a downward-facing $y$-axis as the default condition may be counter-intuitive, it is natural when drawing top-to-bottom and is the convention in most image processing libraries such as OpenCV (Python), Pillow (Python), and Processing (Java, JavaScript, Python), as well as graphic design applications such as Adobe Illustrator. We believe this system is the most often encountered during LM pretraining. However, other libraries with an upward-facing $y$-axis also exist, such as matplotlib (Python), ggplot (R), and D3 (JavaScript).

For the counterfactual setting, we alter the direction--unit vector mapping, and ask for the object coordinates in the new system. We consider two direction-swapped worlds (north-south and east-west), three rotated worlds (by 90\textdegree, 180\textdegree, and 270\textdegree), and a randomly permuted world. We evaluate the relative positions of objects and report the instance-level accuracy that requires all 3 objects in a room to be located correctly as the main metric. The random accuracy is around 16.7\%.\footnote{When not considering cases where objects are placed in the same line, there are 24 permutations for placing 3 objects in 4 different directions, of which 4 can be considered correct.} We also report the object-level accuracy in Table~\ref{tab:numbers-spatial}.
As the CCC, we make sure that the LM understands the permuted world by asking it to also specify the coordinates of the unit vectors representing the 4 cardinal directions in the output.

\subsection{Drawing} \label{sec:drawing-setup}

We choose 100 objects from five Emoji\footnote{\url{https://getemoji.com}} categories: activity, travel \& places, animals \& nature, food \& drink, and objects. 
Since LMs cannot generate images at the pixel level, we use code as an intermediate abstraction for sketch generation. We do our best to select objects that are easy to draw using code, verified by multiple authors. 
We consider the Processing language for our experiment which supports a variety of shapes and colors and is widely used in visualization. Our initial experiments found this language to achieve the best drawing performance compared to other graphics and image processing frameworks, including TikZ, SVG, and matplotlib. 

For the counterfactual settings, we ask the LMs to draw the same object, but vertically flipped (i.e., upside-down), or rotated by 90\textdegree or 180\textdegree. We also ask the LMs to avoid using any transformation functions such as \texttt{rotate} and \texttt{scale} to avoid shortcuts.
Before our quantitative evaluation, we flip/rotate back the generated drawing.

We use human evaluation by asking human annotators to determine whether the drawing matches the object. We instruct the annotators to consider orientation as part of correctness and for objects that have a canonical orientation, they must be drawn in that orientation. We average the results over 4 annotators. We also show a breakdown of accuracy depending on whether an object has a canonical orientation or not, as judged by the annotators, in Table~\ref{tab:numbers-drawing-breakdown}. In addition, we consider multi-class classification accuracy using CLIP~\cite{clip} as an automatic metric, where we ask CLIP to classify the drawing into our 100 categories in a 0-shot fashion. We include the CLIP multi-class classification accuracy in Table~\ref{tab:numbers-drawing}.
We note that the accuracy of the CLIP model for our setup is not guaranteed: first, our generated sketches may be distributionally different from the predominantly photorealistic images in CLIP's training data; also, CLIP might be insensitive to the object's orientation, but that distinguishes between our default and counterfactual settings. Therefore, to verify the reliability of this automatic evaluation, we randomly sample 10 objects for each model and for each default/counterfactual setting, and perform human evaluation on the 240 generated images. We find that CLIP's judgment aligns with human annotators' 84\% of the time, suggesting the reliability of this evaluation.

For this task, we do not consider PaLM-2 due to its limited context length. Our preliminary experiments also found PaLM-2  to struggle in generating parseable Processing code, even in the default setting.

We construct the CCC baseline by requiring the LMs to additionally draw a line at the top of the figure and flip/rotate it as well. A successful flipping/rotation of the line, as judged by the annotators and verified in the generated code if necessary, demonstrates an understanding of the counterfactual world.

\subsection{Music}

\subsubsection{Playing Chords on Instruments}

We measure LMs' abilities to give correct fret placements for ukulele and guitar chords in an existing database\footnote{\url{https://github.com/tombatossals/chords-db}}\footnote{We heuristically filter out incorrect datapoints by filtering out chords that either have the wrong number of notes or lack the root note.}. We include the following kinds of chords from the database: sus2 (suspended second chord), sus4 (suspended fourth chord), min triad (minor triad), maj triad (major triad), dim7 (diminished seventh chord), aug7 (augmented seventh chord), maj7 (major seventh chord), min7 (minor seventh chord), dom7 (dominant seventh chord), 5 (fifth interval), and 6 (sixth chord).

In the counterfactual setting, we instruct LMs to provide fret placements for a ``special'' ukulele or guitar where one of the strings is altered. We experiment with perturbations of different sizes: For guitar, we experiment with one-string changes by one note (\emph{E\textbf{A}DGBE} $\rightarrow$ \emph{E\textbf{B}DGBE}; \emph{\textbf{E}ADGBE} $\rightarrow$ \emph{\textbf{F}ADGBE}), one-string changes by two notes ($\rightarrow$ \emph{E\textbf{C}DGBE}), and two string changes ($\rightarrow$ \emph{E\textbf{CF}GBE}). We also experiment with a one-string change that corresponds to a common alternate tuning of a guitar called drop-D tuning ($\rightarrow$ \emph{\textbf{D}ADGBE}). For ukulele, we experiment with one-string changes by one note (\emph{\textbf{G}CEA} $\rightarrow$ \emph{\textbf{F}CEA}; $\rightarrow$ \emph{\textbf{A}CEA}), one-string change by two notes ($\rightarrow$ \emph{\textbf{B}CEA}), and two-string changes by two notes ($\rightarrow$ \emph{\textbf{BE}EA}).
The generated fret placements for a chord are considered \emph{correct} if all and only the notes in the corresponding chord (\eg \emph{C, E, G} for a C major triad) are produced, irrespective of order. 

As the CCC, we assess LMs' understanding of the given instrument's strings by asking them to identify what notes a given sequence of frets corresponds to; for the CCC, the sequences are either all fret 0, all fret 1, or all fret 2. We compute CCC accuracy at the fret level (as opposed to the sequence level).

\subsubsection{Retrieving Notes of Famous Melodies}

For 8 famous melodies, we prompt LMs to retrieve the $n$-th note in the melody, where $n$ is between 1 and 7 (inclusive). In the counterfactual setting, we prompt the LM to do the same but in a different key. The list of melodies and keys we experiment with is below. 

We use \emph{C Major} as the key for songs as the default condition given its popularity for famous melodies like children's songs. We use other keys as the counterfactual keys.\footnote{We note that some songs may have multiple canonical keys (\eg ``Twinkle Twinkle Little Star'' is also frequently performed in keys like G major or D Major.) In some initial exploration, we validated that C Major was at least one of the canonical keys for the melodies chosen, both by verifying that popular sheet music for these songs was written in C Major, and by asking GPT-3.5 to generate the melodies in an unspecified key and verifying that the generated key was C Major.
}

As the CCC, we assess LMs' understanding of the given keys by asking them to retrieve the $n$-th note of the scale of the given key.

\paragraph{Melodies:} Twinkle Twinkle Little Star, Mary Had a Little Lamb, Happy Birthday to You, Somewhere Over the Rainbow, Row Row Row Your Boat, Old Macdonald Had a Farm, Itsy Bitsy Spider, London Bridge is Falling Down.

\paragraph{Counterfactual Keys:} B\# major, C\# major, Db major, D major, D\# major, Eb major, Fb major, E major, E\# major, F major, F\# major, Gb major, G major, G\# major, Ab major, A major, A\# major, Bb major, Cb major, B major.

\subsection{Chess}

We evaluate an LM's ability to understand chess rules by checking if it can determine whether a 4-move opening follows the rules of chess or not.
In the counterfactual setting, we swap the position of bishops and knights on the board and evaluate the same task. For each setting, we randomly sample 400 unique chess openings via a procedural generation algorithm:
200 are legal for the default setting but not for the counterfactual setting, and vice versa for the other 200, ensuring a more balanced and fair classification problem.
We represent the moves as the LM input using the PGN format, the standard for chess moves description.

For the CCC, we ask an LM for the starting positions of the four knights and four bishops on the board to make sure it understands the new initial board. For both the default and counterfactual settings, we ask for the positions of white knights, white bishops, black knights, and black bishops, totaling 8 pieces, and evaluate using accuracy. Since concluding the effectiveness of our counterfactual prompt using merely 8 CCC may not be statistically significant, we sample 15 LM responses using temperature=0.1 for asking about each piece.

\subsection{SET Game}

We synthetically generate SET boards, consisting of 12 cards, each with exactly one 3-card SET that satisfies the game rules in \S\ref{sec:set-setup}. 
We represent each card with a string representation, e.g., \texttt{(3|open|red|diamond)}. 
In preliminary experiments, we tried to ask the LMs to find the SET directly, but found that they cannot perform this task well (see Figure~\ref{fig:set-nhints}, ``Number of Cards to Find''$=3$).
Therefore, in our main evaluation, we expose 2 cards in the SET and ask the LM to identify the missing one that completes the SET.

In the counterfactual setting, we invert the rule for the \emph{number} attribute to require that two cards in the SET should have the same number but the other card should be different.
For the CCC, we ask the model to verify the validity of a given SET instead of finding it. In each CCC instance, we either give a valid SET from the board, or 3 randomly sampled cards that do not constitute a valid SET. We ask the model to classify whether the given combination is valid or invalid. We note that our counterfactual perturbation ensures that the each SET cannot be simultaneously valid in the default setting and the counterfactual setting, and hence this CCC is discriminative between the two settings.

\section{Prompts} \label{sec:prompts}

We provide the exact prompts that we used to query the LMs in Tables~\ref{tab:prompts-math} to \ref{tab:prompts-set-control}. For clarity, we give a concrete prompt that embeds a test instance, rather than a template. We explain minor design decisions in the respective captions. We do not use the system message field for any model.

\section{Raw Results} \label{sec:raw-results}

We show the numeric results in Tables~\ref{tab:numbers-math} to \ref{tab:numbers-set-hints}.

\begin{table*}[t!]
    \centering
    \begin{tabular}{cl}
        \toprule
        Mode & Prompt \\
        \midrule
        Test & \texttt{\thead{You are a mathematician. Assuming that all numbers are in base-11 where the\\digits are "0123456789A", what is 59+37? \textbf{\{Let's think step by step, and \}}end\\the response with the result in "\textbackslash boxed\{result\}".}} \\
        \midrule
        CCC & \texttt{\thead{You are a mathematician. Assuming that all numbers are in base-11 where the\\digits are "0123456789A", what is the next number after 11A? Do this by\\counting the few preceding numbers and completing the sequence. End\\the response with the result.}} \\
        \midrule
        Few-Shot CoT & \texttt{\thead{You are a mathematician. Assuming that all numbers are in base-11 where the\\digits are "0123456789A", what is 25+68? Let's think step by step, and end\\the response with the result in "\textbackslash boxed\{result\}". We add the ones digits first.\\In base-11, 5+8=12. So the ones digit of the final sum is 2. We need to carry\\over the 1 to the tens place. Then we add the tens digits. In base-11, 2+6=8.\\Since we carried over the 1, 8+1=9. So the tens digit of the final sum is 9.\\Putting the digits of the final sum together, we get \textbackslash boxed\{92\}.\\...[optionally more demonstrations in the same format]...\\You are a mathematician. Assuming that all numbers are in base-11 where the\\digits are "0123456789A", what is 59+37? Let's think step by step, and end\\the response with the result in "\textbackslash boxed\{result\}".}} \\
        \bottomrule
    \end{tabular}
    \caption{\label{tab:prompts-math}
    Prompts for the arithmetic task. \texttt{\textbf{\{Let's think step by step, and \}}} is added only if 0-shot CoT is used (and the following \texttt{e} is capitalized without 0-shot CoT). We use the \texttt{\textbackslash boxed\{result\}} syntax to wrap results because we found in preliminary experiments that the models tend to use this format even without this specification.
    The Few-Shot CoT prompt is used for the analysis in \S\ref{sec:analysis-icl}.
    }
\end{table*}

\begin{table*}[t!]
    \centering
    \scalebox{0.85}{
    \begin{tabular}{cl}
        \toprule
        Mode & Prompt \\
        \midrule
        Default & \texttt{\thead{You are an expert programmer. What does the following code snippet in\\Python 3.7 print?\\\textasciigrave\textasciigrave\textasciigrave python\\def function(lst):\\\ \ \ \ return sum([lst[i] for i in range(1, len(lst), 2) if lst[i] \% 2 == 0])\\\\print([function([4, 88])])\\print([function([4, 5, 6, 7, 2, 122])])\\print([function([4, 0, 6, 7])])\\print([function([4, 4, 6, 8])])\\print([list(range(3))])\\print([[4, 5, 6].pop(2)])\\print(["qrs"[:2]])\\print(["qrstu"[4]])\\print([list(enumerate("qrstuv"))])\\\textasciigrave\textasciigrave\textasciigrave\\\textbf{\{Let's think step by step. Write out intermediate results and reasoning}\\\textbf{processes as needed. \}}End the response by saying "The final output is:"\\and a unified summary \textasciigrave\textasciigrave\textasciigrave python\textasciigrave\textasciigrave\textasciigrave\ code block with *ALL* the output, in\\which each line represents the output of each print statement.}} \\
        \midrule
        CF & \texttt{\thead{You are an expert programmer who can readily adapt to new programming\\languages. There is a new programming language, ThonPy, which is identical to\\Python 3.7 except all variables of the \textasciigrave list\textasciigrave , \textasciigrave tuple\textasciigrave , and \textasciigrave str\textasciigrave\ types use\\1-based indexing, like in the MATLAB and R languages, where sequence indices\\start from 1. That is, index \textasciigrave n\textasciigrave\ represents the \textasciigrave n\textasciigrave-th element in a sequence,\\NOT the \textasciigrave n+1\textasciigrave-th as in 0-based indexing. This change only affects when the\\index is non-negative. When the index is negative, the behavior is the same\\as Python 3.7. This also affects methods of these classes such as \textasciigrave index\textasciigrave\ and\\\textasciigrave pop\textasciigrave. The built-in functions \textasciigrave enumerate\textasciigrave\ and \textasciigrave range\textasciigrave\ also use 1-based\\indexing: by default, the index of \textasciigrave enumerate\textasciigrave\ starts from 1, and so does the\\lower bound of \textasciigrave range\textasciigrave\ when not supplied (the higher bound is unchanged).\\\\For example,\\\textasciigrave\textasciigrave\textasciigrave thonpy\\assert (7, 8, 9)[1] == 7\\assert ["abc", "def", "ghi"][3] == "ghi"\\assert "abcde"[4] == "d"\\assert "abc"[:2] == "a"\\assert [7, 8, 9][1:] == [7, 8, 9][1:5] == [7, 8, 9][1::1] == [7, 8, 9][:4]\\== [9, 8, 7][::-1] == [9, 8, 7, 6][3::-1] == [7, 8, 9]\\assert list(enumerate([7, 8, 9])) == [(1, 7), (2, 8), (3, 9)]\\assert list(range(2)) == [1]\\assert list(range(2, 4)) == [2, 3]\\assert \{0: 7, 1: 8, 2: 9\}[1] == 8\\assert [7, 8, 9].index(8) == 2\\\textasciigrave\textasciigrave\textasciigrave\\\\What does the following code snippet in ThonPy print?\\\textasciigrave\textasciigrave\textasciigrave thonpy\\def function(lst):\\\ \ \ \ return sum([lst[i] for i in range(1, len(lst), 2) if lst[i] \% 2 == 0])\\\\print([function([4, 88])])\\print([function([4, 5, 6, 7, 2, 122])])\\print([function([4, 0, 6, 7])])\\print([function([4, 4, 6, 8])])\\print([list(range(3))])\\print([[4, 5, 6].pop(2)])\\print(["qrs"[:2]])\\print(["qrstu"[4]])\\print([list(enumerate("qrstuv"))])\\\textasciigrave\textasciigrave\textasciigrave\\\textbf{\{Let's think step by step. Write out intermediate results and reasoning}\\\textbf{processes as needed. \}}End the response by saying "The final output is:"\\and a unified summary \textasciigrave\textasciigrave\textasciigrave thonpy\textasciigrave\textasciigrave\textasciigrave\ code block with *ALL* the output, in\\which each line represents the output of each print statement.}} \\
        \bottomrule
    \end{tabular}
    }
    \caption{\label{tab:prompts-programming}
    Prompts for the program execution task. \texttt{\textbf{\{Let's think step by step. Write out intermediate results and reasoning processes as needed. \}}} is added only if 0-shot CoT is used. All the print statements wrap the expression in a singleton list for the ease of parsing, so that (a) each output always takes a single line even with line breaks in the middle, and (b) we can distinguish between a string representation of e.g. an integer and the integer type.
    }
\end{table*}

\begin{table*}[t!]
    \centering
    \begin{tabular}{cl}
        \toprule
        Mode & Prompt \\
        \midrule
        Default & \texttt{\thead{You are an expert programmer. Complete the following function in Python 3.7. Please\\only output the code for the completed function.\\\\\\def add(lst):\\\ \ \ \ """Given a non-empty list of integers lst. add the even elements that are at odd\\indices..\\\\\\\ \ \ \ Examples:\\\ \ \ \ \ \ \ \ add([4, 2, 6, 7]) ==> 2 \\\ \ \ \ """\\}} \\
        \midrule
        CF & \texttt{\thead{You are an expert programmer who can readily adapt to new programming\\languages. There is a new programming language, ThonPy, which is identical to\\Python 3.7 except all variables of the \textasciigrave list\textasciigrave , \textasciigrave tuple\textasciigrave , and \textasciigrave str\textasciigrave\ types use\\1-based indexing, like in the MATLAB and R languages, where sequence indices\\start from 1. That is, index \textasciigrave n\textasciigrave\ represents the \textasciigrave n\textasciigrave-th element in a sequence,\\NOT the \textasciigrave n+1\textasciigrave-th as in 0-based indexing. This change only affects when the\\index is non-negative. When the index is negative, the behavior is the same\\as Python 3.7. This also affects methods of these classes such as \textasciigrave index\textasciigrave\ and\\\textasciigrave pop\textasciigrave. The built-in functions \textasciigrave enumerate\textasciigrave\ and \textasciigrave range\textasciigrave\ also use 1-based\\indexing: by default, the index of \textasciigrave enumerate\textasciigrave\ starts from 1, and so does the\\lower bound of \textasciigrave range\textasciigrave\ when not supplied (the higher bound is unchanged).\\\\For example,\\\textasciigrave\textasciigrave\textasciigrave thonpy\\assert (7, 8, 9)[1] == 7\\assert ["abc", "def", "ghi"][3] == "ghi"\\assert "abcde"[4] == "d"\\assert "abc"[:2] == "a"\\assert [7, 8, 9][1:] == [7, 8, 9][1:5] == [7, 8, 9][1::1] == [7, 8, 9][:4]\\== [9, 8, 7][::-1] == [9, 8, 7, 6][3::-1] == [7, 8, 9]\\assert list(enumerate([7, 8, 9])) == [(1, 7), (2, 8), (3, 9)]\\assert list(range(2)) == [1]\\assert list(range(2, 4)) == [2, 3]\\assert \{0: 7, 1: 8, 2: 9\}[1] == 8\\assert [7, 8, 9].index(8) == 2\\\textasciigrave\textasciigrave\textasciigrave\\\\Complete the following function in ThonPy. Please only output the code for the\\completed function.\\\\\\def add(lst):\\\ \ \ \ """Given a non-empty list of integers lst. add the even elements that are at odd\\indices..\\\\\\\ \ \ \ Examples:\\\ \ \ \ \ \ \ \ add([4, 2, 6, 7]) ==> 2 \\\ \ \ \ """\\}} \\
        \bottomrule
    \end{tabular}
    \caption{\label{tab:prompts-programming-gen}
    Prompts for the program generation task.
    }
\end{table*}

\begin{table*}[t!]
    \centering
    \scalebox{0.85}{
    \begin{tabular}{cl}
        \toprule
        Mode & Prompt \\
        \midrule
        Default & \texttt{\thead{You are an expert in linguistics. Your task is to identify the main verb and the main subject \\ of a sentence in English.   Show the main verb (a single word) and its subject (also a single word) after \\ the prefix `Main verb and subject: '.  \\ Sentence:  japan had just opened its doors to the world after about 250 years of isolation . \\ 
        \textbf{\{Let's think step by step. \}}
        }}  \\
        \midrule
        CF & \texttt{\thead{You are an expert in linguistics. Imagine a language that is the same as English with the only \\ exception being that it uses the verb-object-subject order instead of the subject-verb-object order. \\ Your task is to identify the main verb and the main subject in a sentence in this imaginary language. \\ Show the main verb (a single word) and its subject (also a single word) after the prefix \\ `Main verb and subject: '. \\ Sentence: had just opened its doors japan to the world after about 250 years of isolation . \\ 
        \textbf{\{Let's think step by step. \}}
        }}  \\
        \midrule
        CCC & \texttt{\thead{You are an expert in linguistics. Imagine a language that is the same as English with the only \\ exception being that it uses the verb-subject-object order instead of the subject-verb-object order. \\ Your task is to reconstruct the original sentence in English. You should  only use the words in the same \\ form as they appear in the given sentence. \\ 
        Sentence: saw anna john \\
        Show the original sentence at the end after the prefix `Original sentence: '.   \\  
        \textbf{\{Let's think step by step. \}}  
        }} \\
        \bottomrule 
    \end{tabular}
    }
    \label{prompt:word_order_raw}
    \caption{
    Prompts for the basic syntactic reasoning task. \texttt{\textbf{\{Let's think step by step. \}}} is added only if 0-shot CoT is used. 
    }
\end{table*}

\begin{table*}[t!]
    \centering
    \begin{tabular}{cl}
        \toprule
        Mode & Prompt \\
        \midrule
        Test & \texttt{\thead{Consider the following premises: "All corgis are reptiles. All reptiles are plants."\\Assuming no other commonsense or world knowledge, is the sentence "All corgis are\\plants." necessarily true, necessarily false, or neither? \textbf{\{Let's think step by step, }\\\textbf{and \}}end the response with either "necessarily true", "necessarily false", or "neither".}} \\
        \midrule
        CCC & \texttt{\thead{Consider the following premises: "All corgis are reptiles. All reptiles are plants."\\Assuming no other commonsense or world knowledge, which sentence between (a) "All\\corgis are reptiles." and (b) "All corgis are mammals." is definitely true? Answer just\\"(a)" or "(b)" and nothing else. You MUST choose one and only one, so DO NOT say neither\\or both.}} \\
        \bottomrule
    \end{tabular}
    \caption{\label{tab:prompts-folio}
    Prompts for the natural language reasoning task. \texttt{\textbf{\{Let's think step by step, and \}}} is added only if 0-shot CoT is used (and the following \texttt{e} is capitalized without 0-shot CoT). We only use a made-up example here rather than one in the dataset due to the non-publicness of the dataset~(\S\ref{sec:setup-folio}). Default and counterfactual tasks share the same test template, but the instances themselves are changed to be counterfactual. For the CCC, we separate each changed premise in an instance into a separate prompt. The default statement and the counterfactual statement are matched to (a) and (b) randomly. We do not distinguish between CCC with or without 0-shot CoT.
    }
\end{table*}

\begin{table*}[t!]
    \centering
    \begin{tabular}{cl}
        \toprule
        Mode & Prompt \\
        \midrule
        Default & \texttt{\thead{You are in the middle of a kitchen. There is a microwave on the south side. There is a \\fridge on the west side. There is a coffee machine on the north side. We define the \\following directions. The north direction is (0, -1). The south direction is (0, 1). \\The east direction is (1, 0). The west direction is (-1, 0). What's the layout of the \\room in the following format? You can estimate the size of the objects. \\\\ \textasciigrave\textasciigrave\textasciigrave json \\
\{"name": "??", "width": 500, "height": 500, "directions": \{"north": "??", "south": "??",\\ "east": "??", "west": "??"\}, "objects": [\{"name": "??", "x": "??", "y": "??"\}]\}
\\\textasciigrave\textasciigrave\textasciigrave \\\\\textbf{\{Let's think step by step.\}}}} \\\midrule
        CF & \texttt{\thead{You are in the middle of a kitchen. There is a microwave on the south side. There is a \\fridge on the west side. There is a coffee machine on the north side. We define the \\following directions. The north direction is (-1, 0). The south direction is (1, 0). \\The east direction is (0, -1). The west direction is (0, 1). What's the layout of the \\room in the following format? You can estimate the size of the objects. \\\\ \textasciigrave\textasciigrave\textasciigrave json \\
\{"name": "??", "width": 500, "height": 500, "directions": \{"north": "??", "south": "??",\\ "east": "??", "west": "??"\}, "objects": [\{"name": "??", "x": "??", "y": "??"\}]\}
\\\textasciigrave\textasciigrave\textasciigrave \\\\\textbf{\{Let's think step by step.\}}}} \\     
        \bottomrule
    \end{tabular}
    \caption{\label{tab:prompts-spatial}
    Prompts for the spatial reasoning task. \texttt{\textbf{\{Let's think step by step.\}}} is added only if 0-shot CoT is used.} 
\end{table*}

\begin{table*}[t!]
    \centering
    \begin{tabular}{cl}
        \toprule
        Mode & Prompt \\
        \midrule
        Default 1 & \texttt{\thead{You are an expert programmer and drawer. Imagine an image: there is a line at the top \\ and a house in the middle. Can you try your best to draw this image using the\\ \textasciigrave processing\textasciigrave~language? Please do not use any built-in transformation functions, such as \\ \textasciigrave rotate\textasciigrave and \textasciigrave scale\textasciigrave. Also, avoid defining any custom transformation functions yourself. \\ Do not load any existing images. Please include as many details of the house as \\possible and put everything together in the end. \\\\\textbf{\{Let's think step by step.\}}}} \\
        \midrule
        Default 2 & \texttt{\thead{You are an expert programmer and drawer. Imagine an image: there is a line at the top \\ and a house in the middle. Can you try your best to draw the line and the house \\ using the \textasciigrave processing\textasciigrave~language? Please do not use any built-in transformation \\ functions, such as \textasciigrave rotate\textasciigrave, \textasciigrave scale\textasciigrave, and \textasciigrave translate\textasciigrave. Also, avoid defining any \\ custom transformation functions yourself. Do not load any existing images. Please \\ include as  many details of the house as possible and put everything together in the \\ end. \\\\\textbf{\{Let's think step by step.\}}}} \\\midrule
        CF 1 & \texttt{\thead{You are an expert programmer and drawer. Imagine an image: there is a line at the top \\ and a house in the middle. Can you rotate this image 180 degrees and try your best to \\ draw it using the \textasciigrave processing\textasciigrave~language? Please do not use any built-in transformation \\ functions, such as \textasciigrave rotate\textasciigrave and \textasciigrave scale\textasciigrave. Also, avoid defining any custom transformation \\ functions yourself. Do not load any existing images. Do not draw the original objects. \\ Please include as many details of the house as possible and put everything together \\ in the end. \\\\\textbf{\{Let's think step by step.\}}}} \\    
        \midrule
        CF 2 & \texttt{\thead{You are an expert programmer and drawer. Imagine an image: there is a line at the top \\ and a house in the middle. Can you rotate this image 180 degrees and try your best to \\ draw the 180-degree rotated line and the 180-degree rotated house using the \textasciigrave processing\textasciigrave \\ language? Please do not use any built-in transformation functions, such as \textasciigrave rotate\textasciigrave, \\ \textasciigrave scale\textasciigrave, and \textasciigrave translate\textasciigrave. Also, avoid defining any custom transformation functions \\ yourself. Do not load any existing images. Do not draw the original objects. Please \\ include as many details of the house as possible and put everything together in the \\ end. \\\\\textbf{\{Let's think step by step.\}}}} \\    
        \bottomrule
    \end{tabular}
    \caption{\label{tab:prompts-drawing}
    Prompts for the drawing task. \texttt{\textbf{\{Let's think step by step.\}}} is added only if 0-shot CoT is used. We use prompt 1 for GPT-4 and prompt 2 for GPT-3.5 and Claude. We chose the prompt based on the best CCC accuracy for each respective model. In our preliminary experiments, we found that switching the prompt hurts CCC accuracy by more than 20\% for both GPT-4 and GPT-3.5. Claude does not follow our instructions when using prompt 1, leading to almost 0\% CCC's accuracy.} 
\end{table*}

\begin{table*}[t!]
    \centering
    \begin{tabular}{cl}
        \toprule
        Mode & Prompt \\
        \midrule
        Default & \texttt{\thead{You are an expert guitar player. I have a guitar with standard strings E-A-D-G-B-E.\\\\I want you to tell me how I could play the E minor triad on this guitar.\\\\Make sure to choose one final answer, which you should start with 'ANSWER:' and specify in\\ the following format: \\\\ANSWER:\\E string: fret FRET\\A string: fret FRET\\D string: fret FRET\\G string: fret FRET\\B string: fret FRET\\E string: fret FRET\\\\Use fret 0 to indicate an open string and fret X to indicate not playing a string. Each\\ increase in fret corresponds to an increase in half a note.\\\\ \textbf{\{Let's think step by step.\}}}} \\\midrule
        CF & \texttt{\thead{You are an expert guitar player. I have a special guitar with strings tuned to\\E-C-F-G-B-E instead of the standard E-A-D-G-B-E. Note that what is the standard A string\\ is instead tuned to C, and the standard D string is instead tuned to F. All other strings\\ are the same.\\\\I want you to tell me how I could play the E minor triad on this guitar.\\\\Make sure to choose one final answer, which you should start with 'ANSWER:' and specify in\\ the following format: \\\\ANSWER:\\E string: fret FRET\\C string: fret FRET\\F string: fret FRET\\G string: fret FRET\\B string: fret FRET\\E string: fret FRET\\\\Use fret 0 to indicate an open string and fret X to indicate not playing a string. Each\\ increase in fret corresponds to an increase in half a note.\\\\ \textbf{\{Let's think step by step.\}}}} \\
        \bottomrule
    \end{tabular}
    \caption{\label{tab:prompts-chords-guitar}
    Prompts for chord fingering: guitar. \texttt{\textbf{\{Let's think step by step.\}}} is added only if 0-shot CoT is used.
    }
\end{table*}

\begin{table*}[t!]
    \centering
    \begin{tabular}{cl}
        \toprule
        Mode & Prompt \\
        \midrule
        Default & \texttt{\thead{You are an expert guitar player. I have a guitar with standard strings E-A-D-G-B-E.\\\\I want you to tell me what notes the following sequences of finger positions corresponds to: \\\\E string: fret 0\\A string: fret 0\\D string: fret 0\\G string: fret 0\\B string: fret 0\\E string: fret 0\\\\Note that fret 0 indicates an open string, and each increase in fret corresponds to an\\ increase in half a note. \\\\Make sure to choose one final answer, which you should start with 'ANSWER:' and format\\ with dash-separated notes in the order of strings E-A-D-G-B-E.\\\\\textbf{\{Let's think step by step.\}}}} \\
        \midrule
        CF & \texttt{\thead{You are an expert guitar player. I have a special guitar with strings tuned to\\ E-C-F-G-B-E instead of the standard E-A-D-G-B-E. Note that what is the standard A string\\ is instead tuned to C, and the standard D string is instead tuned to F. All other strings\\ are the same.\\\\I want you to tell me what notes the following sequences of finger positions corresponds to: \\\\E string: fret 0\\C string: fret 0\\F string: fret 0\\G string: fret 0\\B string: fret 0\\E string: fret 0\\\\Note that fret 0 indicates an open string, and each increase in fret corresponds to an\\ increase in half a note. \\\\Make sure to choose one final answer, which you should start with 'ANSWER:' and format\\ with dash-separated notes in the order of strings E-C-F-G-B-E.\\\\\textbf{\{Let's think step by step.\}}}} \\
        \bottomrule
    \end{tabular}
    \caption{\label{tab:prompts-chords-controls-guitar}
    CCC prompts for chord fingering: guitar. \texttt{\textbf{\{Let's think step by step.\}}} is added only if 0-shot CoT is used.
    }
\end{table*}

\begin{table*}[t!]
    \centering
    \begin{tabular}{cl}
        \toprule
        Mode & Prompt \\
        \midrule
        Default & \texttt{\thead{You are an expert ukulele player. I have a ukulele with standard strings G-C-E-A.\\\\I want you to tell me how I could play the E minor triad on this ukulele.\\\\Make sure to choose one final answer, which you should start with 'ANSWER:' and specify in\\ the following format: \\\\ANSWER:\\G string: fret FRET\\C string: fret FRET\\E string: fret FRET\\A string: fret FRET\\\\Use fret 0 to indicate an open string and fret X to indicate not playing a string. Each\\ increase in fret corresponds to an increase in half a note.\\\\ \textbf{\{Let's think step by step.\}}}} \\\midrule
        CF & \texttt{\thead{You are an expert ukulele player. I have a special ukulele with strings tuned to F-C-E-A\\ instead of the standard G-C-E-A. Note that what is the standard G string is instead tuned to F.\\ All other strings are the same.\\\\I want you to tell me how I could play the E minor triad on this ukulele.\\\\Make sure to choose one final answer, which you should start with 'ANSWER:' and specify in\\ the following format: \\\\ANSWER:\\F string: fret FRET\\C string: fret FRET\\E string: fret FRET\\A string: fret FRET\\\\Use fret 0 to indicate an open string and fret X to indicate not playing a string. Each\\ increase in fret corresponds to an increase in half a note.\\\\\textbf{\{Let's think step by step.\}}}} \\
        \bottomrule
    \end{tabular}
    \caption{\label{tab:prompts-chords-ukulele}
    Prompts for chord fingering: ukulele. \texttt{\textbf{\{Let's think step by step.\}}} is added only if 0-shot CoT is used. 
    } 
\end{table*}

\begin{table*}[t!]
    \centering
    \begin{tabular}{cl}
        \toprule
        Mode & Prompt \\
        \midrule
        Default & \texttt{\thead{You are an expert ukulele player. I have a ukulele with standard strings G-C-E-A.\\\\I want you to tell me what notes the following sequences of finger positions corresponds to: \\\\G string: fret 0\\C string: fret 0\\E string: fret 0\\A string: fret 0\\\\Note that fret 0 indicates an open string, and each increase in fret corresponds to an\\ increase in half a note. \\\\Make sure to choose one final answer, which you should start with 'ANSWER:' and format\\ with dash-separated notes in the order of strings G-C-E-A.\\\\\textbf{\{Let's think step by step.\}}}} \\
        \midrule
        CF & \texttt{\thead{You are an expert ukulele player. I have a special ukulele with strings tuned to F-C-E-A\\ instead of the standard G-C-E-A. Note that what is the standard G string is instead tuned to F.\\ All other strings are the same.\\\\I want you to tell me what notes the following sequences of finger positions corresponds to: \\\\F string: fret 0\\C string: fret 0\\E string: fret 0\\A string: fret 0\\\\Note that fret 0 indicates an open string, and each increase in fret corresponds to an\\ increase in half a note. \\\\Make sure to choose one final answer, which you should start with 'ANSWER:' and format\\ with dash-separated notes in the order of strings F-C-E-A.\\\\\textbf{\{Let's think step by step.\}}}} \\
        \bottomrule
    \end{tabular}
    \caption{\label{tab:prompts-chords-controls-ukulele}
    CCC prompts for chord fingering: ukulele. \texttt{\textbf{\{Let's think step by step.\}}} is added only if 0-shot CoT is used.
    }
\end{table*}

\begin{table*}[t!]
    \centering
    \begin{tabular}{cl}
        \toprule
        Mode & Prompt \\
        \midrule
        Default & \texttt{\thead{You are an expert musician. What is the second note of the melody of the song 'Twinkle Twinkle\\Little Star' in \textbf{C major}? Make sure to choose one final answer, which you should start with\\ 'ANSWER:' and specify in the following format: NOTE=\{note\}. \\\\\textbf{\{Let's think step by step.\}}}} \\\midrule
        CF & \texttt{\thead{You are an expert musician. What is the second note of the melody of the song 'Twinkle Twinkle\\Little Star' in \textbf{Db major}? Make sure to choose one final answer, which you should start with\\ 'ANSWER:' and specify in the following format: NOTE=\{note\}. \\\\\textbf{\{Let's think step by step.\}}}} \\
        \bottomrule
    \end{tabular}
    \caption{\label{tab:prompts-songs}
    Prompts for melody retrieval. \texttt{\textbf{\{Let's think step by step.\}}} is added only if 0-shot CoT is used.} 
\end{table*}

\begin{table*}[t!]
    \centering
    \begin{tabular}{cl}
        \toprule
        Mode & Prompt \\
        \midrule
        Default & \texttt{\thead{You are an expert musician. What is the second note of the C major scale? Make sure to\\ choose one final answer, which you should start with 'ANSWER:' and specify in the\\ following format: NOTE=\{note\}. \\\\\textbf{\{Let's think step by step.\}}}} \\\midrule
        CF & \texttt{\thead{You are an expert musician. What is the second note of the Db major scale? Make sure to\\ choose one final answer, which you should start with 'ANSWER:' and specify in the\\ following format: NOTE=\{note\}. \\\\\textbf{\{Let's think step by step.\}}}}\\       
        \bottomrule
    \end{tabular}
    \caption{\label{tab:prompts-songs-controls}
    CCC prompts for melody retrieval. \texttt{\textbf{\{Let's think step by step.\}}} is added only if 0-shot CoT is used. 
    } 
\end{table*}

\begin{table*}[t!]
    \centering
    \begin{tabular}{cl}
        \toprule
        Mode & Prompt \\
        \midrule
        Default & \texttt{\thead{You are a chess player. Given an opening, determine whether the opening is legal. The \\opening doesn't need to be a good opening. Answer "yes" if all moves are legal. Answer "no"\\ if the opening violates any rules of chess. Is the new opening "1. e4 e6 2. Be2 Bc5" legal? \\\textbf{\{Let's think step by step\}}}} \\
        \midrule
        CF & \texttt{\thead{You are a chess player. You are playing a chess variant where the starting positions for\\ knights and bishops are swapped. For each color, the knights are at placed that where \\ bishops used to be and the bishops are now placed at where knights used to be. Given an \\opening, determine whether the opening is legal. The opening doesn't need to be a good\\ opening. Answer "yes" if all moves are legal. Answer "no" if the opening violates any\\ rules of chess. Under the custom variant, is the new opening "1. e4 e6 2. Nfe2 Nc5" legal? \\\textbf{\{Let's think step by step\}}}} \\
        \bottomrule
    \end{tabular}
    \caption{\label{tab:prompts-chess}
    Prompts for the chess task. \texttt{\textbf{\{Let's think step by step\}}} is added only if 0-shot CoT is used.
    }
\end{table*}

\begin{table*}[t!]
    \centering
    \begin{tabular}{cl}
        \toprule
        Mode & Prompt \\
        \midrule
        Default & \texttt{\thead{You are a chess player. Question: The two bishops on the board should be initially at \\which squares? Answer: \textbf{\{Let's think step by step\}}}} \\
        \midrule
        CF & \texttt{\thead{You are a chess player. You are playing a chess variant where the starting positions for\\ knights and bishops are swapped. For each color, the knights are at placed that where \\ bishops used to be and the bishops are now placed at where knights used to be. Question:\\ In this chess variant, the two bishops on the board should be initially at which squares?\\ Answer: \textbf{\{Let's think step by step\}}}} \\
        \bottomrule
    \end{tabular}
    \caption{\label{tab:prompts-chess-ccc}
    CCC prompts for the chess task. \texttt{\textbf{\{Let's think step by step\}}} is added only if 0-shot CoT is used.
    }
\end{table*}

\begin{table*}[t!]
    \centering
    \begin{tabular}{cl}
        \toprule
        Mode & Prompt \\
        \midrule
        Default & \texttt{\thead{You will be shown 12 cards. Each card has a figure and a number.\\
A figure is a combination of a color, a shape, and a fill.\\ Set of colors are: red , green , blue . \\Set of shapes are: squiggle , diamond , oval .\\\\
Set of fills are: solid , striped , open . \\ \\--THE RULE OF THE GAME-- \\A GAME-SET is set of three cards: For each attribute, (color, shape, fill, number), \\ the three cards should either be ALL the SAME or NONE the SAME (=ALL DIFFERENT, \\ e.g. if 2 of the cards have \\ the same value, and 1 of them has a different value, the set is NOT valid; for example, \\(blue, green, blue) is MIXED and does not satisfy any of the rule, \\whereas (oval, diamond, squiggle) is all different.\\\\Here is the board:\\
(2 | green | oval | open) ... [truncated]\\\\You can pick a set by typing the cards in the below format:\\
First card: CARD1\\
Second card: CARD2\\
Third card: CARD3\\
Now remember the rule and tell me which three cards here constitutes a GAME-SET \\in the same format.
\\I will give you 2 cards as a hint, and you tell me the third one.\\
First card: (2 | green | oval | open)\\
Second card: (1 | green | diamond | solid)\\
\textbf{\{Let's think step by step.\}}}} \\
        \midrule
        CF & \texttt{\thead{You will be shown 12 cards. Each card has a figure and a number.\\
A figure is a combination of a color, a shape, and a fill.\\ Set of colors are: red , green , blue . \\Set of shapes are: squiggle , diamond , oval .\\\\
Set of fills are: solid , striped , open . \\ \\--THE RULE OF THE GAME-- \\(This is not the original SET game. It has a tweaked rule.)\\
In this version, a GAME-SET is a set of three cards:\\
- For each figure attribute except the number (color, shape, fill), the three cards \\ should either be ALL the SAME or NONE the SAME (=ALL DIFFERENT, e.g. if 2 of the cards \\ have the same value, and 1 of them has a different value, the set is NOT valid; \\ for example, (blue, green, blue) is MIXED and does not satisfy any of the rule, whereas \\ (oval, diamond, squiggle) is all different.
\\- But only for the number attribute, 2 of the cards should have the same number, \\ and 1 of them should have a different number in order for the set to be valid.\\\\Here is the board:\\
(2 | green | oval | open) ... [truncated]\\\\You can pick a set by typing the cards in the below format:\\
First card: CARD1\\
Second card: CARD2\\
Third card: CARD3\\
Now remember the rule and tell me which three cards here constitutes a GAME-SET \\in the same format.
\\I will give you 2 cards as a hint, and you tell me the third one.\\
First card: (2 | green | oval | open)\\
Second card: (1 | green | diamond | solid)\\
\textbf{\{Let's think step by step.\}}}}\\    
    \bottomrule
    \end{tabular}
    \caption{\label{tab:prompts-set-test}
    Prompts for the SET task. \texttt{\textbf{\{Let's think step by step\}}} is added only if 0-shot CoT is used.
    }
\end{table*}

\begin{table*}[t!]
    \centering
    \begin{tabular}{cl}
        \toprule
        Mode & Prompt \\
        \midrule
        Default & \texttt{\thead{Each card has a figure and a number.\\
A figure is a combination of a color, a shape, and a fill.\\ Set of colors are: red , green , blue . \\Set of shapes are: squiggle , diamond , oval .\\\\
Set of fills are: solid , striped , open . \\ \\--THE RULE OF THE GAME-- \\A GAME-SET is set of three cards: For each attribute, (color, shape, fill, number), \\ the three cards should either be ALL the SAME or NONE the SAME (=ALL DIFFERENT, \\ e.g. if 2 of the cards have \\ the same value, and 1 of them has a different value, the set is NOT valid; for example, \\(blue, green, blue) is MIXED and does not satisfy any of the rule, \\whereas (oval, diamond, squiggle) is all different.\\\\
I will give you three cards from the board, and you will tell me whether this\\constitutes a GAME-SET. \\\\
First card: (1 | blue | oval | striped) \\
Second card: (2 | red | squiggle | striped) \\
Third card: (3 | green | diamond | striped) \\\\
Is this a GAME-SET? \textbf{\{Answer with yes or no in the last line.}\\\textbf{Let's verify rules for each attribute step-by-step:\}}}} \\
        \midrule
        CF & \texttt{\thead{Each card has a figure and a number.\\
A figure is a combination of a color, a shape, and a fill.\\ Set of colors are: red , green , blue . \\Set of shapes are: squiggle , diamond , oval .\\\\
Set of fills are: solid , striped , open . \\ \\--THE RULE OF THE GAME-- \\(This is not the original SET game. It has a tweaked rule.)\\
In this version, a GAME-SET is a set of three cards:\\
- For each figure attribute except the number (color, shape, fill), the three cards \\ should either be ALL the SAME or NONE the SAME (=ALL DIFFERENT, e.g. if 2 of the cards \\ have the same value, and 1 of them has a different value, the set is NOT valid; \\ for example, (blue, green, blue) is MIXED and does not satisfy any of the rule, whereas \\ (oval, diamond, squiggle) is all different.
\\- But only for the number attribute, 2 of the cards should have the same number, \\ and 1 of them should have a different number in order for the set to be valid.\\\\I will give you three cards from the board, and you will tell me whether this\\constitutes a GAME-SET. \\\\
First card: (1 | blue | oval | striped) \\
Second card: (2 | red | squiggle | striped) \\
Third card: (3 | green | diamond | striped) \\\\
Is this a GAME-SET? \textbf{\{Answer with yes or no in the last line.}\\\textbf{Let's verify rules for each attribute step-by-step:\}}}} \\ 
    \bottomrule
    \end{tabular}
    \caption{\label{tab:prompts-set-control}
    CCC prompts for the SET experiments. \texttt{\textbf{\{bold text\}}} is added only if 0-shot CoT is used. Note that we removed the board information for simplicity as it is not required for this CCC test.
    }
\end{table*}

\begin{table*}[t!]
    \centering
    \scalebox{0.8}{
    \begin{tabular}{lccccccccccccccc}
        \toprule
        & \multicolumn{10}{c}{Tests} & \multicolumn{5}{c}{\multirow{2}{*}{CCC}} \\
        \cmidrule(lr){2-11}
        & \multicolumn{5}{c}{w/o 0-CoT} & \multicolumn{5}{c}{w/ 0-CoT} \\
        \cmidrule(lr){2-6}
        \cmidrule(lr){7-11}
        \cmidrule(lr){12-16}
        Base & 8 & 9 & 10 & 11 & 16 & 8 & 9 & 10 & 11 & 16 & 8 & 9 & 10 & 11 & 16 \\
        \midrule
        \# instances & \multicolumn{10}{c}{1,000} & \multicolumn{5}{c}{200} \\
        \midrule
        GPT-4 & 82.3 & 23.4 & 100.0 & 38.4 & 63.0 & 60.2 & 38.6 & 98.2 & 56.5 & 74.0 & 98.0 & 90.0 & 100.0 & 91.0 & 100.0 \\
        GPT-3.5 & \phantom{0}8.3 & \phantom{0}6.6 & 100.0 & \phantom{0}3.8 & 17.7 & 12.6 & \phantom{0}9.8 & 99.0 & \phantom{0}2.7 & 17.7 & 96.5 & 77.0 & 100.0 & 56.0 & \phantom{0}95.5 \\
        Claude & 22.3 & \phantom{0}0.2 & \phantom{0}99.8 & \phantom{0}6.6 & 32.4 & \phantom{0}1.4 & \phantom{0}0.9 & 98.7 & \phantom{0}4.0 & \phantom{0}6.6 & 64.5 & 47.5 & 100.0 & 41.0 & \phantom{0}77.5 \\
        PaLM-2 & \phantom{0}6.4 & \phantom{0}2.2 & \phantom{0}98.7 & \phantom{0}3.4 & 23.4 & \phantom{0}1.1 & \phantom{0}0.6 & 82.2 & \phantom{0}0.5 & \phantom{0}1.2 & 51.5 & 53.5 & 100.0 & 72.0 & \phantom{0}93.5 \\
        \bottomrule        
    \end{tabular}
    }
    \caption{\label{tab:numbers-math}
    Results for the arithmetic task (in accuracy; \%).
    }
\end{table*}

\begin{table*}[t!]
    \centering
    \begin{tabular}{cc|ccccc}
        \toprule
        \# digits & \# shots & 8 & 9 & 10 & 11 & 16 \\
        \midrule
        2 & 0 & 60.2 & 38.6 & 98.2 & 56.5 & 74.0 \\
        3 & 0 & 56.8 & 32.2 & 87.1 & 24.2 & 33.2 \\
        4 & 0 & 24.0 & 14.6 & 83.4 & \phantom{0}8.9 & \phantom{0}9.1 \\
        2 & 1 & 97.3 & 48.1 & 99.7 & 25.7 & 49.1 \\
        2 & 2 & 99.1 & 67.0 & 99.9 & 44.0 & 57.8 \\
        2 & 4 & 99.4 & 79.7 & 99.9 & 68.4 & 70.6 \\
        2 & 8 & 99.7 & 85.8 & 100.0 & 79.6 & 83.5 \\
        2 & 16 & 99.9 & 88.4 & 99.9 & 86.9 & 88.7 \\
        \bottomrule        
    \end{tabular}
    \caption{\label{tab:numbers-math-analysis}
    Results for the arithmetic task for various analyses in \S\ref{sec:analysis} (in accuracy; \%). Only for GPT-4 with 0-shot CoT.
    }
\end{table*}

\begin{table*}[t!]
    \centering
    \begin{tabular}{lcccccccc}
        \toprule
        & \multicolumn{4}{c}{Tests} & \multicolumn{4}{c}{CCC} \\
        \cmidrule(lr){2-5}
        \cmidrule(lr){6-9}
        & \multicolumn{2}{c}{w/o 0-CoT} & \multicolumn{2}{c}{w/ 0-CoT} & \multicolumn{2}{c}{w/o 0-CoT} & \multicolumn{2}{c}{w/ 0-CoT} \\
        \cmidrule(lr){2-3}
        \cmidrule(lr){4-5}
        \cmidrule(lr){6-7}
        \cmidrule(lr){8-9} 
        & Default & CF & Default & CF & Default & CF & Default & CF \\
        \midrule
        \# instances & \multicolumn{4}{c}{113} & \multicolumn{4}{c}{750} \\
        \midrule
        GPT-4 & 58.4 & 18.6 & 73.5 & 24.8 & 95.3 & 78.1 & 99.7 & 90.9 \\
        GPT-3.5 & 39.8 & \phantom{0}9.7 & 54.0 & 10.6 & 97.1 & 21.3 & 94.1 & 25.9 \\
        Claude & 35.4 & 13.3 & 36.3 & \phantom{0}6.2 & 96.5 & 31.1 & 85.1 & 37.3 \\
        \bottomrule
    \end{tabular}
    \caption{
    Results for the programming execution task (in accuracy; \%).
    }
\end{table*}

\begin{table*}[t!]
    \centering
    \begin{tabular}{lcccccccc}
        \toprule
        & \multicolumn{4}{c}{HumanEval (All)} & \multicolumn{4}{c}{HumanEval (Subset)} \\
        \cmidrule(lr){2-5}
        \cmidrule(lr){6-9}
        & \multicolumn{2}{c}{pass@1} & \multicolumn{2}{c}{pass@10} & \multicolumn{2}{c}{pass@1} & \multicolumn{2}{c}{pass@10} \\
        \cmidrule(lr){2-3}
        \cmidrule(lr){4-5}
        \cmidrule(lr){6-7}
        \cmidrule(lr){8-9}
        & Default & CF & Default & CF & Default & CF & Default & CF \\
        \midrule
        \# instances & \multicolumn{4}{c}{164} & \multicolumn{4}{c}{53} \\
        \midrule
        GPT-4 & 87.4 & 68.2 & 95.3 & 83.4 & 82.5 & 40.5 & 93.3 & 64.9 \\
        GPT-3.5 & 73.8 & 41.8 & 88.4 & 67.6 & 68.9 & 25.1 & 81.0 & 45.8 \\
        Claude & 53.7 & 39.6 & 78.1 & 64.2 & 47.6 & 15.7 & 74.0 & 41.9 \\
        PaLM-2 & 27.3 & 20.8 & 55.8 & 42.6 & 29.2 & \phantom{0}7.4 & 55.3 & 21.0 \\
        \bottomrule
    \end{tabular}
    \caption{ \label{tab:programming-gen-results}
    Results for the programming generation task (in pass@1 and pass@10; \%).
    We report both the results on the entire HumanEval dataset for comparability with other work, as well as the subset on which evaluating the original program under 1-based indexing would not pass the test cases.
    Figure~\ref{fig:results-1} only showed the results on this subset.
    }
\end{table*}

\begin{table*}[t!]
    \centering
    \footnotesize
    \begin{tabular}{lcccccccccccc}
    \toprule 
     & \multicolumn{12}{c}{Test Accuracy} \\ \cmidrule(lr){2-13}
     & \multicolumn{6}{c}{w/o 0-CoT} & \multicolumn{6}{c}{w/ 0-CoT} \\
     \cmidrule(lr){2-7} \cmidrule(lr){8-13} \\
      & SVO & SOV & VSO & VOS & OVS & OSV  & SVO & SOV & VSO & VOS & OVS & OSV \\
     \midrule
     \# instances & \multicolumn{12}{c}{100} \\
    \midrule
    GPT-4 & 88.0 & 63.0 & 66.0 & 63.0 & 68.0 & 68.0 & 76.0 & 66.0 & 69.0 & 70.0 & 70.0 & 68.0 \\
    GPT-3.5 & 72.0 & 51.0 & 60.0 & 39.0 & 41.0 & 64.0 & 50.0 & 51.0 & 63.0 & 44.0 & 27.0 & 51.0 \\
    Claude & 55.0 & 58.0 & 65.0 & 62.0 & 59.0 & 56.0 & 51.0 & 57.0 & 62.0 & 62.0 & 62.0 & 59.0 \\
    PaLM-2 &  40.0 & 22.0 &  48.0 & 28.0 & 37.0 & 23.0 & 35.0 & 29.0 & 56.0 & 29.0 & 22.0 & 19.0 \\
    \midrule
     & \multicolumn{12}{c}{CCC} \\ \cmidrule(lr){2-13} & \multicolumn{6}{c}{w/o 0-CoT} & \multicolumn{6}{c}{w/ 0-CoT} \\
     \cmidrule(lr){2-7} \cmidrule(lr){8-13} \\
      & SVO & SOV & VSO & VOS & OVS & OSV  & SVO & SOV & VSO & VOS & OVS & OSV \\
     \midrule
     \# instances & \multicolumn{12}{c}{100} \\
    \midrule
    GPT-4 & -- & 100.0 & 92.0 & 98.0 & 100.0 & 94.0 & -- & 95.0 & 95.0 & 99.0 & 96.0 & 100.0 \\
    GPT-3.5 & -- & \phantom{0}88.0 & 72.0 & 25.0 & \phantom{0}61.0 & 64.0 & -- & 68.0 & 40.0 & 37.0 & \phantom{0}5.0 & \phantom{0}31.0 \\
    Claude & -- & \phantom{0}89.0 & 60.0 & 97.0 & \phantom{0}99.0 & 98.0 & -- & 79.0 & 91.0 & 80.0 & \phantom{0}8.0 & \phantom{0}87.0 \\
    PaLM-2 & -- & \phantom{00}7.0 & \phantom{0}8.0  & 96.0 & 100.0 & 62.0 & -- & 28.0 & 29.0 &91.0 & 94.0 & \phantom{0}70.0    \\
    \bottomrule
    \end{tabular}
    \caption{Results for the basic syntactic reasoning task (in accuracy; \%).
    }
    \label{tab:word_order_raw}
\end{table*}

\begin{table*}[t!]
    \centering
    \begin{tabular}{lccccc}
        \toprule
        & \multicolumn{4}{c}{Tests} & \multirow{3}{*}{CCC} \\
        \cmidrule(lr){2-5}
        & \multicolumn{2}{c}{w/o 0-CoT} & \multicolumn{2}{c}{w/ 0-CoT} \\
        \cmidrule(lr){2-3}
        \cmidrule(lr){4-5}
        & Default & CF & Default & CF \\
        \midrule
        \# instances & \multicolumn{4}{c}{81} & 310 \\
        \midrule
        GPT-4 & 93.8 & 74.1 & 98.8 & 82.7 & 97.4 \\
        GPT-3.5 & 79.0 & 43.2 & 65.4 & 42.0 & 90.3 \\
        Claude & 27.2 & 16.0 & 40.7 & 17.3 & 55.2 \\
        PaLM-2 & 84.0 & 66.7 & 88.9 & 74.1 & 70.6 \\
        \bottomrule
    \end{tabular}
    \caption{
    Results for the logical reasoning task (in accuracy; \%).
    }
\end{table*}

\begin{table*}[t!]
    \centering
    \scalebox{0.8}{
    \begin{tabular}{lcccccccccccccc}
    \toprule
     & \multicolumn{14}{c}{Tests Accuracy} \\
     \cmidrule(lr){2-15}
      & \multicolumn{7}{c}{w/o 0-CoT} & \multicolumn{7}{c}{w/ 0-CoT} \\
     \cmidrule(lr){2-8} \cmidrule(lr){9-15}
    & Default & S-NS & S-WE & R90 & R180 & R270 & Rand. & Default & S-NS & S-WE & R90 & R180 & R270 & Rand. \\
    \midrule
    \# instances &  \multicolumn{7}{c}{100} & \multicolumn{7}{c}{100}  \\
    \midrule
    GPT-4 & 79.0 & 57.0 & 29.0 & 34.0 & 6.0 & 22.0 & 34.0 & 98.0 & 71.0 & 23.0 & 24.0 & 9.0 & 13.0 & 13.0 \\
    GPT-3.5 & 87.0 & 56.0 & 32.0 & 27.0 & 12.0 & 17.0 & 15.0 & 82.0 & 66.0 & 36.0 & 27.0 & 29.0 & 22.0 & 22.0 \\
    Claude & 86.0 & 51.0 & 72.0 & 35.0 & 45.0 & 15.0 & 51.0 & 85.0 & 50.0 & 71.0 & 30.0 & 49.0 & 11.0 & 39.0 \\
    PaLM-2 & 90.0 & 88.0 & 86.0 & 50.0 & 93.0 & 39.0 & 64.0 & 84.0 & 95.0 & 80.0 & 38.0 & 91.0 & 38.0 & 54.0 \\
    \midrule
     & \multicolumn{14}{c}{Tests Object-level Accuracy} \\
     \cmidrule(lr){2-15}
      & \multicolumn{7}{c}{w/o 0-CoT} & \multicolumn{7}{c}{w/ 0-CoT} \\
     \cmidrule(lr){2-8} \cmidrule(lr){9-15}
    & Default & S-NS & S-WE & R90 & R180 & R270 & Rand. & Default & S-NS & S-WE & R90 & R180 & R270 & Rand. \\
    \midrule
    \# instances &  \multicolumn{7}{c}{100} & \multicolumn{7}{c}{100}  \\
    \midrule
    GPT-4 & 86.0 & 74.3 & 55.7 & 56.0 & 34.0 & 53.0 & 61.7 & 99.0 & 85.3 & 57.0 & 49.7 & 36.0 & 46.3 & 46.0 \\
    GPT-3.5 & 92.3 & 77.7 & 62.3 & 54.7 & 41.3 & 42.3 & 47.0 & 92.7 & 82.7 & 64.0 & 53.0 & 50.7 & 54.7 & 53.3 \\
    Claude & 93.7 & 75.3 & 87.7 & 65.7 & 70.3 & 46.7 & 76.0 & 91.7 & 74.7 & 86.3 & 63.0 & 73.0 & 44.0 & 69.3 \\
    PaLM-2 & 96.3 & 95.7 & 94.3 & 71.3 & 97.7 & 64.7 & 79.7 & 94.0 & 98.3 & 91.0 & 63.7 & 96.0 & 64.0 & 75.0 \\
    \midrule
     & \multicolumn{14}{c}{CCC} \\
     \cmidrule(lr){2-15}
      & \multicolumn{7}{c}{w/o 0-CoT} & \multicolumn{7}{c}{w/ 0-CoT} \\
     \cmidrule(lr){2-8} \cmidrule(lr){9-15}
    & Default & S-NS & S-WE & R90 & R180 & R270 & Rand. & Default & S-NS & S-WE & R90 & R180 & R270 & Rand. \\
    \midrule
    \# instances &  \multicolumn{7}{c}{100} & \multicolumn{7}{c}{100}  \\
    \midrule
    GPT-4 & 100.0 & \phantom{0}99.0 & 100.0 & 100.0 & 100.0 & 100.0 & 100.0 & 100.0 & 100.0 & 100.0 & 100.0 & 100.0 & 100.0 & 100.0 \\
    GPT-3.5 & 100.0 & 100.0 & \phantom{0}92.0 & \phantom{0}83.0 & \phantom{0}90.0 & 100.0 & \phantom{0}99.0 & \phantom{0}98.0 & \phantom{0}89.0 & \phantom{0}88.0 & \phantom{0}83.0 & \phantom{0}78.0 & \phantom{0}94.0 & \phantom{0}92.0 \\
    Claude & 100.0 & 100.0 & 100.0 & 100.0 & 100.0 & 100.0 & \phantom{0}99.0 & 100.0 & 100.0 & 100.0 & 100.0 & 100.0 & 100.0 & \phantom{0}91.0 \\
    PaLM-2 & 100.0 & 100.0 & 100.0 & 100.0 & 100.0 & \phantom{0}39.0 & 100.0 & 100.0 & 100.0 & 100.0 & 100.0 & 100.0 & \phantom{0}59.0 & 100.0 \\
    \bottomrule
    \end{tabular}
    }
    \caption{\label{tab:numbers-spatial}Results for the spatial reasoning task (in accuracy; \%). The first section (Tests Accuracy) requires all 3 objects to be correctly placed. The second section (Test Object-Level Accuracy) refers to accuracy averaged over objects. `S' denotes to swapping, `R' denotes to rotation, `Rand.' denotes to random permutation.}
\end{table*}

\begin{table*}[t!]
    \centering
    \scalebox{0.8}{
    \begin{tabular}{lcccccccc}
    \toprule
    & \multicolumn{8}{c}{Human Evaluation Binary Classification Test Accuracy}  \\
     \cmidrule(lr){2-9}
     & \multicolumn{4}{c}{w/o 0-CoT} & \multicolumn{4}{c}{w/ 0-CoT}  \\
     \cmidrule(lr){2-5} \cmidrule(lr){6-9} 
     & Default & VFlip & R90 & R180 & Default & Flip & R90 & R180  \\
     \midrule
    \# instances &  \multicolumn{8}{c}{100}  \\
    \midrule
    GPT-4 & 82.0 & 57.0 & 59.0 & 53.0 & 86.0 & 51.0 & 47.0 & 52.0  \\
    GPT-3.5 & 31.0 & 19.0 & \phantom{0}9.0 & \phantom{0}8.0 & 46.0 & \phantom{0}7.0 & \phantom{0}5.0 & \phantom{0}8.0  \\
    Claude & 42.0 & 13.0 & 15.0 & 13.0 & 34.0 & 15.0 & 30.0 & 12.0  \\
    \midrule
     & \multicolumn{8}{c}{CLIP Evaluation Multi-class Classification Tests Accuracy}  \\
    \cmidrule(lr){2-9}
     & \multicolumn{4}{c}{w/o 0-CoT} & \multicolumn{4}{c}{w/ 0-CoT}  \\
     \cmidrule(lr){2-5} \cmidrule(lr){6-9} 
     & Default & VFlip & R90 & R180 & Default & Flip & R90 & R180  \\
     \midrule
    \# instances &  \multicolumn{8}{c}{100}  \\
    \midrule
    GPT-4 & 58.0 & 30.0 & 23.0 & 31.0 & 53.0 & 26.0 & 33.0 & 28.0 \\
    GPT-3.5 & 20.0 & \phantom{0}9.0 & \phantom{0}7.0 & \phantom{0}8.0 & 32.0 & \phantom{0}4.0 & \phantom{0}6.0 & \phantom{0}6.0  \\
    Claude & 19.0 & 11.0 & \phantom{0}9.0 & \phantom{0}8.0 & 18.0 & \phantom{0}9.0 & 13.0 & \phantom{0}8.0  \\
    \midrule
    & \multicolumn{8}{c}{CCC Accuracy}  \\
    \cmidrule(lr){2-9}
     & \multicolumn{4}{c}{w/o 0-CoT} & \multicolumn{4}{c}{w/ 0-CoT}  \\
     \cmidrule(lr){2-5} \cmidrule(lr){6-9} 
     & Default & VFlip & R90 & R180 & Default & Flip & R90 & R180  \\
     \midrule
    \# instances &  \multicolumn{8}{c}{100}  \\
    \midrule
    GPT-4 &  100.0 & 99.0 & 55.0 & 89.0 & 100.0 & 99.0 & 87.0 & 87.0 \\
    GPT-3.5 & \phantom{0}78.0 & 56.0 & 53.0 & 44.0 & \phantom{0}99.0 & 62.0 & 43.0 & 46.0 \\
    Claude &  100.0 & 84.0 & 86.0 & 57.0 & \phantom{0}99.0 & 90.0 & 99.0 & 54.0 \\
    \bottomrule
    \end{tabular}
}
    \caption{Results for the drawing task (in accuracy; \%). VFlip corresponds to vertical flipping, R90 and R180 correspond to rotation by 90 degrees and 180 degrees, respectively.}
    \label{tab:numbers-drawing}
\end{table*}

\begin{table*}[t!]
    \centering
    \scalebox{0.8}{
    \begin{tabular}{lcccccccc}
    \toprule
    & \multicolumn{8}{c}{Accuracy for Objects w/o Canonical Orientation}  \\
     \cmidrule(lr){2-9}
     & \multicolumn{4}{c}{w/o 0-CoT} & \multicolumn{4}{c}{w/ 0-CoT}  \\
     \cmidrule(lr){2-5} \cmidrule(lr){6-9} 
     & Default & VFlip & R90 & R180 & Default & Flip & R90 & R180  \\
    \midrule
    GPT-4   &  72.1 &   68.6 & 67.3 &  57.4 &  76.2 &   58.0 & 69.8 &  63.3 \\
    GPT-3.5 &  31.8 &   22.0 & 14.3 &  13.0 &  39.0 &    \phantom{0}4.8 &  \phantom{0}7.1 &   \phantom{0}7.0 \\
    Claude  &  51.1 &   22.9 & 20.9 &  23.3 &  39.5 &   27.9 & 34.8 &  19.1 \\
    \midrule
     & \multicolumn{8}{c}{Accuracy for Objects w/ Canonical Orientation}  \\
    \cmidrule(lr){2-9}
     & \multicolumn{4}{c}{w/o 0-CoT} & \multicolumn{4}{c}{w/ 0-CoT}  \\
     \cmidrule(lr){2-5} \cmidrule(lr){6-9} 
     & Default & VFlip & R90 & R180 & Default & Flip & R90 & R180  \\
     \midrule
    GPT-4   &  89.5 &   44.9 & 50.0 &  49.1 &  93.1 &   44.0 & 21.3 &  41.2 \\
    GPT-3.5 &  30.4 &   16.0 &  \phantom{0}5.2 &   \phantom{0}3.7 &  50.8 &    \phantom{0}8.6 &  \phantom{0}3.4 &   \phantom{0}8.8 \\
    Claude  &  34.5 &    \phantom{0}3.8 & 10.5 &   \phantom{0}5.3 &  29.8 &    \phantom{0}5.3 & 25.9 &   \phantom{0}5.7 \\
    \bottomrule
    \end{tabular}
}
    \caption{Results for the drawing task, as measured by human evaluation accuracy (\%), broken down by objects with or without a canonical orientation as judged by human annotators. If an object has a canonical orientation, such as the house in Figure~\ref{fig:drawing_vis}, it is only considered correct if the orientation is correct.}
    \label{tab:numbers-drawing-breakdown}
\end{table*}

\begin{table*}[t!]
    \centering
    \footnotesize
    \begin{tabular}{lcccccccccccc}
    \toprule 
     & \multicolumn{12}{c}{Test Accuracy} \\ \cmidrule(lr){2-13}
     & \multicolumn{6}{c}{w/o 0-CoT} & \multicolumn{6}{c}{w/ 0-CoT} \\
     \cmidrule(lr){2-7} \cmidrule(lr){8-13} \\
     & Default & \textbf{D}AD- & \textbf{F}AD- & E\textbf{B}D- & E\textbf{C}D- & E\textbf{CF}- & Default & \textbf{D}AD- & \textbf{F}AD- & E\textbf{B}D- & E\textbf{C}D- & E\textbf{CF}- \\
     \midrule
     \# instances & \multicolumn{12}{c}{120} \\
    \midrule
    GPT-4 & 47.5 & 22.5 & 11.7 & 18.3 & 7.5 & 1.7 & 42.5 & 25.8 & 12.5 & 24.2 & 14.2 & 10.0 \\
    GPT-3.5 & 30.8 & \phantom{0}4.2 & \phantom{0}5.0 & \phantom{0}5.8 & 3.3 & 1.7 & 30.8 & \phantom{0}5.0 & \phantom{0}5.0 & \phantom{0}0.8 & \phantom{0}1.7 & \phantom{0}1.7 \\
    Claude & \phantom{0}5.0 & \phantom{0}0.8 & \phantom{0}0.8 & \phantom{0}0.8 & 0.8 & 1.7 & 10.0 & \phantom{0}5.8 & \phantom{0}4.2 & \phantom{0}5.8 & \phantom{0}5.0 & \phantom{0}0.8 \\
    PaLM-2 & \phantom{0}0.8 & \phantom{0}0.0 & \phantom{0}0.0 & \phantom{0}0.0 & 0.0 & 0.0 & \phantom{0}0.8 & \phantom{0}0.0 & \phantom{0}0.0 & \phantom{0}0.0 & \phantom{0}0.8 & \phantom{0}0.8 \\
    \midrule
     & \multicolumn{12}{c}{CCC} \\ \cmidrule(lr){2-13} & \multicolumn{6}{c}{w/o 0-CoT} & \multicolumn{6}{c}{w/ 0-CoT} \\
     \cmidrule(lr){2-7} \cmidrule(lr){8-13} \\
     & Default & \textbf{D}AD- & \textbf{F}AD- & E\textbf{B}D- & E\textbf{C}D- & E\textbf{CF}- & Default & \textbf{D}AD- & \textbf{F}AD- & E\textbf{B}D- & E\textbf{C}D- & E\textbf{CF}- \\
     \midrule
     \# instances & \multicolumn{12}{c}{18} \\
    \midrule
    GPT-4 & 100.0 & 100.0 & 94.4 & 94.4 & 100.0 & 83.3 & 100.0 & 100.0 & 100.0 & 100.0 & 100.0 & 100.0 \\
    GPT-3.5 & \phantom{0}50.0 & \phantom{0}27.8 & 11.1 & 44.4 & \phantom{0}50.0 & 61.1 & \phantom{0}94.4 & \phantom{0}66.7 & \phantom{0}88.9 & \phantom{0}88.9 & \phantom{0}94.4 & \phantom{0}77.8 \\
    Claude & \phantom{0}77.8 & \phantom{0}38.9 & 55.6 & 55.6 & \phantom{0}50.0 & 50.0 & 100.0 & \phantom{0}38.9 & \phantom{0}55.6 & \phantom{0}55.6 & \phantom{0}50.0 & \phantom{0}50.0 \\
    PaLM-2 & \phantom{0}33.3 & \phantom{0}33.3 & 33.3 & 38.9 & \phantom{0}50.0 & 38.9 & \phantom{0}33.3 & \phantom{0}33.3 & \phantom{0}50.0 & \phantom{0}38.9 & \phantom{0}38.9 & \phantom{0}38.9 \\
    \bottomrule
    \end{tabular}
    \caption{Results for the chord fingering task (in accuracy; \%): guitar. Default corresponds to EADGBE. Counterfactuals show the first three strings (the remaining three strings, GBE, are the same).
    }
\end{table*}

\begin{table*}[t!]
    \centering
    \footnotesize
    \begin{tabular}{lcccccccccccc}
    \toprule 
     & \multicolumn{12}{c}{Test Accuracy} \\ \cmidrule(lr){2-13}
     & \multicolumn{6}{c}{w/o 0-CoT} & \multicolumn{6}{c}{w/ 0-CoT} \\
     \cmidrule(lr){2-7} \cmidrule(lr){8-13} \\
     & Default & \textbf{D}AD- & \textbf{F}AD- & E\textbf{B}D- & E\textbf{C}D- & E\textbf{CF}- & Default & \textbf{D}AD- & \textbf{F}AD- & E\textbf{B}D- & E\textbf{C}D- & E\textbf{CF}- \\
     \midrule
     \# instances & \multicolumn{12}{c}{12} \\
    \midrule
    maj triad & 66.7 & 50.0 & 33.3 & 41.7 & \phantom{0}8.3 & \phantom{0}0.0 & 58.3 & 41.7 & \phantom{0}8.3 & 41.7 & \phantom{0}8.3 & \phantom{0}0.0 \\
    min triad & 58.3 & 33.3 & 25.0 & 41.7 & 25.0 & \phantom{0}0.0 & 58.3 & 25.0 & 16.7 & 25.0 & 16.7 & \phantom{0}8.3 \\
    5 & 83.3 & 41.7 & 33.3 & 16.7 & \phantom{0}0.0 & \phantom{0}0.0 & 75.0 & 50.0 & 25.0 & 33.3 & 33.3 & 50.0 \\
    dom7 & 25.0 & 25.0 & 16.7 & \phantom{0}8.3 & 25.0 & \phantom{0}0.0 & 33.3 & 16.7 & \phantom{0}8.3 & 25.0 & 16.7 & \phantom{0}0.0 \\
    6 & 25.0 & 16.7 & \phantom{0}0.0 & 16.7 & \phantom{0}0.0 & \phantom{0}0.0 & 33.3 & 25.0 & \phantom{0}8.3 & 16.7 & \phantom{0}8.3 & \phantom{0}8.3 \\
    sus4 & 50.0 & 33.3 & \phantom{0}0.0 & 33.3 & \phantom{0}0.0 & \phantom{0}0.0 & 58.3 & 33.3 & 16.7 & 33.3 & 25.0 & 25.0 \\
    dim7 & 25.0 & \phantom{0}0.0 & \phantom{0}0.0 & \phantom{0}0.0 & \phantom{0}8.3 & 16.7 & \phantom{0}0.0 & \phantom{0}0.0 & \phantom{0}8.3 & 16.7 & \phantom{0}0.0 & \phantom{0}0.0 \\
    aug7 & \phantom{0}0.0 & \phantom{0}0.0 & \phantom{0}0.0 & \phantom{0}0.0 & \phantom{0}0.0 & \phantom{0}0.0 & \phantom{0}8.3 & \phantom{0}0.0 & \phantom{0}0.0 & 16.7 & 16.7 & \phantom{0}0.0 \\
    sus2 & 58.3 & \phantom{0}0.0 & \phantom{0}8.3 & 16.7 & \phantom{0}0.0 & \phantom{0}0.0 & 58.3 & 50.0 & 16.7 & 25.0 & 16.7 & \phantom{0}8.3 \\
    min7 & 83.3 & 25.0 & \phantom{0}0.0 & \phantom{0}8.3 & \phantom{0}8.3 & \phantom{0}0.0 & 41.7 & 16.7 & 16.7 & \phantom{0}8.3 & \phantom{0}0.0 & \phantom{0}0.0 \\
    \bottomrule
    \end{tabular}
    \caption{Results broken down by chords for the chord fingering task as analyzed in \S\ref{sec:analysis} (in accuracy; \%): guitar, GPT-4. Default corresponds to EADGBE. Counterfactuals show the first three strings (the remaining three strings, GBE, are the same).
    }
\end{table*}

\begin{table*}[t!]
    \centering
    \footnotesize
    \begin{tabular}{lcccccccccc}
    \toprule 
     & \multicolumn{10}{c}{Test Accuracy} \\ \cmidrule(lr){2-11}
     & \multicolumn{5}{c}{w/o 0-CoT} & \multicolumn{5}{c}{w/ 0-CoT} \\
     \cmidrule(lr){2-6} \cmidrule(lr){7-11} \\
     & Default & \textbf{F}C- & \textbf{A}C- & \textbf{B}C- & \textbf{BE}-& Default & \textbf{F}C- & \textbf{A}C- & \textbf{B}C- & \textbf{BE}- \\
     \midrule
     \# instances & \multicolumn{10}{c}{108} \\
    \midrule
    GPT-4 & 39.8 & 1.9 & 1.9 & 2.8 & 0.9 & 20.4 & 2.8 & 16.7 & 11.1 & 10.2\\
    GPT-3.5 & 14.8 & 0.0 & 2.8 & 3.7 & 0.0 & 6.5 & \phantom{0}1.9 & \phantom{0}2.8 & \phantom{0}1.9 & \phantom{0}0.0\\
    Claude & \phantom{0}0.0 & 1.9 & 0.0 & 0.0 & 0.9 & 6.5 & \phantom{0}0.0 & \phantom{0}2.8 & \phantom{0}1.9 & \phantom{0}2.8\\
    PaLM-2 & \phantom{0}0.0 & 0.0 & 0.0 & 0.0 & 0.0 & 0.9 & \phantom{0}0.0 & \phantom{0}1.9 & \phantom{0}0.9 & \phantom{0}0.0\\
    \midrule
     & \multicolumn{10}{c}{CCC} \\ \cmidrule(lr){2-11} & \multicolumn{5}{c}{w/o 0-CoT} & \multicolumn{5}{c}{w/ 0-CoT} \\
     \cmidrule(lr){2-6} \cmidrule(lr){7-11} \\
      & Default & \textbf{F}C- & \textbf{A}C- & \textbf{B}C- & \textbf{BE}-& Default & \textbf{F}C- & \textbf{A}C- & \textbf{B}C- & \textbf{BE}- \\
     \midrule
     \# instances & \multicolumn{10}{c}{12} \\
    \midrule
    GPT-4 & 100.0 & 100.0 & 100.0 & 100.0 & 100.0 & 100.0 & 100.0 & 100.0 & 100.0 & 100.0\\
    GPT-3.5 & \phantom{0}33.3 & \phantom{0}41.7 & \phantom{0}91.7 & \phantom{0}41.7 & 100.0 & \phantom{0}91.7 & \phantom{0}83.3 & 100.0 & \phantom{0}75.0 & \phantom{0}75.0 \\
    Claude & \phantom{0}58.3 & \phantom{0}50.0 & \phantom{0}50.0 & \phantom{0}41.7 & \phantom{0}33.3 & 100.0 & 100.0 & \phantom{0}75.0 & \phantom{0}66.7 & \phantom{0}50.0\\
    PaLM-2 & \phantom{0}41.7 & \phantom{0}41.7 & \phantom{0}41.7 & \phantom{0}50.0 & \phantom{0}33.3 & \phantom{0}41.7 & \phantom{0}50.0 & \phantom{0}41.7 & \phantom{0}58.3 & \phantom{0}33.3\\
    \bottomrule
    \end{tabular}
    \caption{Results for the chord fingering task (in accuracy; \%): ukulele. Default corresponds to GCEA. Counterfactuals show the first two strings (the remaining two strings, EA, are the same).
    }
\end{table*}

\begin{table*}[t!]
    \centering
    \begin{tabular}{lcccccccc}
        \toprule
        & \multicolumn{4}{c}{Tests} & \multicolumn{4}{c}{CCC} \\
        \cmidrule(lr){2-5}
        \cmidrule(lr){6-9}
        & \multicolumn{2}{c}{w/o 0-CoT} & \multicolumn{2}{c}{w/ 0-CoT} & \multicolumn{2}{c}{w/o 0-CoT} & \multicolumn{2}{c}{w/ 0-CoT} \\
        \cmidrule(lr){2-3}
        \cmidrule(lr){4-5}
        \cmidrule(lr){6-7}
        \cmidrule(lr){8-9} 
        & Default & CF & Default & CF & Default & CF & Default & CF \\
        \midrule
        \# instances & 56 & 1120 & 56 & 1120 & 7 & 140 & 7 & 140 \\
        \midrule
        GPT-4 & 48.2 & 20.4 & 64.3 & 44.9 & 100.0 & 87.1 & 100.0 & 94.3\\
        GPT-3.5 & 17.9 & 17.4 & 44.6 & 21.8 & 100.0 & 70.0 & 100.0 & 95.7\\
        Claude & 17.9 & 17.2 & 16.1 & 20.1 & 100.0 & 74.3 & 100.0 & 80.7\\
        PaLM-2 & 28.6 & 19.5 & 19.6 & 19.3 & 100.0 & 48.6 & 100.0 & 47.1\\
        \bottomrule
    \end{tabular}
    \caption{
    Results for the melody retrieval task (in accuracy; \%). Default corresponds to C major, and CF corresponds to other keys.}
\end{table*}

\begin{table*}[t!]
    \centering
    \begin{tabular}{lcccccccc}
        \toprule
        & \multicolumn{4}{c}{Tests} & \multicolumn{4}{c}{CCC} \\
        \cmidrule(lr){2-5}
        \cmidrule(lr){6-9}
        & \multicolumn{2}{c}{w/o 0-CoT} & \multicolumn{2}{c}{w/ 0-CoT} & \multicolumn{2}{c}{w/o 0-CoT} & \multicolumn{2}{c}{w/ 0-CoT} \\
        \cmidrule(lr){2-3}
        \cmidrule(lr){4-5}
        \cmidrule(lr){6-7}
        \cmidrule(lr){8-9} 
        & Default & CF & Default & CF & Default & CF & Default & CF \\
        \midrule
        \# instances & 8 & 160 & 8 & 160 & 1 & 20 & 1 & 20 \\
        \midrule
        1 & 87.5 & 63.1 & 87.5 & 70.0 & 100.0 & \phantom{0}95.0 & 100.0 & \phantom{0}95.0\\
        2 & 87.5 & 10.6 & 62.5 & 46.9 & 100.0 & \phantom{0}95.0 & 100.0 & 100.0\\
        3 & 12.5 & 11.9 & 62.5 & 51.2 & 100.0 & \phantom{0}80.0 & 100.0 & \phantom{0}90.0\\
        4 & 25.0 & 21.2 & 62.5 & 42.5 & 100.0 & 100.0 & 100.0 & \phantom{0}85.0\\
        5 & 50.0 & \phantom{0}3.1 & 37.5 & 35.0 & 100.0 & \phantom{0}95.0 & 100.0 & 100.0\\
        6 & 37.5 & 5.0 & 75.0 & 38.8 & 100.0 & \phantom{0}75.0 & 100.0 & \phantom{0}95.0\\
        7 & 37.5 & 27.5 & 62.5 & 30.0 & 100.0 & \phantom{0}70.0 & 100.0 & \phantom{0}95.0\\
        \bottomrule
    \end{tabular}
    \caption{
    Results broken down by \emph{n} for the melody retrieval task as analyzed in \S\ref{sec:analysis} (in accuracy; \%): GPT-4. Default corresponds to C major, and CF corresponds to other keys.}
\end{table*}

\begin{table*}[t!]
    \centering
    \begin{tabular}{lcccccccc}
        \toprule
        & \multicolumn{4}{c}{Tests} & \multicolumn{4}{c}{CCC} \\
        \cmidrule(lr){2-5}
        \cmidrule(lr){6-9}
        & \multicolumn{2}{c}{w/o 0-CoT} & \multicolumn{2}{c}{w/ 0-CoT} & \multicolumn{2}{c}{w/o 0-CoT} & \multicolumn{2}{c}{w/ 0-CoT} \\
        \cmidrule(lr){2-3}
        \cmidrule(lr){4-5}
        \cmidrule(lr){6-7}
        \cmidrule(lr){8-9} 
        & Default & CF & Default & CF & Default & CF & Default & CF \\
        \midrule
        \# instances & \multicolumn{4}{c}{400} & \multicolumn{4}{c}{120} \\
        \midrule
        GPT-4 & 73.8 & 50.0 & 87.8 & 53.6 & 100.0 & 100.0 & 100.0 & 100.0 \\
        GPT-3.5 & 60.1 & 54.4 & 60.2 & 48.9 & \phantom{0}85.8 & \phantom{0}90.0 & \phantom{0}91.7 & \phantom{0}93.3 \\
        Claude & 59.2 & 50.0 & 60.2 & 50.4 & 100.0 & \phantom{0}50.0 & \phantom{0}76.7 & 100.0 \\
        PaLM-2 & 48.0 & 51.5 & 48.5 & 51.2 & \phantom{0}61.7 & \phantom{0}23.3 & \phantom{0}50.0 & \phantom{0}23.3 \\
        \bottomrule
    \end{tabular}
    \caption{
    Results for the chess task with $4$ moves (in accuracy; \%). CF refers to the setting where the initial positions of knights and bishops are swapped. We generate a balanced classification problem with $400$ openings via procedure generation.
    }
\end{table*}

\begin{table*}[t!]
    \centering
    \begin{tabular}{lcccccccc}
        \toprule
        & \multicolumn{4}{c}{Tests} & \multicolumn{4}{c}{CCC} \\
        \cmidrule(lr){2-5}
         \cmidrule(lr){6-9}
        & \multicolumn{2}{c}{w/o 0-CoT} & \multicolumn{2}{c}{w/ 0-CoT} & \multicolumn{2}{c}{w/o 0-CoT} & \multicolumn{2}{c}{w/ 0-CoT}  \\
        \cmidrule(lr){2-3}
        \cmidrule(lr){4-5}
         \cmidrule(lr){6-7}
          \cmidrule(lr){8-9}
        & Default & CF & Default & CF & Default & CF & Default & CF\\
        \midrule
        \# instances & \multicolumn{4}{c}{100} & \multicolumn{4}{c}{100}  \\
        \midrule
        GPT-4 & 100.0 &	21.0&100.0 &	61.0 & 89.0	& 74.0& 100.0 & 	96.0 \\
        GPT-3.5 & \phantom{0}73.0 &	4.0	& \phantom{0}37.0	& \phantom{0}7.0 & 68.0 &	55.0& \phantom{0}77.0 &	78.0\\
        Claude & \phantom{0}55.0	& 21.0	&\phantom{0}64.0 & 35.0 &92.0 & 62.0& \phantom{0}59.0	& 65.0 \\
        PaLM-2 & \phantom{0}55.0 &	17.0&\phantom{0}62.0	 & 13.0 &67.0 &	68.0& \phantom{0}47.0	& 46.0 \\
        \bottomrule
    \end{tabular}
    \caption{Results for the SET game (in accuracy; \%).}
\end{table*}

\begin{table*}[t!]
    \centering
    \begin{tabular}{lcccccccc}
        \toprule
        & \multicolumn{6}{c}{Tests} \\
        \midrule
        & Default (c=1) & CF (c=1) & Default (c=2)& CF (c=2) & Default (c=3) & CF (c=3) \\
        \midrule
        \# instances & \multicolumn{6}{c}{100}  \\
        \midrule
        GPT-4 & 100.0	&61.0	&24.0	&6.0	&15.0	&3.0 \\
        GPT-3.5 & \phantom{0}37.0	&\phantom{0}7.0	&\phantom{0}7.0	&0.0	&\phantom{0}1.0	&0.0 \\
        Claude & \phantom{0}64.0	&35.0	&10.0	&4.0	&\phantom{0}5.0	&1.0 \\
        PaLM-2 & \phantom{0}62.0	&13.0	&10.0	&1.0	&\phantom{0}3.0	&1.0 \\
        \bottomrule
    \end{tabular}
    \caption{Breakdown for the SET game test results (with 0-CoT) when the model needs to find different number of cards (c) in a SET as analyzed in \S\ref{sec:analysis} (in accuracy; \%).}
    \label{tab:numbers-set-hints}
\end{table*}

\end{document}